\pdfoutput=1

\ifdefined\usemsstyle

\documentclass[mnsc,blindrev]{informs3} % current default for manuscript submission

\OneAndAHalfSpacedXI % current default line spacing
\usepackage{natbib}
 \bibpunct[, ]{(}{)}{,}{a}{}{,}%
 \def\bibsep{\smallskipamount}%
\TheoremsNumberedThrough     % Preferred (Theorem 1, Lemma 1, Theorem 2)
\ECRepeatTheorems

\EquationsNumberedThrough    % Default: (1), (2), ...
\MANUSCRIPTNO{}

\usepackage{hyperref}
\usepackage{microtype}
\usepackage{booktabs} % for professional tables
\usepackage{./statistics-macros-ms}
\usepackage{pgfplotstable}
\usepackage{graphicx}
\usepackage{subcaption}
\usepackage{float}
\usepackage{soul,color}

\usepackage{algorithm}
\usepackage{algorithmic}

\usepackage{overpic}
\usepackage{tikz}
\usepackage{rotating}
\usepackage{psfrag}
\usepackage{bm}
\usepackage{placeins}
\usepackage{textcomp}
\usepackage{url}
\def\UrlBreaks{\do\/\do-}

\providecommand{\comment}[1]{}

\usepackage{manyfoot}

\DeclareNewFootnote{A}
\DeclareNewFootnote{B}

\let\footnote\footnoteA

\begin{document}
%%%%%%%%%%%%%%%%

%\setcounter{page}{1}
%\DOI{10.1287/xxxx.0000.0000}%


\RUNTITLE{Assessing External Validity Over Worst-case Subpopulations}

% \TITLE{Bundling Information Goods of Decreasing Value}
\TITLE{Assessing External Validity Over Worst-case Subpopulations}

\ARTICLEAUTHORS{%
  \AUTHOR{Sookyo Jeong}
\AFF{Lyft Inc., \EMAIL{sjeong@lyft.com}} %, \URL{}}

\AUTHOR{Hongseok Namkoong}
\AFF{Decision, Risk, and Operations Division, Columbia Business School, New York, NY 10027, \EMAIL{namkoong@gsb.columbia.edu}} , \URL{hsnamkoong.github.io}
} % end of the block

\ABSTRACT{Study populations are typically sampled from limited points in space and time,
and marginalized groups are underrepresented. To assess the external validity
of randomized and observational studies, we propose and evaluate the
worst-case treatment effect (WTE) across all subpopulations of a given size,
which guarantees positive findings remain valid over subpopulations. We
develop a semiparametrically efficient estimator for the WTE that analyzes the
external validity of the augmented inverse propensity weighted estimator for
the average treatment effect.  Our cross-fitting procedure leverages flexible
nonparametric and machine learning-based estimates of nuisance parameters and
is a regular root-$n$ estimator even when nuisance estimates converge more
slowly. On real examples where external validity is of core concern, our
proposed framework guards against brittle findings that are invalidated by
unanticipated population shifts.

%%% local Variables:
%%% mode: latex
%%% TeX-master: "main"
%%% End:

}

%\KEYWORDS{deterministic inventory theory; infinite linear programming duality;
%  existence of optimal policies; semi-Markov decision process; cyclic schedule}

\HISTORY{This paper was first submitted on Jan, 2022.}

\maketitle
%%%%%%%%%%%%%%%%%%%%%%%%%%%%%%%%%%%%%%%%%%%%%%%%%%%%%%%%%%%%%%%%%%%%%%

% Samples of sectioning (and labeling) in MNSC
% NOTE: (1) \section and \subsection do NOT end with a period
%       (2) \subsubsection and lower need end punctuation
%       (3) capitalization is as shown (title style).
%
%\section{Introduction.}\label{intro} %%1.
%\subsection{Duality and the Classical EOQ Problem.}\label{class-EOQ} %% 1.1.
%\subsection{Outline.}\label{outline1} %% 1.2.
%\subsubsection{Cyclic Schedules for the General Deterministic SMDP.}
%  \label{cyclic-schedules} %% 1.2.1
%\section{Problem Description.}\label{problemdescription} %% 2.

% Text of your paper here

\else

\documentclass[11pt]{article}
\usepackage[numbers]{natbib}
\usepackage{fullpage}
\usepackage{./statistics-macros}
\usepackage{setspace}

\usepackage{hyperref}
\usepackage{pgfplotstable}
\usepackage{graphicx}
\usepackage{subcaption}
\usepackage{float}

\usepackage{algorithm}
\usepackage{algpseudocode}
\usepackage{tabularx}

\usepackage{overpic}
\usepackage{tikz}
\usepackage{rotating}
\usepackage{psfrag}

\usepackage{url}
\def\UrlBreaks{\do\/\do-}

\providecommand{\comment}[1]{}

%%% New version of \caption puts things in smaller type, single-spaced 
%%% and indents them to set them off more from the text.
\makeatletter
\long\def\@makecaption#1#2{
  \vskip 0.8ex
  \setbox\@tempboxa\hbox{\small {\bf #1:} #2}
  \parindent 1.5em  %% How can we use the global value of this???
  \dimen0=\hsize
  \advance\dimen0 by -3em
  \ifdim \wd\@tempboxa >\dimen0
  \hbox to \hsize{
    \parindent 0em
    \hfil 
    \parbox{\dimen0}{\def\baselinestretch{0.96}\small
      {\bf #1.} #2
      %%\unhbox\@tempboxa
    } 
    \hfil}
  \else \hbox to \hsize{\hfil \box\@tempboxa \hfil}
  \fi
}
\makeatother

\begin{document}
% Control whitespace around equations
\abovedisplayskip=8pt plus0pt minus3pt
\belowdisplayskip=8pt plus0pt minus3pt

% ------------------------------------------------------------------------
% Default title and authorship
% ------------------------------------------------------------------------

\begin{center}
  {\LARGE Assessing External Validity Over  Worst-case Subpopulations\footnote{An extended abstract for an earlier version of this work appeared at
  the Conference in Learning Theory 2020 entitled ``Robust Causal Inference
  Under Covariate Shift via Worst-Case Subpopulation Treatment Effects''.}} \\
  \vspace{.5cm}
  % {\Large Authors blinded for review } \\
  {\Large Sookyo Jeong$^{1}$ ~~~~  Hongseok Namkoong$^{2}$ } \\
  \vspace{.2cm} {\large
    $^1$Lyft Inc. \\
    \vspace{.1cm}   $^2$Decision, Risk, and Operations Division, Columbia Business School
    \\ \vspace{.1cm} }
  \vspace{.2cm}
  {\tt sjeong@lyft.com, namkoong@gsb.columbia.edu}
\end{center}

% ------------------------------------------------------------------------
% Abstract
% ------------------------------------------------------------------------

\begin{abstract}
  Study populations are typically sampled from limited points in space and time,
and marginalized groups are underrepresented. To assess the external validity
of randomized and observational studies, we propose and evaluate the
worst-case treatment effect (WTE) across all subpopulations of a given size,
which guarantees positive findings remain valid over subpopulations. We
develop a semiparametrically efficient estimator for the WTE that analyzes the
external validity of the augmented inverse propensity weighted estimator for
the average treatment effect.  Our cross-fitting procedure leverages flexible
nonparametric and machine learning-based estimates of nuisance parameters and
is a regular root-$n$ estimator even when nuisance estimates converge more
slowly. On real examples where external validity is of core concern, our
proposed framework guards against brittle findings that are invalidated by
unanticipated population shifts.

%%% local Variables:
%%% mode: latex
%%% TeX-master: "main"
%%% End:

\end{abstract}

\fi

% ------------------------------------------------------------------------
% Main Paper Body
% ------------------------------------------------------------------------

\newcommand{\makesmall}[1]{\ifdefined\useectastyle
  {\small
    #1
  }
  \else
  {#1}
  \fi
  }

\definecolor{innerboxcolor}{rgb}{.9,.95,1}
\definecolor{outerlinecolor}{rgb}{.6,0,.2}

\newcommand{\hn}[1]{\fcolorbox{outerlinecolor}{innerboxcolor}{
    \begin{minipage}{.9\textwidth}
      \red{\bf [ HN Comment:} {#1}
      \textbf{\red ]}
  \end{minipage}} \\
}

\newcommand{\sj}[1]{\fcolorbox{outerlinecolor}{innerboxcolor}{
    \begin{minipage}{.9\textwidth}
      \red{\bf [ SJ Comment:} {#1}
      \textbf{\blue ]}
  \end{minipage}} \\
}

\newcommand{\hnshort}[1]{
  \red{\bf [ HN: {#1} ]}
}

\newcommand{\sjshort}[1]{
  \blue{\bf [ SJ: {#1} ]}
}

\providecommand{\comment}[1]{}

\newcommand{\rem}{\mathfrak{R}_k}
\newcommand{\hbound}{M_h}

% population quantities: p for population
\newcommand{\pprop}{e\opt}
\newcommand{\pmu}{\mu\opt}
\newcommand{\pmzero}{\mu_0\opt}
\newcommand{\pmone}{\mu_1\opt}
\newcommand{\pthr}{h\opt} % threshold indicator
\newcommand{\pall}{\pmzero, \pmone, \pprop, \pthr}

% estimates for each bin: h for hat
\newcommand{\hprop}{\what{e}_{\indfold}}
\newcommand{\hmu}{\what{\mu}_{\indfold}}
\newcommand{\hmzero}{\what{\mu}_{0, \indfold}}
\newcommand{\hmone}{\what{\mu}_{1, \indfold}}
\newcommand{\hthr}{\what{h}_{\indfold}}
\newcommand{\hall}{\hmzero, \hmone, \hprop, \hthr}

\newcommand{\rprop}{\what{e}_{\indfold, r}}
\newcommand{\rmu}{\what{\mu}_{\indfold, r}}
\newcommand{\rmzero}{\what{\mu}_{0, \indfold, r}}
\newcommand{\rmone}{\what{\mu}_{1, \indfold, r}}
\newcommand{\rthr}{\what{h}_{\indfold, r}}
\newcommand{\rall}{\rmzero, \rmone, \rprop, \rthr}

\newcommand{\rtall}{\what{\mu}_{0, \indfold, r + t}, \what{\mu}_{1, \indfold, r}, \what{e}_{\indfold, r+t}, \what{h}_{\indfold, r + t}}

\newcommand{\rrprop}{\what{e}_{\indfold, r+t}}
\newcommand{\rrmu}{\what{\mu}_{\indfold, r+t}}
\newcommand{\rrmzero}{\what{\mu}_{0, \indfold, r+t}}
\newcommand{\rrmone}{\what{\mu}_{1, \indfold, r+t}}
\newcommand{\rrthr}{\what{h}_{\indfold, r+t}}
\newcommand{\rrall}{\rmzero, \rmone, \rprop, \rthr}

\newcommand{\empfold}{\what{P}_k}

\newcommand{\hq}{\what{q}_{\indfold}}
\newcommand{\quantile}[2]{P_{#1}^{-1}(#2)}
\newcommand{\aq}[1]{\quantile{1-\alpha}{#1}}

\newcommand{\fold}{I_k}
\newcommand{\cfold}{I_{k}^c}
\newcommand{\cfoldinf}{I_{k}^{c, \infty}}

\newcommand{\env}{\bar{\mu}}

\newcommand{\wte}[1]{\mbox{\rm WTE}_{#1}}
\newcommand{\rvs}{W}
\newcommand{\indfold}{k}
\newcommand{\empcrossfit}{\what{P}_{n, \indfold}}
\newcommand{\augment}{\kappa}
\newcommand{\indica}[1]{\mbf{1}_a\left\{#1\right\}} % Indicator function
\newcommand{\indican}[1]{\mbf{1}_{a_n}\left\{#1\right\}} % Indicator function

\section{Introduction}
\label{section:introduction}

% Outline:
% 1. External validity breaks when population not representative. Examples. 
% 2. Particularly problematic when there is heterogeneity.
% 3. OHIE example.
% 4. Framework. Really explain our setting.
%
%
When the study population is different from those affected by the treatment,
the external validity of a study's finding may be called into question. Study
populations are often sampled from a particular set of points in space and
time and may not represent future populations of interest~\citep{CampbellSt63,
  Manski13, RosenzweigUd20, DehejiaPoSa21}. Furthermore, study populations
often lack diversity, and minority groups are underrepresented. For example,
out of $10,000+$ cancer clinical trials funded by the National Cancer
Institute, less than 5\% of participants were non-white~\citep{ChenLaDaPaKe14,
  ShenoyHa15}.
% less than 2\% focused on racial minorities, and
When the treatment effect is heterogeneous, both randomized and observational
studies lose external validity outside the study population. While large-scale
randomized trials offer a ``gold standard'' for internal validity, their
external validity can be nevertheless called into question over spatiotemporal
changes in the population~\citep{Deaton10, BasuSuHa17}.

% ACCORD and SPRINT trials studied the effect of treatments to lower blood
% pressure on cardiovascular disease but reached opposite conclusions despite
% large sample sizes ($n = 4733$ for ACCORD and $n = 9361$ for SPRINT). The
% mechanism behind the difference could not be explained by experts even
% ex-post~\citep{Accord10, Sprint15, BasuSuHa17}.

\begin{figure}[t]
  \centering
  \begin{tabular}{cc}
      \includegraphics[scale=.3]{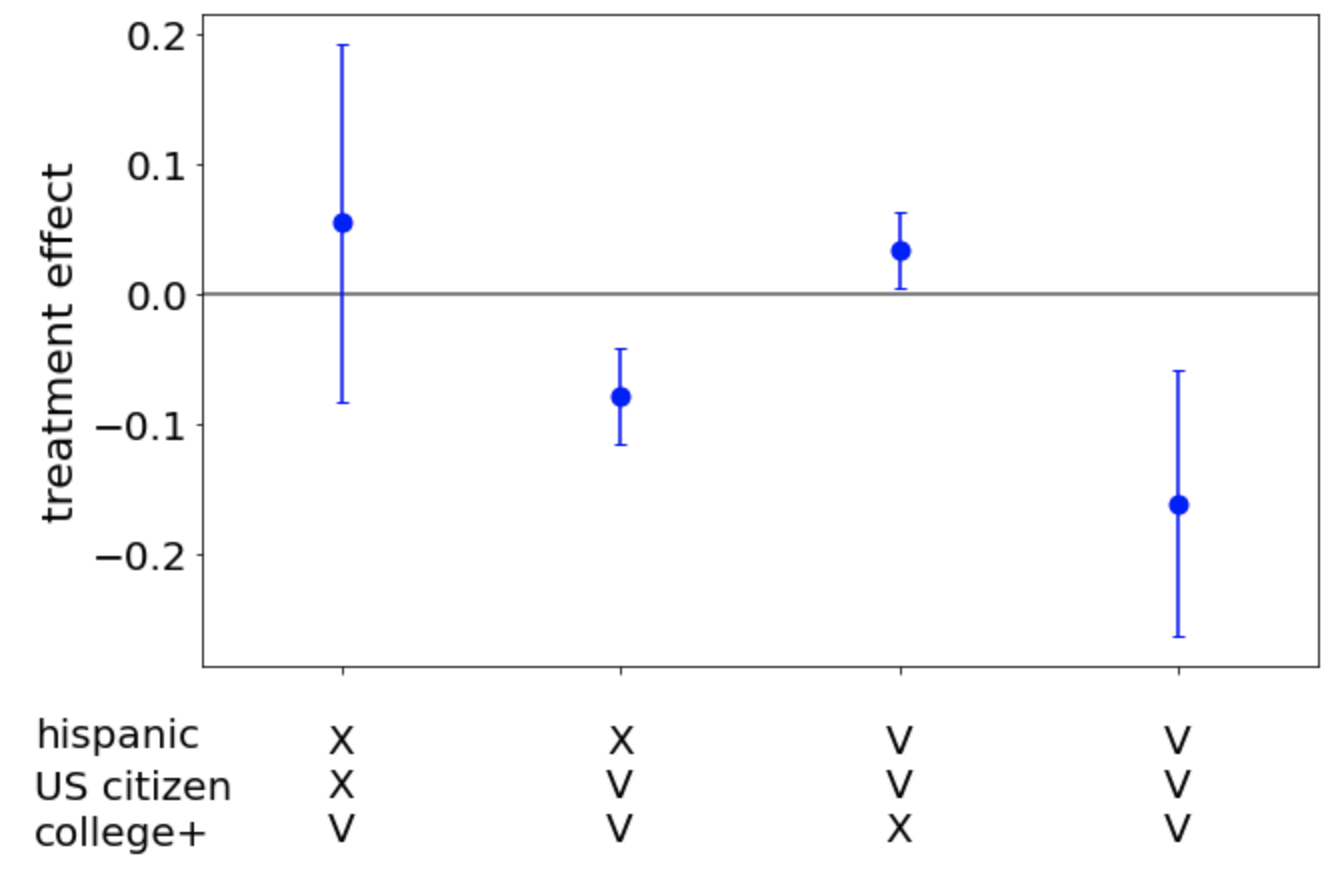}
      &
          \includegraphics[scale=.3]{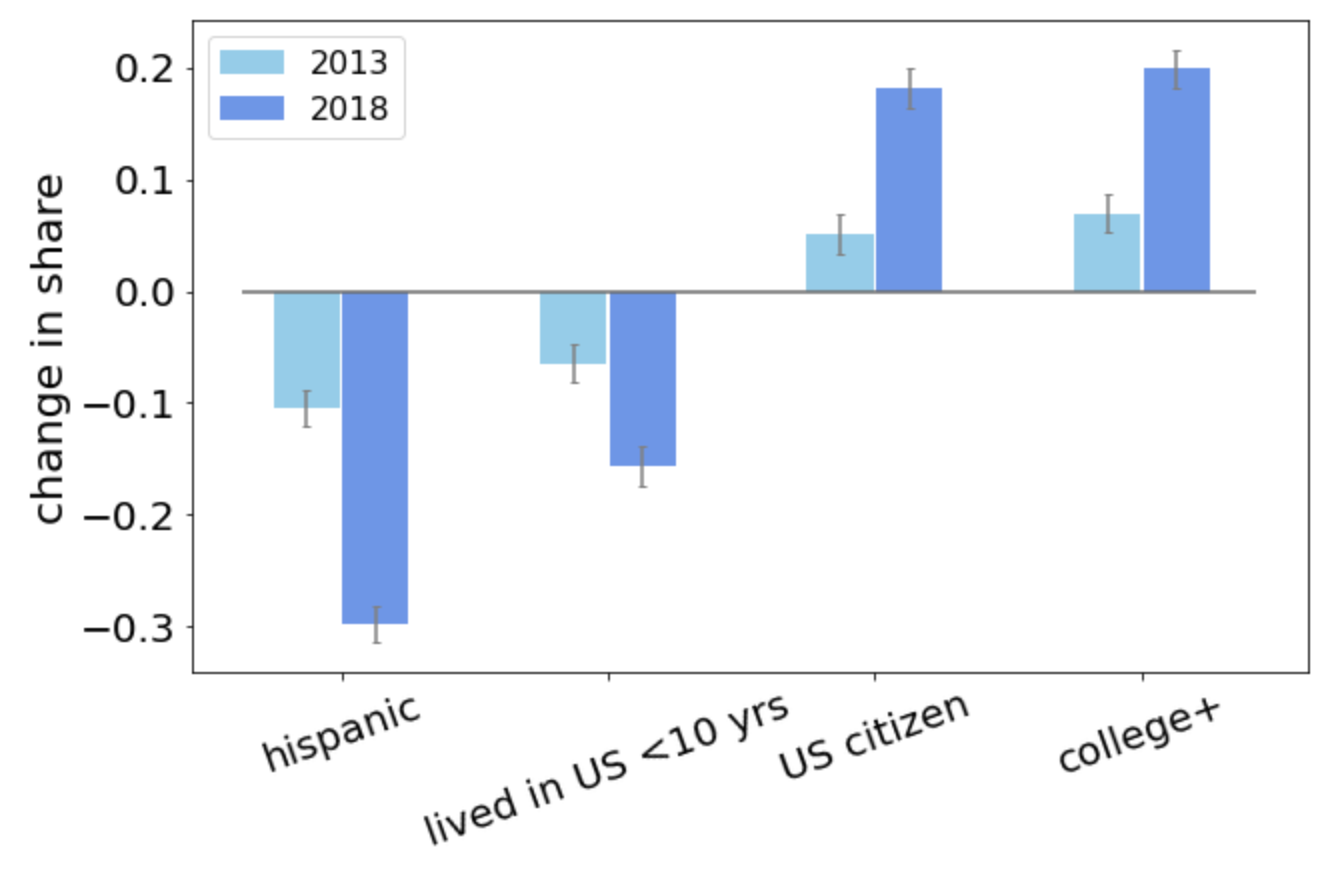}
    \\
    \small{(a) Intersectionality of treatment effects (2009)}

      & \small{(b) Change in shares from 2009 to 2013 \& 2018 } 
  \end{tabular}
  \vspace{10pt}
  \caption{   
    \label{fig:intro} Effect of Medicaid enrollment on doctors' visit for
    low-income adults (NHIS) } \end{figure}

Existing approaches for assessing external validity require the knowledge of
the target population~\citep{HotzImMo05, AngristFe10, ColeSt10,
  StuartCoBrLe11, Tipton13, LeskoBuWeEdHuCo17, AndrewsOs17,
  Meager19}. However, target populations chosen at the time of analysis may
not be sufficient to guarantee external validity over unforeseen shifts in the
population that occur post-analysis.  Heuristic approaches such as estimating
treatment effects over fixed subgroups are similarly limited as effects
typically vary over a combination of multiple characteristics like race,
gender, age, and income, a phenomenon we refer to as \emph{intersectionality}.

To illustrate these challenges, consider estimating the effect of Medicaid
enrollment on doctors' office utilization based on the National Health
Interview Survey (NHIS) in 2009, where we focus on whether the individual made
any visits to doctors two weeks prior to the survey date as the main
outcome. Observed treatment effects and resulting decisions in 2009 must
remain valid over \emph{a priori unknown shifts} in the population over time.
We observe substantial intersectionality in the treatment effect
(Figure~\ref{fig:intro}a): for Hispanic U.S.  citizens, the treatment effect
changes signs depending on professional degree attainment.  In the decade
following 2009 (``future''), there is a major shift in the underlying
population (Figure~\ref{fig:intro}b), and observed effects in 2009 are no
longer valid in the future, as we later illustrate in
Section~\ref{section:experiments}.

% We use the potential outcomes notation to denote counterfactual
% outcomes. Focusing on binary treatments for simplicity, let $Y(1)$ and $Y(0)$
% be (real-valued) outcomes corresponding to treatment and control
% respectively~\citep{Rubin74, Holland86, Rubin05}. To evaluate the causal
% effect of a treatment, a standard goal is to estimate the \emph{average
%   treatment effect (ATE)}
% \begin{equation}
%   \label{eqn:ate}
%   \mbox{ATE} \defeq \E[Y(1) - Y(0)] = \E_{X \sim P_X}\left[\pmu(X)\right]
%   ~~\mbox{where}~~\pmu(X) \defeq \E[Y(1) - Y(0)|X].
% \end{equation}
% The ATE is an average of $\pmu(X)$, the \emph{conditional average treatment
%   effect (CATE)}, under the data-generating distribution
% $X \sim P_X$. Letting $Z \in \{0, 1\}$ be the assigned treatment, we only
% observe the outcome $Y \defeq Y(Z)$ associated with the assigned treatment,
% with corresponding pretreatment covariates $X \in \R^d$. In particular, the
% counterfactual outcome $Y(1-Z)$ is always missing.

To assess external validity over unanticipated population shifts in the
population, we propose and study the worst-case treatment effect,
$\mbox{WTE}_{\alpha}$, defined over \emph{all subpopulations} that comprise at
least $\alpha$-fraction of the study population (see Eq.~\eqref{eqn:wte} to
come).  As the $\mbox{WTE}_{\alpha}$ bounds the average treatment effect (ATE)
and reduces to it when $\alpha=100\%$, the WTE analyzes the sensitivity of an
average-case finding under population shifts, guaranteeing that conservative
findings using the WTE remain valid uniformly over subpopulations. For
example, if low-income Hispanic U.S. citizens with professional degrees
comprise at least $20\%$ of the study population,  positive findings with
respect to $\mbox{\rm WTE}_{.2}$ guarantee the treatment remains effective
over this subgroup.

% Letting $Z \in \{0, 1\}$ be the assigned treatment, we only observe the
% outcome $Y \defeq Y(Z)$ associated with the assigned treatment, with
% corresponding pretreatment covariates $X \in \R^d$. In particular, the
% counterfactual outcome $Y(1-Z)$ is always missing.

% We propose a $K$-fold cross-fitting procedure to estimate
% $\mbox{\rm WTE}_{\alpha}$ (Section~\ref{section:approach}), where we split the
% data into auxiliary ($K-1$ folds) and main samples ($1$ fold). On the
% auxiliary sample, we fit ML models of CATE and use it to estimate the
% threshold that defines the worst-case subpopulation; on the main sample, we
% combine estimates of these nuisance parameters to evaluate the treatment
% effect on the worst-case subpopulation. By switching the role of auxiliary and
% main samples, we utilize all samples, similar to the cross-fitting procedure
% for the ATE proposed by~\citet{ChernozhukovChDeDuHaNeRo18}.
% By switching the role of auxiliary and main samples across the $K$ folds, our
% procedure is able to utilize all samples, and allows using large scale machine
% learning models to predict infinite-dimensional nuisance parameters. As we
% show in the sequel, our sample splitting approach removes bias induced by both
% regularization and overfitting, similar to that for the
% ATE~\citep{ChernozhukovChDeDuHaNeRo18}.

We develop a semiparametrically efficient estimator of
$\mbox{\rm WTE}_{\alpha}$, analyzing the external validity of the augmented
inverse propensity weighted (AIPW) estimator~\citep{RobinsRoZh94, RobinsRo95}
for the ATE. Our $K$-fold cross-fitting procedure leverages machine
learning-based and nonparametric estimators of nuisance parameters and
provides a worst-case bound on the cross-fitted AIPW for the
ATE~\citep{ChernozhukovChDeDuHaNeRo18}; our estimator reduces to the AIPW when
$\alpha = 100\%$. On real datasets where external validity is of core concern,
our worst-case sensitivity approach identifies disadvantaged subpopulations
based on a priori nontrivial demographic groupings and guards against brittle
findings that are invalidated under population shifts
(Section~\ref{section:experiments}).

Specifically, we exploit the dual representation of the worst-case over
subpopulations to derive our semiparametric estimator
(Section~\ref{section:approach}).  By virtue of satisfying an orthogonality
property (similar to the AIPW for the ATE), under standard assumptions
required for the identification and estimation of the ATE, our augmented
estimator of $\mbox{WTE}_{\alpha}$ enjoys central limit rates even when
estimates of the nuisance parameters converge at slower-than-parameteric rates
(Section~\ref{section:asymptotics}).  Since $\mbox{WTE}_{\alpha}$ is nonlinear
in the underlying probability measure, our main asymptotic result
(Theorem~\ref{theorem:clt}) requires a novel theoretical analysis different from
the estimating equations framework (method of moments) studied
by~\citet{ChernozhukovChDeDuHaNeRo18}.

We prove that our augmented estimator for the $\mbox{WTE}_{\alpha}$ is
semiparametrically efficient, in both observational and randomized studies
(Section~\ref{section:efficiency}). Our semiparametric efficiency bound
informs experimental design with external validity as a central
concern. % Just as standard statistical power
% calculations for the ATE give the minimal number of study participants
% required to detect a specified effect size,
Power calculations based on our efficiency bound provide the minimal sample
size required to detect a specified effect size for the
$\mbox{WTE}_{\alpha}$. Our bounds quantify how testing external validity
against smaller subpopulations ($\alpha$) requires a correspondingly larger
sample sizes.

\paragraph*{Related work}

Assessing the external validity of randomized and observational studies is an
active area of research in causal inference~\citep{DahabrehHe19}.
% the set of efforts that aim to extend inferences from an experiment to a
% target population is referred to as generalizability and transportability. We
% refer readers to~\citet{DahabrehHe19, BareinboimPe16} and references therein
% for a detailed discussion of recent results. 
% ~\citet{DahabrehHe19} defines \emph{generalizability} as the ``extension of
% inferences from the trial to a target population that coincides, or is a
% subset of, the trial-eligible population'', and \emph{transportability} as the
% ``extension of inferences from the trial to a target population that includes
% individuals who are not part of the trial-eligible population''. For
% \emph{generalizability}, 
When the target population is \emph{known} but different from the study
population, many authors have leveraged the relationship between the two
populations to guarantee external validity. A prevalent approach is to view
selection into the study as another ``treatment'', adjusting estimates of the
ATE based on some information about the target population~\citep{ColeSt10,
  StuartCoBrLe11, Tipton13, KernStHiGr16, LeskoBuWeEdHuCo17, AndrewsOs17,
  Meager19}.~\citet{StuartCoBrLe11} and~\citet{Tipton13} use the probability
of being included in the study to adjust for population bias, assuming that
sample selection decisions only depend on observed covariates.
\citet{HotzImMo05} apply bias-corrected matching methods to predict the impact
of a program by using observations collected from a different location. In the
context of structural causal models, Bareinboim, Pearl and colleagues identify
settings that allow external validity in a series of
works~\citep{BareinboimPe12, BareinboimPe16}.

External validity is of particular concern when identification strategies only
allow studying a local notion of treatment effect. In such scenarios, several
authors aim to connect estimates for a local population to a (known) broader
population by leveraging a postulated structure between the two
populations. For instrumental variable strategies, there is a line of work
(see, for example,~\citep{AngristImRu96, Angrist04, AngristFe10}) studying
when the treatment effects for compliers (LATE) can inform effects for a
broader population. Most recently,~\citet{RosenzweigUd20} directly estimate
external validity over time when exogenous aggregate shocks are observed. For
regression discontinuity designs,~\citet{DongLe15, AngristRo15, BertanhaIm20}
analyze settings where local estimates are externally valid and develop
corresponding statistical tests.

% ~\citet{StuartCoBrLe11, Tipton14} propose using propensity scores of being
% included in the study to quantify the extent to which results of an
% experiment can generalize to the overall population.~\citet{ColeSt10,
% StuartCoBrLe11, KernStHiGr16, LeskoBuWeEdHuCo17} uses propensity-based or
% matching-based methods is used to correct for the selection bias into the
% study.

% Another missing BoGa, and RES20

% \citep{BareinboimPe12, BareinboimPe13, BareinboimPe16, BareinboimTiPe14,
% PearlBa14}.  Another related literature is domain adaptation, where a
% prediction model is learned on a supervised data from one domain with the
% purpose of being used on a pre-specified target domain. These methods
% reweight the distribution $P$ to adjust for distributional differences
% between the source and target domains~\citep{Shimodaira00,
% HuangGrBoKaScSm07, BickelBrSc07, SugiyamaKrMu07, SugiyamaNaKaBuKa08,
% TsuboiKaHiBiSu09}. In statistics, mixture model approaches are frequently
% used model latent subpopulations directly~\citep{AitkinRu85, FigueiredoJa02,
% McLachlanPe04, CappeMoRy05}.  When subpopulations of interest are clearly
% defined and membership is observed, Meinshausen, Buhlmann, and colleagues
% study models that achieve good prediction performance on all (known)
% subpopulations~\citep{MeinshausenBu15, RothenhauslerMeBu16, BuhlmannMe16,
% RothenhauslerBuMePe18}.

Compared to the above methods, our worst-case sensitivity approach does not
assume knowledge of the target population. Our conservative approach is
agnostic to the unknown shifts in the population and provides uniform
guarantees over subpopulations comprising at least $\alpha$-fraction of the
study population. This is conceptually related to recent works on
distributionally robust optimization in operations research and supervised
learning, where models are trained to optimize a worst-case loss over
distribution shifts~\citep{DuchiHaNa20}. Relatedly,~\citet{BoGa21} studied
external validity as a notion of stability of the joint distribution between
the observed outcome and treatment assignments.

Study populations must be designed to be as diverse as possible across
demographics, space, and time.  Our approach can guarantee meaningful external
validity only if the study includes heterogeneous subpopulations. (When the
study population is not representative, our approach can nevertheless raise
alarms.)  A \emph{design-based} approach to external validity complements our
sensitivity framework by promoting diversity in the study population as a
central concern. Several works in development and labor economics aim to
improve the external validity of a (quasi-) experiment by collecting data over
multiple sites and temporal points~\citep{CrucesGa07, BanerjeeKaZi15,
  GertlerShAlCaMaPa15, DupasKaRoUb18, RosenzweigUd20, DehejiaPoSa21}.  Tipton
and colleagues develop methodologies for measuring the diversity of a study
population alongside practical experimental design guidelines~\citep{Tipton14,
  TiptonPe17, TiptonRo18}.
% Collecting data over multiple locations and time can also improve the
% external validity of quasi-experiments~\citep{RosenzweigUd20,DehejiaPoSa21}.

Our worst-case approach is broadly related to previous works that estimate
treatment effects beyond mean differences~\citep{Rothe10,
  KimKiKe18}.~\citet{ChernozhukovFeLu18} study \emph{sorted effects}, a
collection of sorted quantiles of the conditional average treatment effect
(CATE). They develop central limit results for this nonparametric estimand
under Donsker conditions (i.e., functional CLT) on CATE estimates. The
$\mbox{WTE}_{\alpha}$ we introduce in Section~\ref{section:approach} is a
tail-average of sorted effects, and our semiparametric approach extends the
AIPW under unanticipated shifts in the population.  Theoretically, we prove
central limit results for our estimator \emph{without requiring Donsker
  conditions on CATE estimates}. Our approach is not to be confused with
quantile treatment effects~\citep{Firpo07}, which measures the difference
between quantiles of $Y(1)$ and $Y(0)$.

\section{Approach}
\label{section:approach}

Using the potential outcomes notation to denote counterfactuals, we let $Y(1)$
and $Y(0)$ be outcomes corresponding to treatment and control and let
$Z \in \{0, 1\}$ be the assigned treatment~\citep{Rubin74}.  We study both
randomized and observational studies, assuming the analyst has access to
observed covariates $X \sim P_X$. A standard goal is to estimate the
\emph{average treatment effect},
$\mbox{ATE} \defeq \E[Y(1) - Y(0)] = \E_{X \sim P_X}\left[\pmu(X)\right]$,
where $\pmu(X) \defeq \E[Y(1) - Y(0)|X]$ is the \emph{conditional average
  treatment effect (CATE)}.  Throughout, we assume the distribution
$Y(1), Y(0) \mid X$ remains unchanged over subpopulations.

As the study population $P_X$ may not be representative of those affected by
the treatment, we are interested in measuring the sensitivity of a study's
finding to shifts in the underlying population. We consider the set of all
subpopulations (probabilities) $Q_X$ that comprise more than
$\alpha \in (0, 1]$ fraction of the study population $P_X$
\begin{equation}
  \label{eqn:subpopulations}
  \mc{Q}_\alpha \defeq \left\{Q_X \mid
    P_X = a Q_X + (1-a) Q_X'~\mbox{for some}~a \ge \alpha,~\mbox{and subpopulation}~Q_X'
  \right\}.
\end{equation}
As a convention, we assume the desired sign of the treatment effect is
negative (the positive case is symmetric).  We propose and study the worst-case
subpopulation treatment effect
\begin{equation}
  \label{eqn:wte}
  \mbox{\rm WTE}_{\alpha} \defeq \sup_{Q_X \in \mc{Q}_{\alpha}} \E_{X \sim
    Q_X}\E[Y(1) - Y(0)| X].
\end{equation}
When treatment effects are highly heterogeneous and external validity is of
particular concern, the worst-case subpopulation treatment
effect~\eqref{eqn:wte} will be substantially different from the ATE. The
worst-case bound $\mbox{WTE}_{\alpha}$ reduces to the ATE when
$\alpha = 100\%$.

 % Compared to the
% ATE---corresponding to $\mbox{WTE}_{\alpha}$ with $\alpha=100\%$---which
% solely relies on the study population $P_X$, the WTE~\eqref{eqn:wte} analyzes
% the sensitivity of a study's findings to unanticipated shifts in
% population. Conservative findings with respect to the WTE guarantees uniformly
% valid treatment effects across all subpopulations larger than $\alpha$.  For
% example, if low-income Black females without insurance \emph{and} diabetic
% history represent at least $20\%$ of the collected data, then
% $\mbox{\rm WTE}_{.2} < 0$ guarantees the treatment remains effective over the
% subgroup.

The modeling choice of $\alpha$ in the definition~\eqref{eqn:wte} is
important, and it should be informed by domain knowledge. When a study
population is not representative, we recommend selecting a smaller value of
$\alpha$. For example, the analyst may reason about the level of bias
anticipated in the data collection process or use the size of proxy target
groups in the study.  In the latter case, positive findings with the chosen
level of $\alpha$ guarantee uniformly valid treatment effects over all
minority subpopulations of size $\alpha$, not just the proxy targets.  The
choice of $\alpha$ should also consider the data size: as $\alpha$ becomes
small, inference becomes difficult, as our semiparametric efficiency bounds
demonstrate in Section~\ref{section:efficiency}. Even when the analyst does
not commit to a single level of $\alpha$, evaluating the WTE over a range of
$\alpha$'s can offer a practical diagnostic. The level of $\alpha$ at which
$\mbox{WTE}_{\alpha}$ crosses a threshold (e.g., 0) is often of particular
practical interest as it represents the smallest subpopulation size over which
average-case findings remain valid.

To derive our augmented estimator, we begin by simplifying the primal
problem~\eqref{eqn:wte} over (infinite-dimensional) covariate distributions
$Q_X \in \mc{Q}_{\alpha}$ to its dual representation over a one-dimensional
threshold on the CATE $\pmu(X) = \E[Y(1) - Y(0) \mid X]$. The dual
reformulation shows an equivalence between worst-case subpopulation
performances and tail-averages. We rely on this relationship heavily to derive
our augmented estimator and to prove its asymptotic properties.
% Using the likelihood ratio $L(X) \defeq \frac{dQ_X}{dP_X}$, we can rewrite
% $\mbox{WTE}_{\alpha}$~\eqref{eqn:wte} as
% \begin{align*}
%   \mbox{\rm WTE}_{\alpha}
%   = \sup_{L: \mc{X} \to [0, \alpha^{-1}]~\mbox{\scriptsize meas.}~}
%   \left\{ \E_{X \sim P_X}[L(X) \pmu(X)]:
%   \E_{X \sim P_X}[L(X)] = 1
%   \right\}.
% \end{align*}
We make the dependence on the underlying probability explicit and write
$\E_Q[X]$, except for when $Q = P$, the data-generating distribution.  The
following lemma is a consequence of~\citet[Example 6.19]{ShapiroDeRu09}.
\begin{lemma}
  \label{lemma:dual}
  Let $\aq{\pmu}$ be the $(1-\alpha)$-quantile of $\pmu(X)$, and denote
  $\hinge{\cdot} \defeq \max(\cdot, 0)$ and
  $\pthr(x) \defeq \frac{1}{\alpha} \indic{\pmu(x) \ge \aq{\pmu}}$. If
  $\E[\pmu(X)_+] < \infty$, then
  {\small
  \begin{align*}
    \mbox{WTE}_{\alpha}
    = \inf_{\eta \in \R} \left\{
    \frac{1}{\alpha} \E\hinge{\pmu(X) - \eta} + \eta
      \right\} 
    = \E \left[ \pmu(X) \mid \pmu(X) \ge
    \aq{\pmu} \right] = \E[\pmu(X) h\opt(X)].
  \end{align*}
  }
\end{lemma}

The dual optimum is attained at $\aq{\pmu}$ giving the second equality. This
tail-average is known as the conditional value-at-risk (CVaR), a common risk
measure in portfolio optimization~\citep{RockafellarUr00}. In contrast to the
$\mbox{WTE}_{\alpha}$ involving an unknown nuisance parameter $\pmu(X)$ that
needs to be estimated, the CVaR is typically considered over an observable
random variable---this gives rise to a salient semiparametric structure. The
dual shows the worst-off subpopulation is given by those who get
disproportionately and adversely affected by the treatment, measured by $X$
such that $\pmu(X) \ge \aq{\pmu}$.  To illustrate how the WTE~\eqref{eqn:wte}
accounts for heterogeneity across subpopulations, consider
$\pmu(X) \sim N(-.1, 1)$ so that there is substantial heterogeneity across
covariates. Although the $\mbox{ATE} = -.1$ suggests a negative treatment
effect, % elementary calculations show
% $\mbox{WTE}_{.9} = 0.095 \approx - \mbox{ATE}$; for $90\%$ of the
% population, the treatment effect is the opposite to the ATE. The contrast
% becomes stark as $\alpha$ decreases:
$\mbox{WTE}_{.9} = 0.184$, meaning the treatment effect goes in the reverse
direction of the ATE for even $90\%$ of the study population.

% Talk about identification We use dual to derive estimator.  state basic
% conditions
To identify causal effects, we assume no unobserved confounding and overlap
between the treated and control groups.
\begin{assumption}
  \label{assumption:ig}
  Ignorability~ $Y(0), Y(1) \indep ~Z \mid X$
\end{assumption}
\begin{assumption}
  \label{assumption:overlap}
  Overlap: There exists $c> 0$ such that  $\P(\pprop(X) \in [c, 1-c]) = 1$.
\end{assumption}
\noindent We also assume that units do not interact with each other (no
interference), and that we observe i.i.d. units $D_i = (X_i, Y_i, Z_i)$ for
$i = 1, \ldots, n$ (stable unit treatment value
assumption~\citep{Rubin80}). Finally, we require the following standard
condition that uniformly bounds the conditional variance of the residuals for
$z \in \{0, 1\}$.
\begin{assumption}
  \label{assumption:residuals}
  Bounded residuals
  {\small $\E[Y(z)^2] + \linfstatnorm{\E[ (Y(z) - \mu_z\opt(X))^2 \mid X]} < \infty$} 
\end{assumption}
\noindent These standard assumptions are also required to identify and
estimate the ATE~\citep{ImbensRu15, ChernozhukovChDeDuHaNeRo18}.

Recalling that $\pthr$ is a nuisance parameter determining the worst-off
subpopulation in Lemma~\ref{lemma:dual}, we consider the following key nuisance parameters
\begin{equation}
  \label{eqn:nuisance-def}
  \begin{split}
  \mbox{outcome models}~& ~\mu\opt_z(x) \defeq \E[Y(0) \mid X =x, Z=z]~~\mbox{for}~~z \in \{0,1\},\\
  \mbox{propensity score}~& ~\pprop(x) \defeq \P(Z = 1 \mid X = x) \\
  \mbox{threshold function}~& ~\pthr(x) \defeq \frac{1}{\alpha} \indic{\pmu(x) \ge \aq{\pmu}}.
  \end{split}
  \end{equation}
Letting $D = (X, Y, Z)$ be the tuple of observed data and
$(\mu_0, \mu_1, e, h)$ be the tuple of nuisance parameters, we consider the
augmentation term
\begin{equation}
  \label{eqn:aug}
  \kappa(D; (\mu_0, \mu_1, e, h)) \defeq
  h(X) \left(
    \frac{Z}{e(X)} (Y - \mu_1(X)) - \frac{1-Z}{1-e(X)} (Y - \mu_0(X))
  \right).
\end{equation}
Under the stated assumptions, we have $\E[\kappa(D; \pall)] = 0$. Instead of
estimating $\mbox{WTE}_{\alpha}$, we estimate the augmented form
$\mbox{WTE}_{\alpha} + \E[\kappa(D; \pall)]$. When $\alpha = 1$ so that
$\mbox{WTE}_{1} = \mbox{ATE}$, our estimator reduces to the augmented inverse
probability weighted (AIPW) estimator for the ATE. Thus, our estimator can be
viewed as an extension of the AIPW estimator under shifts in the underlying
population.

\ifdefined\usemsstyle
\begin{algorithm}[t]
  \caption{\label{alg:cross-fitting} Cross-fitting for $\mbox{WTE}_{\alpha}$}
  \begin{algorithmic}[1]
    \STATE \textsc{Input: $K$-fold partition $\cup_{k=1}^K I_k = [n]$ of
     $\{(X_i, Y_i, Z_i)\}_{i=1}^n$ s.t. $|I_k| = \frac{n}{K}$} 
   \STATE \textbf{\textsc{For}} $\indfold \in [K]$
   \STATE \hspace{10pt} \textbf{Estimate nuisance parameters}
   Using the data $\{D_i\}_{i \in \cfold}$, fit estimators
   \STATE \hspace{10pt} 1. $\what{\mu}_{z, \indfold}(\cdot)$ of
      $\mu_z\opt(\cdot) = \E[Y(z) \mid X = \cdot, Z = z]$ for $z \in \{0, 1\}$ % (e.g. calibrated
      % predictor of $Y(z)$)
      \STATE \hspace{10pt} 2. $\hprop(\cdot)$ of $e\opt(\cdot) = \P(Z = 1 \mid X =\cdot)$
      % (e.g. calibrated binary classifier of $Z = 1$ vs. $Z = 0$)
      \STATE \hspace{10pt} 3. $\hthr(x) \defeq \frac{1}{\alpha} \indic{\hmu(x) \ge \hq}$, where
      $\hq$ is an estimator of $\aq{\pmu}$ 
    \STATE \hspace{10pt} \textbf{Compute augmented estimator} Using the data
    $\{D_i = (X_i, Y_i, Z_i) \}_{i \in I_k}$, compute
    \begin{align*}
      \what{\omega}_{\alpha, \indfold}
      & \defeq \inf_{\eta}
        \left\{ \frac{1}{\alpha} \E_{X \sim \empfold}
        \hinge{\hmu(X) - \eta} + \eta  \right\}
        + \E_{D \sim \empfold}\left[\augment\left(D; \hall\right)\right] \\
      \what{\sigma}^2_{\alpha, k}
      & \defeq \frac{1}{\alpha^2} \var_{X \sim \empfold} \hinge{\hmu(X) - \hq}
        + \var_{D \sim \empfold} \left( \augment\left(D; \hall\right)\right)
    \end{align*}
    \STATE \textbf{\textsc{Return}} Estimator
    $\what{\omega}_{\alpha} = \frac{1}{K} \sum_{k \in [K]}
    \what{\omega}_{\alpha, k}$, and variance estimate
    $\what{\sigma}^2_{\alpha} = \frac{1}{K} \sum_{k \in [K]}
    \what{\sigma}^2_{\alpha, k}$
  \end{algorithmic}
\end{algorithm}
\else
\begin{algorithm}[t]
  \caption{\label{alg:cross-fitting} Cross-fitting for $\mbox{WTE}_{\alpha}$}
  \begin{algorithmic}[]
    \State \textsc{Input: $K$-fold partition $\cup_{k=1}^K I_k = [n]$ of
     $\{(X_i, Y_i, Z_i)\}_{i=1}^n$ s.t. $|I_k| = \frac{n}{K}$} 
   \State \textbf{\textsc{For}} $\indfold \in [K]$
   \State \hspace{10pt} \textbf{Estimate nuisance parameters}
   Using the data $\{D_i\}_{i \in \cfold}$, fit estimators
   \State \hspace{10pt} 1. $\what{\mu}_{z, \indfold}(\cdot)$ of
      $\mu_z\opt(\cdot) = \E[Y(z) \mid X = \cdot, Z = z]$ for $z \in \{0, 1\}$ % (e.g. calibrated
      % predictor of $Y(z)$)
      \State \hspace{10pt} 2. $\hprop(\cdot)$ of $e\opt(\cdot) = \P(Z = 1 \mid X =\cdot)$
      % (e.g. calibrated binary classifier of $Z = 1$ vs. $Z = 0$)
      \State \hspace{10pt} 3. $\hthr(x) \defeq \frac{1}{\alpha} \indic{\hmu(x) \ge \hq}$, where
      $\hq$ is an estimator of $\aq{\pmu}$ 
    \State \hspace{10pt} \textbf{Compute augmented estimator} Using the data
    $\{D_i = (X_i, Y_i, Z_i) \}_{i \in I_k}$, compute
    \begin{align*}
      \what{\omega}_{\alpha, \indfold}
      & \defeq \inf_{\eta}
        \left\{ \frac{1}{\alpha} \E_{X \sim \empfold}
        \hinge{\hmu(X) - \eta} + \eta  \right\}
        + \E_{D \sim \empfold}\left[\augment\left(D; \hall\right)\right] \\
      \what{\sigma}^2_{\alpha, k}
      & \defeq \frac{1}{\alpha^2} \var_{X \sim \empfold} \hinge{\hmu(X) - \hq}
        + \var_{D \sim \empfold} \left( \augment\left(D; \hall\right)\right)
    \end{align*}
    \State \textbf{\textsc{Return}} Estimator
    $\what{\omega}_{\alpha} = \frac{1}{K} \sum_{k \in [K]}
    \what{\omega}_{\alpha, k}$, and variance estimate
    $\what{\sigma}^2_{\alpha} = \frac{1}{K} \sum_{k \in [K]}
    \what{\sigma}^2_{\alpha, k}$
  \end{algorithmic}
\end{algorithm}
\fi

We now formally define our cross-fitted augmented estimator
$\what{\omega}_{\alpha}$ and an estimate $\what{\sigma}^2_{\alpha}$ of its
asymptotic variance
\begin{align}
  \label{eqn:var}
  \sigma^2_{\alpha}
  & \defeq \frac{1}{\alpha^2} \var\left(\hinge{\pmu(X) - \aq{\pmu}} \right)
    + \var\left( \kappa\left(D; \pall\right) \right).
\end{align}
As we show in Section~\ref{section:asymptotics}, these estimates give an
asymptotically exact confidence interval
$\P(\mbox{WTE}_{\alpha} \in [\what{\omega}_{\alpha} \pm z_{\delta}
\what{\sigma}_{\alpha} / \sqrt{n}]) \to 1-\delta$ if we set $z_{\delta}$ to be
the $(1 - \delta/2)$-quantile of a standard normal distribution. The
asymptotic variance~\eqref{eqn:var} is the best attainable in the typical
semiparametric sense as our semiparametric efficiency bounds in
Section~\ref{section:efficiency} show.

% To estimate the nuisance parameters---outcome models
% $\mu\opt_z(x) = \E[Y(0) \mid X =x, Z=z]$ for $z \in \{0,1\}$, propensity score
% $\pprop(x) = \P(Z = 1 \mid X = x)$, and threshold $\pthr(x)$---we split data
% into main and auxiliary samples. 
We fit estimators of nuisance parameters on the auxiliary sample, and combine
them via the augmented dual form~\eqref{eqn:aug} to evaluate the treatment
effect on the worst-off subpopulation. Our approach is agnostic to the
nuisance estimation method, and in particular, allows flexible use of machine
learning models and nonparametric techniques to estimate $\mu\opt_z$ and
$\pprop$. To estimate the threshold function $\pthr(X)$ that determines the
worst-case subpopulation, we first compute an estimator $\what{q}$ of
$\aq{\pmu}$ based on the auxiliary data, and take
$\what{h}(x) \defeq \frac{1}{\alpha} \indic{(\what{\mu}_1 - \what{\mu}_0)(x)
  \ge \what{q}}$.  In some applications, large quantities of \emph{unlabeled}
covariate observations can be cheaply collected even when \emph{labeled}
observations $(X, Y, Z)$ are expensive. Then a particularly nice estimator
$\what{q}$ of $\aq{\pmu}$ can be constructed by evaluating the
$(1-\alpha)$-quantile of the CATE estimator
$\what{\mu} = \what{\mu}_1 - \what{\mu}_0$ on unlabeled observations. With
cheap unlabeled covariates, such an estimator can be made arbitrarily close to
$\aq{\what{\mu}}$, and hence close to $\aq{\pmu}$ if $\what{\mu}$ is
sufficiently close to $\pmu$ as we show in Section~\ref{section:asymptotics}.

% final estimator
% Given estimators of nuisance parameters trained on the auxiliary data, we
% estimate $\mbox{WTE}_{\alpha}$ on the main sample $I$ by taking
% \begin{equation}
%   \label{eqn:estimator}
%   \inf_{\eta} \left\{\frac{1}{\alpha |I|} \sum_{i \in I} \hinge{\what{\mu}(X_i) - \eta} + \eta
%   \right\}
%   + \frac{1}{|I|} \sum_{i \in I} \kappa\left(D_i; \hall\right).
% \end{equation}

To utilize the entire sample, we take a cross-fitting approach, partitioning
the data into $K$ folds and switching the roles of the main and auxiliary
datasets on each fold. We adapt the original cross-fitting algorithm for
estimating equations (due to~\citet{ChernozhukovChDeDuHaNeRo18}) to estimating
the WTE.  Denoting the $k$-th fold $I_k$ and its complement
$\cfold = [n] \backslash I_k$, we fit nuisance parameters $(\hall)$ on the
$k$-th auxiliary data $\{D_i\}_{i \in \cfold}$.
% where $\what{\mu}_{z, k}(\cdot)$ is an estimator of the outcome model
% $\mu\opt_z(x) = \E[Y(z) \mid X =x, Z=z]$ for $z \in \{0, 1\}$, and
% $\hprop(\cdot)$ is an estimator of the propensity score
% $\pprop(x) = \P(Z=1 \mid X = x)$. Letting
% $\hmu(x) \defeq \hmone(x) - \hmzero(x)$ be the estimator of CATE
% $\pmu(\cdot)$, and $\hq$ be an estimate of $\aq{\hmu}$, our estimate of the
% threshold function $\pthr(x) = \frac{1}{\alpha} \indic{\pmu(x) \ge \aq{\pmu}}$
% is given by $\hthr(x) \defeq \frac{1}{\alpha} \indic{\hmu(x) \ge \hq}$.
Using $\what{P}_k$ to denote the empirical distribution on the $k$-th main
data $\{D_i\}_{i \in I_k}$, we summarize our procedure in
Algorithm~\ref{alg:cross-fitting}.  Our estimator can be computed in both
randomized control trials using the true propensity score or in observational
studies where $\hprop$ needs to be estimated using suitable statistical
models.

We can also derive natural analogues of the direct method (DM) and the inverse
probability weighted estimator (IPW) for estimating $\mbox{WTE}_{\alpha}$
{\small
\begin{subequations}
  \label{eqn:dm-ipw}
\begin{align}
  \what{\mbox{DM}}_{\alpha}
   & \defeq \frac{1}{K} \sum_{k=1}^K \inf_{\eta} \left\{ \frac{1}{\alpha} \E_{\what{P}_k}
    \hinge{\what{\mu}_k(X) - \eta} + \eta \right\}, \\
  ~\what{\mbox{IPW}}_{\alpha, k}
   & \defeq \frac{1}{K} \sum_{k=1}^K  \E_{\what{P}_k} \left[\hthr(X) Y \left(
    \frac{Z}{\hprop(X)}  - \frac{1-Z}{1-\hprop(X)} \right) \right].
\end{align}
\end{subequations}
} Again, the above estimators reduce to their counterparts for estimating the
ATE when $\alpha = 1$. In our subsequent analysis and experiments, we focus on
the augmented estimator presented in Algorithm~\ref{alg:cross-fitting} since
unlike the two approaches above~\eqref{eqn:dm-ipw}, the augmented version
satisfies Neyman orthogonality and achieves the semiparametric efficiency
bound.

\section{Empirical results}
\label{section:experiments}

We empirically demonstrate how our worst-case sensitivity approach guards
against spurious findings that are invalidated under shifts in the underlying
population. Our WTE estimator (Algorithm~\ref{alg:cross-fitting})
automatically detects subgroups adversely affected by the treatment and
guarantees validity of findings over subpopulations on both randomized and
observational studies. We observe that our WTE estimator remains stable even
when typical CATE estimators---based on undersmoothed machine learning
models---vary significantly across estimation methods and sample sizes.

Throughout our empirical analysis, we consider $\mbox{WTE}_{\alpha}$ for
$\alpha \in \{ 1., .8, .6, .4, .2\}$ (recall $\mbox{ATE} = \mbox{WTE}_{1}$),
where without mention we replace the supremum with an infimum in the
definition~\eqref{eqn:wte} if the desired sign of the treatment effect is
positive. We use $K = 3$ in our cross-fitting procedure and use random forests
to estimate outcome models with two-fold cross-validation. Our estimator
bounds the usual cross-fitted AIPW~\citep{ChernozhukovChDeDuHaNeRo18} and
reduces to the AIPW when $\alpha = 1$, which we use as the topline estimator
for the ATE.
\ifdefined\usemsstyle
\else
\footnote{The code for all experiments can be found in 
  \url{https://github.com/sookyojeong/worst-ate}.}
\fi

\ifdefined\useectastyle
\vspace{-15pt}
\fi

\subsection{Effect of Medicaid on doctor visits over time}

Since the passage of the Affordable Care Act, Medicaid has expanded to 38
states in the U.S., aiming to increase healthcare coverage for low-income
individuals.  We study the effectiveness of Medicaid in increasing healthcare
access as measured by the post-enrollment change in doctors' office
utilization.  Taking the viewpoint of an analyst in 2009, we illustrate how
findings in 2009 (``present'') may no longer hold in the subsequent decade
(``future'') due to unforeseen population shifts. Our worst-case sensitivity
approach accounts for latent intersectionality using ``present'' data alone
(Figure~\ref{fig:intro}) and calls into question the external validity of
present-day findings.

%% share
% \begin{figure}[ht]
% \centering
% \begin{tabular}{cc}
%   \includegraphics[scale=.29]{./figures/wte_nhis_alpha.png} 
%   &
%   \includegraphics[scale=.24]{./figures/wte_nhis_breakdown.png}
%   \\
%   \small{(a) $\mbox{WTE}_{\alpha}$ in 2009 ($\mbox{WTE}_{1} = \mbox{ATE}$)}
%   & \small{(b) CATE by covariates in 2009}
% \end{tabular}
% \caption{ \label{fig:negex_share} Effect of Medicaid enrollment on doctors visits for low-income adults in 2009}
% \vspace{-10pt}
% \end{figure}

\begin{figure}[t]
%  \vspace{-10pt}
\centering
\includegraphics[scale=.4]{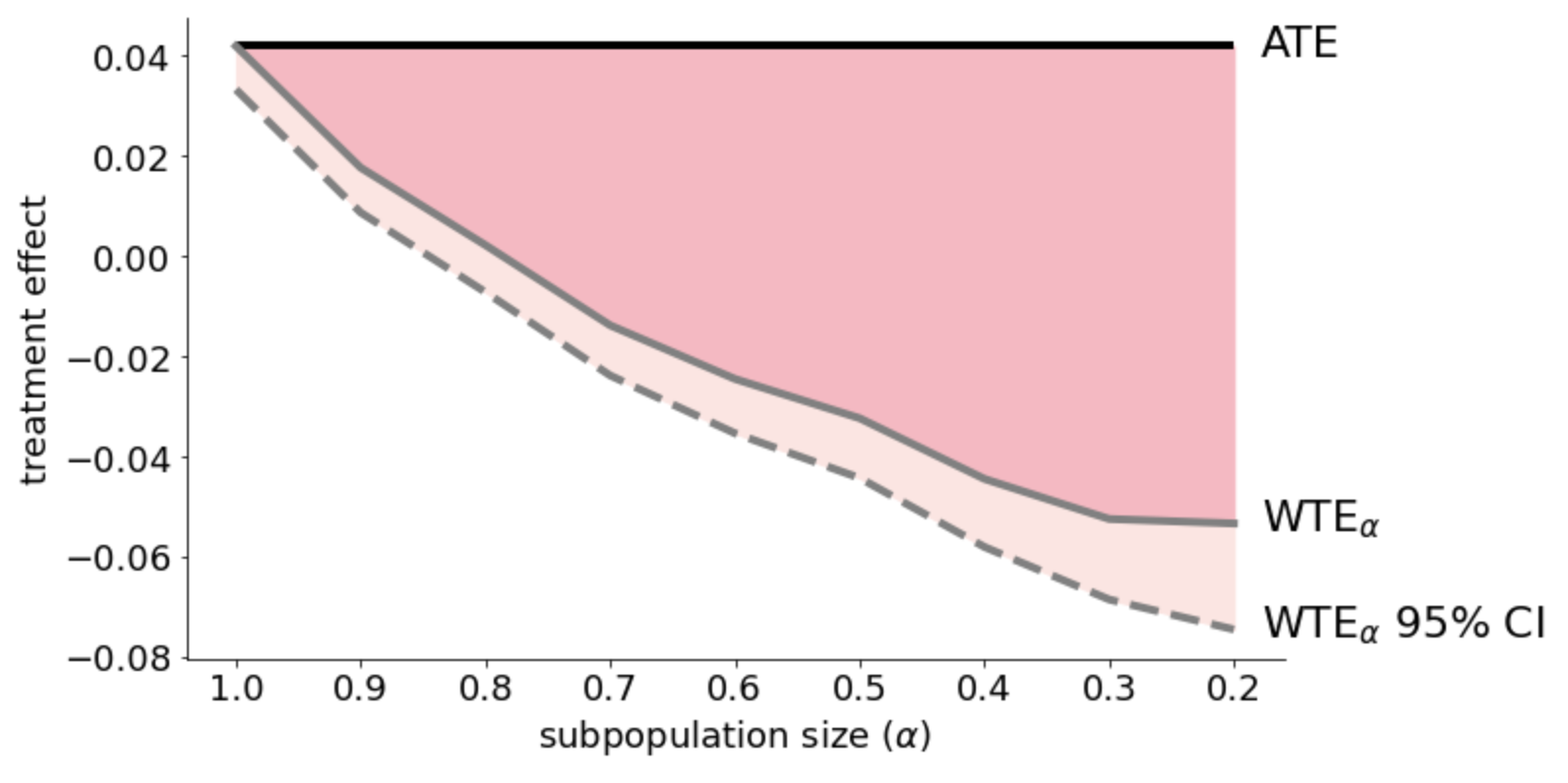} 
%\vspace{-15pt}
\caption{ \label{fig:negex_share} Effect of Medicaid enrollment on doctors
  visits in 2009}
\end{figure}

% with a budget of \$662 billion dollars a year.

Using ten years of data (2009-18) from the National Health Interview Survey
(NHIS), we focus on the binary outcome that indicates whether the individual
made any visits to doctors in the two-weeks prior to the survey
date~\citep{CurrieGr96, LiptonDe15, CurrieFa05,
  KahendeMaEnZhMoXuSeRo17}. Although survey respondents have limited control
on the outcome variable as the date of survey depends on the availability of
field agents, Medicaid enrollments are nevertheless non-random. Those in poor
health and higher intention of seeking treatment are more likely to enroll in
Medicaid and thus more likely to visit the doctors, leading to an upward bias
in confounded estimates. To address the potential bias, we control for a rich
set of baseline covariates ($d=396$) including demographics, medical history,
employment, earnings, health limitations, whether they require help with daily
tasks, and enrollment status in insurance and government programs.  We posit
that there are no unobserved confounders (Assumption~\ref{assumption:ig}); as
this is a strong assumption, we also study a randomized experiment in the
following subsection. We restrict attention to the Census West region to
ensure uniformity in the experiences of the control group across different
states. Although Medicaid eligibility cannot be determined based on the NHIS
survey data (number of family members is missing), we restrict attention to
those with annual income $\le$\$65K so that respondents have a positive
probability of eligibility.

The $\mbox{WTE}_{\alpha}$~\eqref{eqn:wte} allows analyzing external validity
over shifts in the study population. Using only data from 2009 ($n=82,993$
observations), in Figure~\ref{fig:negex_share} we plot the cross-fitted
estimates of $\mbox{WTE}_{\alpha}$ across a range of subpopulation sizes
$\alpha$. While the ATE is positive in 2009 (significant at 99\%), observed
effects fail to be significant even over subpopulations that comprise
$\alpha = 80\%$ of the study population. The WTE identifies subgroups
disparately affected by the treatment and shows the average-case findings are
invalidated under small changes to the study population. Our sensitivity
framework thus brings into question whether the observed effects in 2009 can
endure the test of time.

% \begin{figure}[t]
%   \vspace{-10pt}
% \centering
% \includegraphics[scale=.5]{./figures/wte_nhis_ts.png}
% \vspace{-20pt}
% \caption{ \label{fig:negex_ts} Effect of Medicaid enrollment on doctors visits for low-income adults in Census region West from 2009 to 2018. 95\% confidence intervals
%   for ATE are shown in the error bars.}
% \vspace{-10pt}
% \end{figure}

\begin{figure}[t]
\centering
\begin{tabular}{cc}
%  \hspace{-25pt}
  \includegraphics[scale=.27]{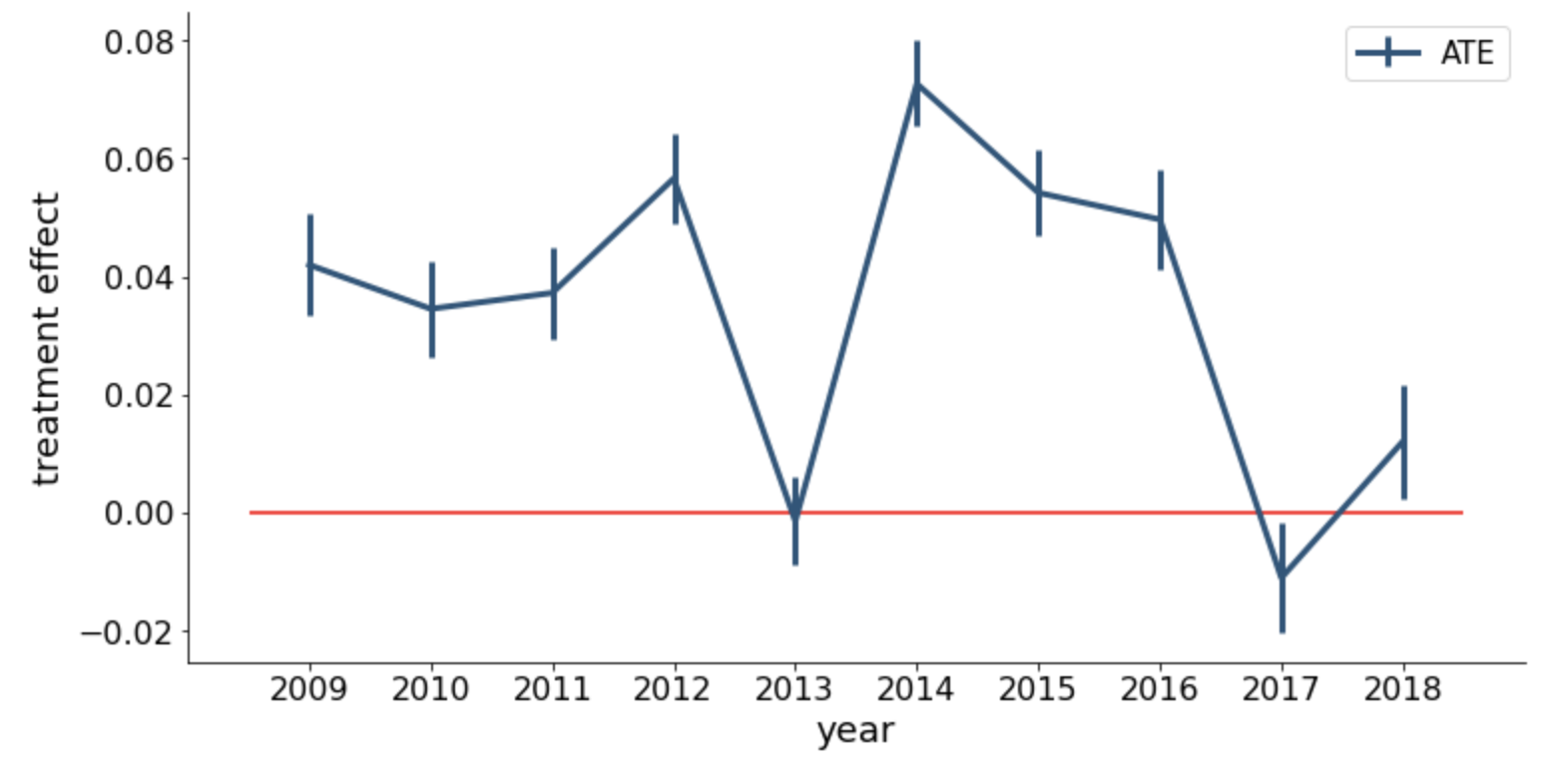}
  &
    \includegraphics[scale=.23]{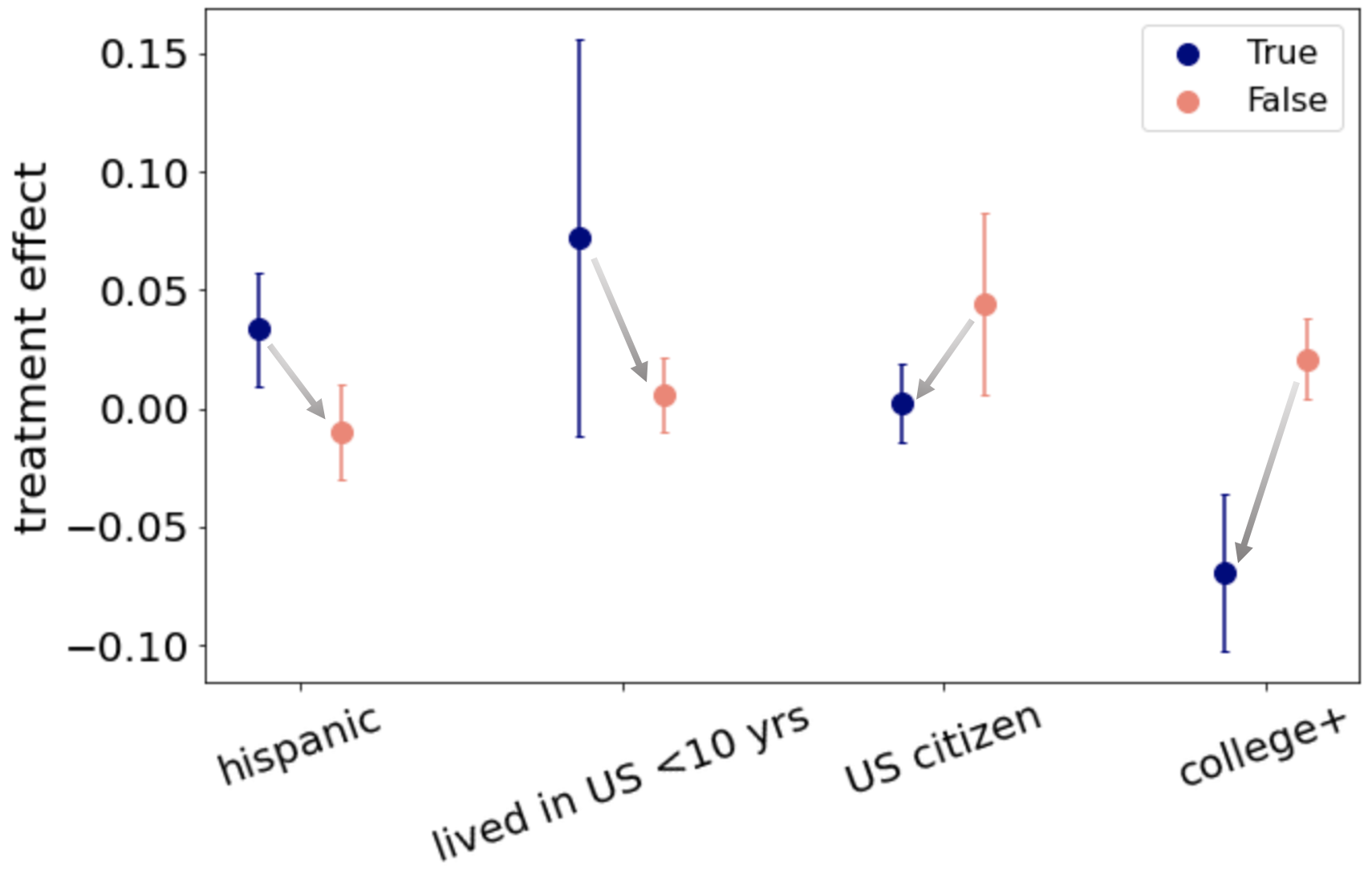}
  \\
  \small{(a) ATE estimates in 2009-18 with 95\% CIs}
  & \small{(b) CATE by covariates in 2009}
\end{tabular}
\vspace{10pt}
\caption{ \label{fig:negex_ts} Temporal changes in the ATE due to population shift}
%\vspace{-10pt}
\end{figure}

As predicted from 2009 data alone, treatment effects fail to be statistically
significant over time (Figure~\ref{fig:negex_ts}a).  To heuristically
investigate the temporal population shift in the ATE, in
Figure~\ref{fig:intro}a we analyzed covariates with the largest mean
difference between 2009 and 2018. The share of Hispanics and those who have
been in the U.S. for less than 10 years decreased, but the share of
U.S. citizens and those with high educational attainment increased. In
Figure~\ref{fig:negex_ts}b, we observe that groups whose share decreased
have higher treatment effects (and vice versa), contributing to the overall
decrease in the  treatment effect over time.

\ifdefined\useectastyle
\vspace{-15pt}
\fi
\subsection{Welfare attitudes experiment}

A large group of Americans harbors antipathy towards programs labeled
``welfare,'' a phenomenon that has generated much interest.  We are interested
in measuring how seemingly insubstantial wording changes in the description of
social welfare programs affect public support. Disdain towards ``welfare'' has
been associated with racist stereotypes towards welfare
recipients~\citep{HenryReWe04, Federico04}
% Gilens09},
and political ideology~\citep{KluegelSm86}.
% , BullockWiLi03}.
Previous works have observed substantial heterogeneity with
respect to variables such as level of racism, education levels, and political
leanings~\citep{Jacoby00, Federico04, GreenKe12}.

We study an experiment on welfare attitudes in the General Social Survey (GSS)
from 1986 to 2010~\citep{GreenKe12}. We focus on the binary outcome indicating
whether the respondents state too much is being spent on either ``welfare''
(treatment) or ``assistance to the poor'' (control). Both questions about
public spending are identical except for the wording change. Covariates
include attitude towards Blacks, political views, party identification,
educational attainment, and age; there are $n = 20,783$ data points, and
$d = 22$ covariates. To illustrate the flexibility of our estimation approach,
we estimate the propensity score using a logistic regression with elastic net
regularization as if it was unknown. (We observe similar results when we use
the true propensity score $\pprop(X) \equiv \half$.)

\begin{figure}[t]
\centering
\begin{tabular}{ccc}
\includegraphics[scale=.34]{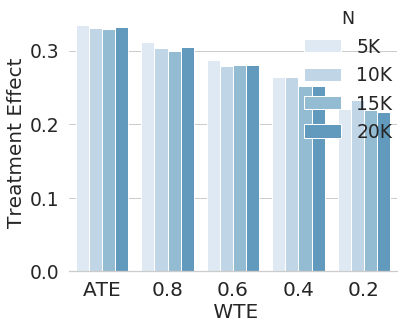}
&
\includegraphics[scale=.34]{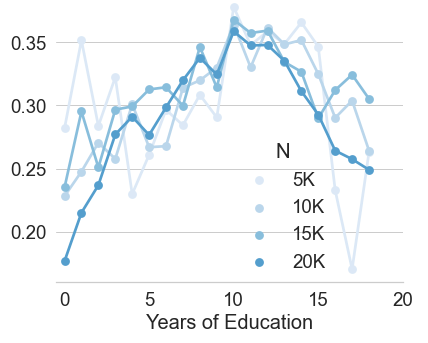}
&
\includegraphics[scale=.34]{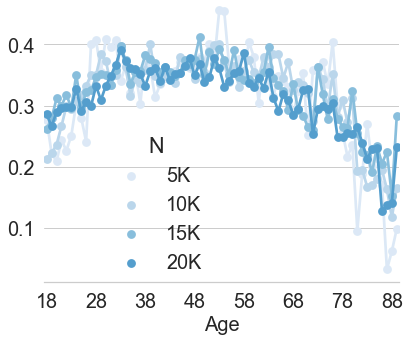}
\\
\small{(a) ATE and WTE$_\alpha$} & \small{(b) CATE by years of education}& \small{(c) CATE by age}
\end{tabular}
\vspace{10pt}
\caption{ \label{fig:posex} Treatment effects by number of observations. }
%\vspace{-10pt}
\end{figure}

By virtue of our semiparametric approach, we observe that even when CATE
estimates vary significantly across sample size and outcome model classes,
estimates of $\mbox{WTE}_{\alpha}$ align around a single value, a (empirical)
stability property shared with estimators of
ATE~\citep{CarvalhoFeMuWoYe19}. To illustrate the stability of WTE estimates,
we plot them over different sample sizes (5-20K) and outcome model classes.
% \citet{CarvalhoFeMuWoYe19} observed that even when ATE estimates
% align around the true value, CATE estimates may vary significantly over
% estimation methods. We observe that WTE estimates also remain stable across
% estimation approaches, for $\alpha$ in the range $20\% \sim 100\%$.
Even at small sample sizes, wording changes in the survey have a resoundingly
strong effect on attitude towards government welfare programs (Figure
\ref{fig:posex}a). This observed effect is uniformly significant over
subpopulations as small as 20\% of the collected data. Such robust evidence
instills confidence in the external validity of the finding across
spatiotemporal changes in the demographic composition of respondents.  While
estimates of $\mbox{WTE}_{\alpha}$ and ATE remain relatively stable across
different sample sizes, CATE estimates vary considerably over different sample
sizes, a typical behavior for undersmoothed ML models (Figure~\ref{fig:posex}b,
c).

\begin{figure}[t]
\centering
\begin{tabular}{ccc}
\includegraphics[scale=.34]{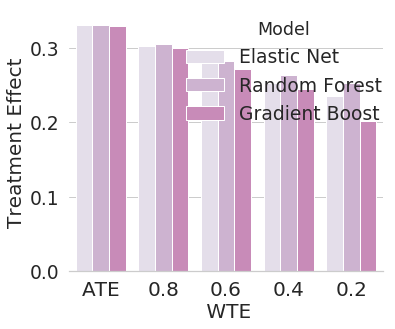}
&
\includegraphics[scale=.34]{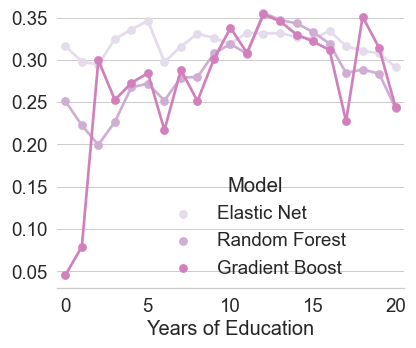}
&
\includegraphics[scale=.34]{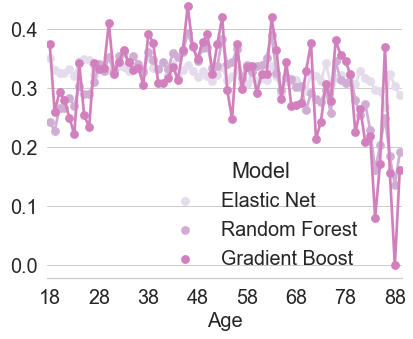}
\\
\small{(a) ATE and WTE$_\alpha$} & \small{(b) CATE by years of education}& \small{(c) CATE by age}
\end{tabular}
\vspace{10pt}
\caption{ \label{fig:posex2} Treatment effects by model class.}
%\vspace{-10pt}
\end{figure}

The stability of WTE estimates persists over different nuisance estimation
approaches.  Figure~\ref{fig:posex2} explores several common model classes for
the conditional outcome $\E[Y(z) \mid X]$: linear models with elastic net
regularizers, random forests, and gradient boosted regression trees. All
hyperparameters are chosen using 2-fold cross-validation as
before. Figure~\ref{fig:posex2}a highlights the stability of WTE and ATE
estimates along different models. In Figures~\ref{fig:posex2}b and c, we
observe that CATE estimates vary up to 25 times, especially around the
endpoint of bins, a common phenomenon often attributed to bias at the
boundaries of the support of the feature space~\citep{WagerAt18}.
% As~\citet{AtheyImPhWa17} noted, CATE estimates are sensitive to the choice
% of different regularization methods and model choices, and inference on
% resulting personalized estimates can be difficult.  WTE provides a way for
% practitioners to account for treatment effect heterogeneity while
% maintaining insensitive to the CATE estimation approach.
We anticipate that WTE estimators will similarly suffer instability issues
when $\alpha$ is exceedingly small due to limited sample size.

\section{Asymptotics}
\label{section:asymptotics}

% Neyman orthogonality
% Semiparametrics, advantages of DML
% quantile can be estimated efficiently

We now show that our cross-fitted augmented estimator
(Algorithm~\ref{alg:cross-fitting}) enjoys central limit behavior
$\frac{\sqrt{n}}{\what{\sigma}_{\alpha}} (\what{\omega}_{\alpha} -
\mbox{WTE}_{\alpha}) \cd N(0, 1)$ even when we can only estimate the nuisance
parameters~\eqref{eqn:nuisance-def} at slower-than-parametric rates.  The
Neyman orthogonality condition~\citep{Neyman59} serves a central role in our
analysis.
\begin{definition}
  \label{def:neyman}
  Let $Q \mapsto T(Q; \gamma)$ be a statistical functional with nuisance
  parameter $\gamma \in \Gamma$, where we take $\Gamma$ to be a subset of a
  normed vector space containing the true nuisance parameter $\gamma\opt$. The
  functional $T$ is Neyman orthogonal at $P$ if for all $\gamma \in \Gamma$,
  the derivative $\frac{d}{dr} \E_P[T(P; \gamma\opt + r(\gamma-\gamma\opt))]$
  exists for $r \in [0, 1)$, and is zero at $r = 0$.
\end{definition}
We use the augmented form~\eqref{eqn:aug} as our statistical functional
$T(Q; \mu_0, \mu_1, e, h)$ 
\begin{align}
  \label{eqn:functional}
  T(Q; \mu_0, \mu_1, e, h)
  \defeq \inf_{\eta} \left\{ \frac{1}{\alpha}
  \E_{D \sim Q}\hinge{\mu(X) - \eta} + \eta
  \right\} + \E_{D \sim Q}[\kappa(D; \mu_0, \mu_1, e, h)],
\end{align}
where we use $\mu(x) \defeq \mu_1(x) - \mu_0(x)$ as usual. The first term is
the dual form of $\mbox{WTE}_{\alpha}$ (Lemma~\ref{lemma:dual}), and the
second term is the augmentation term defined in Eq.~\eqref{eqn:aug}. Since
$\E_{D \sim P}[\kappa(D; \pall)]=0$ under ignorability
(Assumption~\ref{assumption:ig}), we have $T(P; \pall) = \mbox{WTE}_{\alpha}$.
\ifdefined\useectastyle An application of the envelope theorem confirms Neyman
orthogonality of the augmented functional~\eqref{eqn:functional}.
\else

To build intuition, we first informally argue that this augmented functional
satisfies Neyman orthogonality. We first compute the (Gateaux) derivative of
the first dual infimization term in the
functional~\eqref{eqn:functional}. From Danskin's theorem~\citep[Theorem
4.13]{BonnansSh00}, the derivative of the dual formulation is the derivative
of the objective at the unique optimal solution. Under sufficient regularity
conditions, the unique solution to the dual is given by the quantile
$\aq{\pmu}$, and the derivative at $r = 0$ is
\begin{equation*}
  \frac{d}{dr} \inf_{\eta} \left\{ \frac{1}{\alpha}
  \E\hinge{(\pmu + r (\mu - \pmu))(X)  - \eta} + \eta
  \right\} \Bigg|_{r=0} = \E[\pthr(X) (\mu - \pmu)(X)].
\end{equation*}
To compute the derivative of the second term, let
$\gamma = (\mu_0, \mu_1, e, h)$ to ease notation. So long as we can
interchange derivatives and expectations, it is straightforward to calculate
\begin{align*}
  \frac{d}{dr} \E\left[\kappa\left(D; \gamma\opt + r (\what{\gamma}_{k} -
    \gamma)\opt\right)\right] |_{r = 0} = -\E[\pthr(X) (\mu - \pmu)(X)]
\end{align*}
from ignorability (Assumption~\ref{assumption:ig}). This verifies Neyman
orthogonality of the functional~\eqref{eqn:functional}.

\fi

Orthogonality allows us to show a central limit theorem for our augmented
cross-fitting estimator $\what{\omega}_{\alpha}$ under the following weak rate
requirements for the nuisance parameters. Recall the bound $c >0$ on the
propensity score given in Assumption~\ref{assumption:overlap}.
\begin{assumption}
  \label{assumption:neyman} Let $\Lone{\hmu -\pmu} \cas 0$, and let there
  exist an envelope function $\env: \mc{X} \to \R$ satisfying
  $\E[\env(X)^2] < \infty$ and $\max(|\hmzero|, |\hmone|) \le \env$.  There
  exists $\delta_n, \Delta_n \downarrow 0$, and $M_h > 0$ such that with probability at least
  $1-\Delta_n$, for all $\indfold \in [K]$
  \begin{enumerate}[(a)]
  \item
    {\small $\Ltwo{\hprop - \pprop} + \Ltwo{\hmzero - \pmzero} + \Ltwo{\hmone -
      \pmone} \le \delta_n$, $\hprop \in [c, 1-c]$}, and $|\hthr| \le \hbound$
\item {\small $\Ltwo{\hprop - \pprop}\left(\Ltwo{\hmzero - \pmzero} + \Ltwo{\hmone - \pmone}\right)
    \le \delta_n n^{-1/2}$}
  \item $\Linf{\hmu - \pmu} \le \delta_n n^{-1/3}$, and
    $|\hq - \quantile{1-\alpha}{\hmu}| \le \delta_n n^{-1/3}$
  \end{enumerate}
\end{assumption}

In particular, Assumption~\ref{assumption:neyman} does not require a
$\sqrt{n}$-convergence rate (Donsker condition) on the estimators of nuisance
parameters. The first two conditions are standard convergence
rates~\citep{ChernozhukovChDeDuHaNeRo18}, also required for proving a central
limit result for the $\mbox{ATE}$. They hold, in particular, when
$\Ltwo{\what{e} - \pprop} = o_p(n^{-1/4})$ and
$\Ltwo{\what{\mu}_z - \pmu_z} = o_p(n^{-1/4})$ for $z \in \{0, 1\}$. The third
condition guarantees approximation of the threshold function
$\pthr(x) = \alpha^{-1} \indic{\pmu(x) \ge \aq{\pmu}}$ at suitably fast rates.
The requirement $\Linf{\hmu - \pmu} \le \delta_n n^{-1/3}$ states that the
CATE be estimated at a somewhat faster rate compared to the case for
estimating the ATE.  \ifdefined\useectastyle \else In
Appendix~\ref{section:example}, we provide detailed examples of model classes
and learning methods where these convergence rates hold.  \fi As noted in
Section~\ref{section:approach}, when \emph{unlabeled} covariates (those
without corresponding outcomes nor treatments) are cheaply available, the
second part of $(c)$ can be easily achieved.

As the WTE is a tail-average of the CATE above the quantile $\aq{\pmu}$
(recall Lemma~\ref{lemma:dual}), to estimate $\mbox{WTE}_{\alpha}$ we need to
estimate the quantile $\aq{\pmu}$. Towards this goal, we require that a
positive density exists at its $(1-\alpha)$-quantile, a standard condition
required for quantile estimation~\cite[Chapter 3.7]{VanDerVaartWe96}.
Let~$\mc{U}$ be a subset of measurable functions $\mu: \mc{X} \to \R$ such
that the following holds:
\begin{quote}
  $F_{r, \mu}$, the cumulative distribution of
  $(\pmu + r(\mu - \pmu))(X)$, is uniformly differentiable in $r \in [0, 1]$
  at $\aq{\pmu + r(\mu - \pmu)}$, with a positive density. Formally, if we let
  $q_{r, \mu} \defeq \aq{\pmu + r(\mu - \pmu)}$, then for each $r \in [0, 1]$,
  there is a positive density $f_{r, \mu}(q_{r, \mu}) > 0$ such that
\begin{equation}
  \label{eqn:unif-diff}
  \lim_{t \to 0} \sup_{r \in [0, 1]}
  \left| \frac{1}{t} \left(F_{r, \mu}(q_{r, \mu} + t)
      - F_{r, \mu}(q_{r, \mu})\right)
    - f_{r, \mu}(q_{r, \mu}) \right| = 0.
\end{equation}
\end{quote}
We require this holds for our estimators $\mu = \hmu$ with high probability.
\begin{assumption}
  \label{assumption:regularity}
 $\exists \Delta_n' \downarrow 0$ s.t. with probability at least
  $1-\Delta_n'$, $\hmu \in \mc{U}$ for all $\indfold \in [K]$.
\end{assumption}
\noindent In particular, Assumption~\ref{assumption:regularity} requires
$\pmu(X)$ to have a positive density at $\aq{\pmu}$.

We are now ready to give our main technical result which shows that the
augmented cross-fitting estimator $\what{\omega}_{\alpha}$ enjoys central
limit rates with the influence function
\begin{align}
  \label{eqn:influence}
  \psi(D) \defeq
  \frac{1}{\alpha} \hinge{\pmu(X) - \aq{\pmu}} + \aq{\pmu}
  - \mbox{WTE}_{\alpha} 
  + \kappa\left(D; \pall\right).
\end{align}
Indeed, $\psi(D)$ is a valid influence function under
Assumption~\ref{assumption:ig}: we have $\E[\psi(D)] = 0$, and
$\var(\psi(D)) = \sigma^2_{\alpha}$ where $\sigma_{\alpha}^2$ is the
asymptotic variance defined in expression~\eqref{eqn:var}.
\begin{theorem}
  \label{theorem:clt}
  Under
  Assumptions~\ref{assumption:ig}-\ref{assumption:regularity},
  $\sqrt{n} (\what{\omega}_{\alpha} - \mbox{WTE}_{\alpha}) =
  \frac{1}{\sqrt{n}} \sum_{i=1}^n \psi(D_i) + o_p(1)$. 
  Further, we have $\what{\sigma}^2_{\alpha} \cp \sigma^2_{\alpha}$ and
  $\frac{\sqrt{n}}{\what{\sigma}_{\alpha}} (\what{\omega}_{\alpha} -
  \mbox{WTE}_{\alpha}) \cd N(0, 1)$.
\end{theorem}
\noindent Since the proof of Theorem~\ref{theorem:clt} is involved, we give
its sketch below, emphasizing how we leverage Neyman orthogonality of the
augmented functional~\eqref{eqn:functional} to obtain our result. We provide
rigorous details of the proof in
Appendices~\ref{section:proof-clt-classical},~\ref{section:proof-clt-neyman},~\ref{section:proof-clt-var}.
\citet{ChernozhukovChDeDuHaNeRo18} showed that the solution of a Neyman
orthogonal \emph{estimating equation} could be estimated at the usual central
limit rate, even when nuisance parameters converge at slower rates.  Their
results do not apply to the $\mbox{WTE}_{\alpha}$ as it is a nonlinear
statistical functional of the underlying probability measure $P$. To account
for such nonlinearity, our theoretical analysis considerably extends existing
results.  Using tools from empirical process theory~\citep{VanDerVaartWe96},
our main result shows that for uniformly Hadamard differentiable functionals,
orthogonality still allows insensitivity to estimation error in nuisance
parameters. Even when $\alpha = 100\%$ so that $\mbox{WTE}_{1} = \mbox{ATE}$
and our estimator reduces to the cross-fitted AIPW for the ATE, our argument
provides a different proof to that given
by~\citet{ChernozhukovChDeDuHaNeRo18}.

\subsection*{Sketch of proof for Theorem~\ref{theorem:clt}}
\label{section:proof-clt}

Our proof proceeds in three parts.  Recalling $\what{P}_k$, the empirical
distribution on the $k$-th fold, our cross-fitted estimator can be written
succinctly as $\frac{1}{K} \sum_{\indfold = 1}^{K} T(\what{P}_k; \hall)$. We
emphasize (empirical) expectations over $Q$ in the
definition~\eqref{eqn:functional} are taken only over $D$, and not over the
randomness in $(\hall)$.

The first two parts show that for each fold $\indfold \in [K]$,
\ifdefined\useectastyle
{\small
\begin{align}
  \label{eqn:fold-clt}
  \sqrt{|I_k|} \left( T\left(\what{P}_k; \hall\right)
  - T(P; \pall) \right) = \frac{1}{\sqrt{|I_k|}} \sum_{i \in I_k} \psi(D_i)
  + o_p(1).
\end{align}
}
\else
\begin{align}
  \label{eqn:fold-clt}
  \sqrt{|I_k|} \left( T\left(\what{P}_k; \hall\right)
  - T(P; \pall) \right) = \frac{1}{\sqrt{|I_k|}} \sum_{i \in I_k} \psi(D_i)
  + o_p(1).
\end{align}
\fi
Since $T(P; \pall) = \mbox{WTE}_{\alpha}$ by ignorability
(Assumption~\ref{assumption:ig}), this gives our first result. Towards this
goal, decompose the left hand side of the equality~\eqref{eqn:fold-clt} into
\begin{subequations}
  \label{eqn:two-parts}
  \begin{align}
    & \sqrt{|I_k|} \left( T\left(\what{P}_k; \hall\right) - T\left( P; \hall\right)
      \right) \label{eqn:part-classical} \\
    & + \sqrt{|I_k|} \left( T\left(P; \hall\right) - T(P; \pall) \right).
      \label{eqn:part-neyman}
  \end{align}
\end{subequations}
In Part I of the proof (Appendix~\ref{section:proof-clt-classical}), we prove
that the first term~\eqref{eqn:part-classical} is asymptotically equal to
$\frac{1}{\sqrt{|I_k|}} \sum_{i \in I_k} \psi(D_i)$. Leveraging tools from empirical
process theory~\citep{VanDerVaartWe96}, we first show that the functional
$Q \mapsto T(Q; \hall)$ satisfies a uniform variant of Hadamard
differentiability so that the functional delta method applies. A careful
application of a uniform version of the Lindeberg-Feller central limit theorem
over functions gives the desired conclusion.

In Part II (Appendix~\ref{section:proof-clt-neyman}), we use Neyman
orthogonality of our augmented estimator to show that the second
term~\eqref{eqn:part-neyman} vanishes asymptotically.  Define
$\rem: [0, 1] \to \R$ \ifdefined\useectastyle {\small
\begin{align}
  \label{eqn:remainder}
  \rem(r) \defeq
  T\left(P; (1-r) (\pall) + r (\hall)\right) - T(P; \pall),
\end{align}
}
\else
\begin{align}
  \label{eqn:remainder}
  \rem(r) \defeq
  T\left(P; (1-r) (\pall) + r (\hall)\right) - T(P; \pall),
\end{align}
\fi
so that $\rem(0) = 0$, and $\rem(1)$ is equal to the second
term~\eqref{eqn:part-neyman}. $\rem(r)$ is continuously differentiable under
suitable conditions and mean value theorem gives 
\begin{align*}
  \rem(1) = \rem(0) + \rem'(r) \cdot (1-0) = \rem'(r)
\end{align*}
for some $r \in [0, 1]$.  From Neyman orthogonality, we have $\rem'(0) = 0$.
Building on this, we show that all values of $\rem'(r)$ are sufficiently
small: under postulated convergence rates for the nuisance parameters in
Assumption~\ref{assumption:neyman},
$\sup_{r \in [0, 1]} |\rem'(r)| = o_p(n^{-1/2})$.

In Part III (Appendix~\ref{section:proof-clt-var}), we show consistency of our
variance estimator: $\what{\sigma}^2_{\alpha} \cp
\sigma^2_{\alpha}$. Combining this with the central limit
result~\eqref{eqn:fold-clt}, Slutsky's lemma gives our second
result. $\diamond$

%%% Local Variables:
%%% mode: latex
%%% TeX-master: "main"
%%% End:

% approach empirically in Section~\ref{section:experiments}, demonstrating that
% our estimator provides practical and efficient bounds on treatment effects
% under covariate shift.

%%% Local Variables:
%%% mode: latex
%%% TeX-master: "main"
%%% End:

\section{Semiparametric efficiency bound}
\label{section:efficiency}

We now establish a semiparametric efficiency bound, showing that all (regular)
estimators of $\mbox{WTE}_{\alpha}$ necessarily have asymptotic variance
larger than $\sigma_{\alpha}^2$, both when the true propensity score
$\pprop(\cdot)$ is known and unknown. In particular, this implies that our
augmented estimator $\what{\omega}_{\alpha}$ achieves the \emph{optimal}
asymptotic variance and that its influence function~\eqref{eqn:influence} is
the efficient influence function for estimating $\mbox{WTE}_{\alpha}$. Our
efficiency bound i) informs the design of experiments by providing the minimum
number of study participants required to reach a conclusion that is valid
across subpopulations no smaller than $\alpha$ and ii) quantifies how the
required sample size grows with the level of desired external validity
$\alpha$.

For parametric problems, Hajek-Le Cam theorems~\cite[Chapter 8]{VanDerVaart98}
give a lower bound on the asymptotic variance of regular estimators, more
generally, a lower bound on the mean squared error for any estimator. Since
these bounds coincide with the Cramer-Rao bound for unbiased estimators, they
can be considered as an asymptotic Cramer-Rao bound. We consider parametric
submodels of our semiparametric problem, i.e., finite-dimensional
parameterizations that contain the truth. Since the asymptotic variance of any
(smooth enough) semiparametric estimator is worse than the Hajek-Le Cam
bound---equivalently, the Cramer-Rao bound---of any parametric submodel, the
semiparametric efficiency bound is defined as the supremum of the Hajek-Le Cam
bound over all parametric submodels.  As these definitions are standard yet
tedious~\citep{Newey90, BickelKlRiWe93}, we defer a formal treatment to
Appendix~\ref{section:proof-efficiency}.

We now characterize the semiparametric efficiency bound for estimating
$\mbox{WTE}_{\alpha}$.
\begin{assumption}
  \label{assumption:efficiency}
  $\pmu(X)$ has a positive density, and {\small
    $\Linf{\E[|Y(z)| \mid X]} < \infty$, $z \in \{0, 1\}$}.
\end{assumption}
\begin{theorem}
  \label{theorem:efficiency}
  Let  Assumptions~\ref{assumption:ig},~\ref{assumption:overlap},~\ref{assumption:residuals},~\ref{assumption:efficiency}
  hold, and $\sigma^2_{\alpha} = \E[\psi(D)^2] > 0$, where $\psi$ was defined in
  Eq.~\eqref{eqn:influence}. Then, $\mbox{WTE}_{\alpha}$ is a differentiable
  parameter in the sense of Definition~\ref{def:diff}, and its semiparametric
  efficiency bound is given by $\E[\psi(D)^2]$. The efficiency bound is
  identical if the true propensity score is known.
\end{theorem}
\noindent See Appendix~\ref{section:proof-efficiency} for the
proof. Theorem~\ref{theorem:clt} and~\ref{theorem:efficiency} show that our
cross-fitted augmented estimator $\what{\omega}_{\alpha}$ achieves the
semiparametric efficiency bound $\mathbb{V}(\mbox{WTE}_{\alpha}; P)$. Similar
to the efficiency bound for the ATE~\citep{Hahn98}, the knowledge of
$\pprop(\cdot)$ does not affect the efficiency bound
$\mathbb{V}(\mbox{WTE}_{\alpha}; P)$. We conclude that our estimator is
semiparametrically efficient for \emph{both} observational studies and
randomized control trials.

Our semiparametric efficiency bound allows calculating the minimum required
sample size for finding conclusions that are robust against all subpopulations
of size at least $\alpha$. Consider the hypothesis test
$H_0: \mbox{WTE}_{\alpha} \ge 0 ~\mbox{v.}~H_{\rm A}: \mbox{WTE}_{\alpha} <
-\epsilon$, where $\epsilon > 0$ is the specified minimum detectable effect
size (recall that the desired sign of the treatment effect is negative). As an
illustration, consider a test with size ($\P(\mbox{Type I error})$) at most
$5\%$ and power ($1 -\P(\mbox{Type II error})$) at least $80\%$. A standard
power calculation shows $n \ge 6.2 \frac{\sigma_{\alpha}^2}{\epsilon^2}$
samples are needed to detect a worst-case subpopulation treatment effect of
size $-\epsilon$. In particular, this number grows as we require a stronger
level of external validity, or equivalently as the worst-case subpopulation
size $\alpha$ becomes small.

%%% Local Variables:
%%% mode: latex
%%% TeX-master: "main"
%%% End:

\section{Discussion}
\label{section:discussion}

% \paragraph*{Related work}
% \label{section:related-work}

% Distribution shifts occur in many different forms
% across statistics, econometrics, epidemiology, machine learning, and
% operations research. We provide an abridged review of the literature.

% A number of authors have recently leveraged Neyman orthogonality in
% semiparametric inference and statistical
% learning~\citep{ChernozhukovChDeDuHaNeRo18, ChernozhukovNeSeSy18, ZadikMaSy18,
%   FosterSy19}.~
% ~\citet{FosterSy19} studies statistical learning problems with nuisance
% components, and shows insensitivity to nuisance parameter estimation error for
% their two-stage learn-then-optimize procedure. In contrast to their learning
% setting, we are interested in estimating the \emph{optimal objective value}
% given by the $\mbox{WTE}_{\alpha}$~\eqref{eqn:wte}, and our augmented
% estimator is designed specifically to be orthogonal for this purpose.

Motivated by challenges in evaluating treatments under unanticipated
population shifts, we proposed a sensitivity analysis framework based on the
worst-case subpopulation treatment effect. We advocate for such conservatism
in important policy decisions that need to benefit all subpopulations
uniformly.  % We derived a augmented estimator for the WTE by leveraging its
% dual tail-average representation. Our cross-fitting procedure allows
% flexible use of statistical methods for estimating nuisance parameters, and
% enjoys central limit rates even when nuisance estimates converge more
% slowly.
% The WTE is agnostic to the unknown target (sub)population, and positive
% findings with respect to the WTE guarantees a treatment effect over all
% subpopulations of a given size.
While the WTE~\eqref{eqn:wte} provides a strong notion of robustness, it may
be overly conservative in scenarios where one is concerned with more
structured covariate shifts. When covariate shift on only a small subset of
$X$ is of interest, the definition~\eqref{eqn:wte} can be modified over this
subset. More broadly, studying structured shifts in the covariate distribution
$P_X$ is an exciting direction of future work.

Our theoretical developments show that our cross-fitted augmented estimator
inherits two advantageous inferential properties of the AIPW estimator for the
ATE: orthogonality and efficiency. Elementary derivations show, however, that
our estimator does not satisfy the doubly robust property due to the
additional nuisance parameter
$\pthr(x) = \alpha^{-1} \indic{\pmu(x) \ge \aq{\pmu}}$. Another limitation of
our augmented estimator is that it is not necessarily increasing in $\alpha$.
Though the extension of the direct method and the inverse probability weighted
estimator are increasing in $\alpha$, it is unclear whether an orthogonal,
efficient, and monotone estimator of the WTE exists. Finally, estimators of
tail-averages often suffer higher variance and may be more sensitive to lack
of overlap and unobserved confounding. Further investigation of these issues
is a topic for future research.

% If ignorability holds only with respect to the full covariate vector,
% regressing the full CATE estimate on the small subset of covariates gives a
% direct method for this less conservative variant of the WTE.
Sometimes it is of interest to estimate the \emph{average treatment effect on
  the treated} $\E[Y(1) - Y(0) \mid Z = 1]$. There are two natural ways to
extend the worst-case definition~\eqref{eqn:wte} to measure the effect on the
treated, depending on whether the worst-case is still taken over $Q_X$, or
over the conditional distribution $Q_{X \mid Z = 1}$. We leave a systematic
study of the two definitions and corresponding inferential frameworks to
future work.

%%% Local Variables:
%%% mode: latex
%%% TeX-master: "main"
%%% End:

\paragraph*{Acknowledgments} We thank Steve Yadlowsky for helpful comments.

%%% Local Variables:
%%% mode: latex
%%% TeX-master: "colt-main"
%%% End:

%% ========================== Bibliography ========================== %%

\bibliographystyle{abbrvnat}

\setlength{\bibsep}{.7em}
\bibliography{bib}

\ifdefined\usemsstyle

\ECSwitch

%\ECDisclaimer
%%%%%%%%%%%%%%%%%%%%%%%%%%%%%%%%%%%%%%%%%%%%%%%%%%%%%%%%%%

%% ========================== Appendix ========================== %%

%%% Main head for the e-companion
\ECHead{Appendix}

\else
\newpage
\appendix
\fi

\paragraph*{Notation} Throughout, we let $F_{\mu}$ denote the cumulative
distribution of $\mu(X)$, and let
$\aq{\mu} \defeq \inf\{t: F_{\mu}(t) \ge 1-\alpha\}$ be the
$(1-\alpha)$-quantile of $\mu(X)$; when $\mu(\cdot)$ is random (e.g. estimated
from data), the probabilities are taken only over $X \sim P$.  $L^{q}(\mc{X})$
is the $L^q$ norm w.r.t. $X \sim P$.  For $a_n \in \R$, we write
$X_n = O_P(a_n)$ if
$\lim_{b \to \infty} \limsup_n P(|X_n| \ge b \cdot a_n) = 0$, and
$X_n = o_P(a_n)$ if $\limsup P(|X_n| \ge b \cdot a_n) = 0$ $\forall b > 0$.

\section{Review of nuisance estimation approaches}
\label{section:example}

In this section, we review standard loss minimization approaches to estimating the
nuisance parameters $\pmone, \pmzero, \pprop$. For ease of exposition, we focus
on estimation of $\pmone = \E[Y(1) | X = x]$; directly analogous results are
available for $\pmzero$ and $\pprop$. Our starting point is the fact that
$\pmone$ is the solution to the loss minimization problem
\begin{equation}
  \label{eqn:opt}
  \minimize_{\mu_1(\cdot):~\mbox{\scriptsize measurable}}
  \E[ (Y - \mu_1(X))^2 \mid Z = 1].
\end{equation}
We consider empirical plug-in estimators $\hmone$ on the auxiliary sample
$\cfold$, given by the (approximate) solution to the following optimization
problem
\begin{equation}
  \label{eqn:opt-emp}
  \minimize_{\mu_1} 
  \sum_{i \in \cfold} Z_i(Y_i - \mu_1(X_i))^2,
\end{equation}
where the minimization problem is taken over a suitably chosen (possibly
regularized) model class.

The guarantees we focus on below scale separately in the complexity of model
classes for $\hmone$ and $\hmzero$. On the other hand, our most stringent rate
requirement $\Linf{\hmu - \pmu} = o_p(n^{-1/3})$ only requires convergence of
$\hmu = \hmone - \hmzero$; see~\citet{KunzelSeBiYu19} for procedures that
scale with the complexity of $\hmu$, which are advantageous when $\pmu$
has significantly lower complexity than $\pmone$ and $\pmzero$.

\subsection{High-dimensional estimation}
\label{section:example-param}

We consider two typical estimation scenarios involving high dimensional
covariates $X$. Our guarantees in this subsection require that the model class
$\mc{M}$ we optimize over in the empirical problem~\eqref{eqn:opt-emp} be
well-specified (i.e., $\pmone \in \mc{M}$). We relax this condition in
Section~\ref{section:example-sieve}, where we consider nonparameteric
estimators.

\begin{example}[Sparse linear regresion]
  Consider the sparse linear regression problem
  \begin{align*}
    Y(1) = \theta^{\star\top} X + \varepsilon,
  \end{align*}
  where $\varepsilon \indep X$, $\E[\epsilon] = 0$, and
  $\linf{\varepsilon} \le 1$ (so that $\pmone(X) = \theta^{\star \top}X$). We
  consider the scenario where $\theta\opt$ is $s$-sparse,
  $\norm{\theta\opt}_0 \le s$, and satisfies $\lone{\theta\opt} \le 1$. We
  assume $\linf{X} \le 1$ a.s. for simplicity.

  Let $\hmone$ be the solution to the empirical optimization
  problem~\eqref{eqn:opt-emp} over the model class
  \begin{align*}
    \mc{M} = \left\{ x \mapsto \theta^\top x: \norm{\theta}_0 \le s, \lone{\theta}
    \le 1 \right\}.
  \end{align*}
  % cite Bartlett, Wainwright, Van Der Vaart
  A standard localized Rademacher complexity argument~\citep{BartlettBoMe05,
    Wainwright19} and another standard covering bound over a class of linear
  functions~\citep{Zhang02, VanDerVaartWe96} show that with probability at least
  $1-\Delta$,
  \begin{equation*}
    \Ltwo{\hmone - \pmone}
    = O\left(\sqrt{\frac{s \log(d / s)}{n}} + \sqrt{\frac{\log(1/\Delta)}{n}}\right).
  \end{equation*}
  We conclude that the rate requirements for $\hmone$ given in
  Assumption~\ref{assumption:neyman} hold whenever $s \log d \ll n^{1/3}$.

  Similar rates can be shown for the (convex) Lasso-regularized model class
  $\{x \mapsto \theta^\top x: \lone{\theta} \le \lone{\theta\opt}\}$ under the
  standard restricted eigenvalue conditions on $\{X_i\}_{i \in
    \cfold}$. Variants of these results can also be shown when these norm
  constraints appear as regularizers in the objective. We refer the reader
  to~\citet{Wainwright19} and~\citet[Chapter 11]{HastieTiWa15} for a detailed
  overview of related results.
  % convex constraint (lasso) under restricted eigenvalue assumption,
  % regularization
\end{example}

\begin{example}[Neural networks]
  Consider neural networks with ReLU activations $v(a) = \max(a, 0)$
  \begin{align*}
    \mc{M} = \left\{
    \mu_1(x) = A_L v\left( A_{L-1} v\left( \cdots v\left( A_1 x\right)
    \cdots\right) \right):
    A_l \in \R^{d_l \times d_{l-1}}, \Linf{\mu_1} \le C
    % \norm{A_l}_{\rm sp} \le C_{1,l},
    % \norm{A_l}_{2, 1} \le C_{2, l}~~\mbox{for all}~1 \le l \le L
    \right\},
  \end{align*}
  where $d_0 = d$ and $d_L = 1$.  We assume that the depth of the network,
  $L$, is deep enough to represent the true parameter $\pmone$ so that
  $\pmone \in \mc{M}$ and $Y(1) = \pmone(X) + \varepsilon$, where
  $\varepsilon \indep X$, $\E[\epsilon] = 0$, and $\linf{\varepsilon} \le 1$.

  Let $d_{\mc{M}} = \prod_{l=1}^L d_l d_{l-1}$ be the number of parameters
  in the neural network.  Again, standard a localized Rademacher complexity
  argument~\citep{BartlettBoMe05, Wainwright19} and bounds on the VC dimension
  for neural networks~\cite[Theorem 6]{BartlettHaLiMe19} yields
  \begin{align*}
    \Ltwo{\hmone - \pmone}
    = O\left(\sqrt{\frac{d_{\mc{M}} L \log d_{\mc{M}} \log C }{n}}
    + \sqrt{\frac{\log(1/\Delta)}{n}}\right)
  \end{align*}
  with probability at least $1-\Delta$.  Rate requirements for $\hmone$ given
  in Assumption~\ref{assumption:neyman} hold whenever
  $d_{\mc{M}} L \log d_{\mc{M}} \log C \ll n^{1/3}$.
\end{example}

\subsection{Sieve estimation}
\label{section:example-sieve}

\newcommand{\covspace}{\mc{X}} % space where covariate vectors lie
\newcommand{\covdim}{d} % dimension of x \in \covspace
\newcommand{\holderspace}[1]{\Lambda^{#1}(\covspace)} % space of Holder smooth ftns
\newcommand{\holderball}[2]{\Lambda^{#1}_{#2}(\covspace)} % ball with Holder-norm
\newcommand{\holderradius}{R} % Holder-norm radius
\newcommand{\holdersmooth}{p} % Holder smoothness parameter

\newcommand{\comp}{J} % complexity (e.g. order of polynomial)
\newcommand{\pol}[1]{\mbox{Pol}\left(#1\right)} % polynomial class
\newcommand{\tripol}[1]{\mbox{TriPol}\left(#1\right)} % trigonometric polynomial class
\newcommand{\spl}[2]{\mbox{Spl}\left(#1, #2\right)}

\newcommand{\funcparam}{\mu} % placeholder for nuisance
\newcommand{\funcspace}{\mc{M}} % space of nuisance variables
\newcommand{\empfunc}{\what{\funcparam}_1} % population nuisance
\newcommand{\popfunc}{\funcparam_1} % population nuisance

\newcommand{\funcscalar}{t}

\newcommand{\mulower}{\mu_1^-}

\newcommand{\probfunc}{\nu}
\newcommand{\popprob}{\nu_1}
\newcommand{\empprob}{\what{\nu}_1}
\newcommand{\emppopprob}{\what{\nu}'}
\newcommand{\probspace}{\Pi}
\newcommand{\empfuncp}{\what{\funcparam}^{\nu_1}_1}

\newcommand{\probsmooth}{q}
\newcommand{\probradius}{r}

\newcommand{\treatment}{z}
\newcommand{\treatmentrv}{Z}
\newcommand{\treatedE}{\E_1}
\newcommand{\empE}{\E_{n}}
\newcommand{\emptreatedE}{\E_{1, n}}
\newcommand{\emptreatedP}{P_{1, n}}
\newcommand{\treatedP}{P_1}
\newcommand{\treatedp}{p_1}
\newcommand{\numtreated}{{n_1}}
\newcommand{\covariate}{x}
\newcommand{\covariaterv}{X}
\newcommand{\estfunc}{a}
\newcommand{\outcome}{y}
\newcommand{\outcomerv}{Y}

\newcommand{\covmod}{\delta}
\newcommand{\ltwotp}[1]{\norm{#1}_{2, \treatedP}} % l2 norm

We now move away from well-specified parametric approaches, and consider
nonparametric sieve estimators that take an increasing sequence
$\funcspace_n \subset \funcspace_{n+1} \subset \cdots \subset \funcspace\opt$ of
spaces of functions as the model class in the empirical optimization
problem~\eqref{eqn:opt-emp}~\citep{GemanHw82}. We let $\funcspace\opt$ be the
(nonparametric) set of suitably smooth functions, and only require that
$\pmone \in \funcspace\opt$, allowing very general class of functions and
misspecified model classes $\funcspace_n$. By appropriately choosing the
approximation space $\funcspace_n$, which we call \emph{sieves}, we can
provide estimation guarantees required in Assumption~\ref{assumption:neyman}
whenever $\pmone$ is smooth enough. We refer the reader to~\citet{Chen07} for
a detailed overview of sieve estimators.

For concreteness, we consider the following sieve spaces.\\
\begin{example}[Polynomials]
  \label{example:polynomials}
  Let $\pol{\comp}$ be the space of $\comp$-th order polynomials on  $[0, 1]$
  \begin{equation*}
    \pol{\comp} \defeq
    \left\{ [0,1] \ni x \mapsto \sum_{k=0}^{\comp} a_k x^k
    : a_k \in \R \right\},
  \end{equation*}
  and let the sieve be
  $\funcspace_n \defeq \left\{ x \mapsto \Pi_{k=1}^d f_k(x_k) \mid f_k \in
    \pol{J_n}, k = 1,\dots,d \right\},$ for $J_n \to \infty$.
\end{example}

\begin{example}[Splines]
  \label{example:splines}
  Consider knots $0 = t_0 < \ldots < t_{\comp+1} = 1$ such that
  \begin{equation*}
    \frac{\max_{0 \le j \le \comp} (t_{j+1} - t_j)}
    {\min_{0  \le j \le \comp} (t_{j+1} - t_j)}
    \le c
  \end{equation*}
  for some $c > 0$. Let space of $r$-th order splines with $\comp$ knots be
  \begin{equation*}
    \spl{r}{\comp} \defeq
    \left\{ x \mapsto
    \sum_{k=0}^{r-1} a_k x^k + \sum_{j=1}^\comp b_j \hinge{x-t_j}^{r-1}, x\in [0, 1]:
    a_k, b_k \in \R
    \right\},
  \end{equation*}
  and the corresponding sieve
  $\funcspace_n \defeq \left\{ x \mapsto f_1(x_1)f_2(x_2)\dots f_d(x) \mid f_k
    \in \spl{r}{J_n}, k = 1,\dots,d \right\}$ for some integer
  $r \ge \floor{\holdersmooth} + 1$ and $J_n \to \infty.$
\end{example}
\hspace{1pt} \\ \noindent The empirical approximation~\eqref{eqn:opt-emp} over
$\mu_1 \in \funcspace_n$ is a convex optimization problem when $\funcspace_n$
is a finite dimensional linear space, as in Examples~\ref{example:polynomials}
or \ref{example:splines}.

We also consider neural networks with one hidden layer, without requiring that
the model class is well-specified as in Section~\ref{section:example-param}. \\
\begin{example}[Neural network with one hidden layer]
  \label{example:nn}
  Consider neural networks with one hidden layer, with the sigmoid activation
  function $v(a) = (1+\exp^{-a})^{-1}$
  \begin{align*}
    \funcspace_n \defeq \left\{ x\mapsto
    a_0 + \sum_{j=1}^{J_n} a_j v(b_j^\top x + b_{0, j}):
    a_j, b_{0, j} \in \R, b_j \in \R^d,
    \max\left( \sum_{j=0}^{J_n} |a_j|,
    \max_{1\le j\le J_n} \sum_{i=0}^d |b_{ij}| \right) \le C
    \right\},
  \end{align*}
  for some $C> 0$ and $J_n \to \infty$.
\end{example}

To establish asymptotic convergence of $\hmone$, we need some regularity
conditions. First, we assume a condition that allows control of supremum norm
errors using the $L^2(P)$-norm. This is an important requirement used to show
convergence~\citep{Chen07} of sieve estimators.
\begin{assumption}
  \label{assumption:density}
  Let $P_{\covariaterv | \treatmentrv = 1}$ have 
  density $\treatedp(x)$ with respect to the Lebesgue measure, such that
  $0 < \inf_{\covariate \in \covspace} \treatedp(\covariate) \le
  \sup_{\covariate \in \covspace} \treatedp(\covariate) < \infty$.
\end{assumption}

For the first two examples, we will assume that $\pmone$ belong in a
H\"{o}lder class. Recall that the H\"{o}lder class
$\holderball{\holdersmooth}{\holderradius}$ of $\holdersmooth$-smooth
functions for $\holdersmooth_1 = \ceil{\holdersmooth} - 1$ and
$\holdersmooth_2 = \holdersmooth - \holdersmooth_1$ is given by
\begin{equation*}
  \holderball{\holdersmooth}{\holderradius}
  \defeq \left\{ h \in C^{\holdersmooth_1}(\covspace):
  \sup_{\tiny \begin{array}{c}x\in \mc{X} \\ \sum_{l=1}^d \alpha_l<p_1 \end{array}}|D^\alpha h(x)| +
  \sup_{\tiny \begin{array}{c} x \neq x' \in \mc{X} \\ \sum_{l=1}^d \beta_l=p_1\end{array}}\frac{|D^\beta h(x) - D^\beta h(x')|}{\norm{x - x'}^{\holdersmooth_2}}
  \le \holderradius.
  \right\},
\end{equation*}
where $C^{\holdersmooth_1}(\covspace)$ denotes the space of
$\holdersmooth_1$-times continuously differentiable functions on $\covspace,$
and
$D^{\alpha} = \frac{\partial^{\alpha}} {\partial^{\alpha_1} \ldots
  \partial^{\alpha_\covdim}}$, for any $d$-tuple of nonnegative integers
$\alpha = (\alpha_1, \ldots, \alpha_\covdim)$.
\begin{assumption}
  \label{assumption:smooth}
  Let $\covspace = \covspace_1 \times \cdots \times \covspace_\covdim$ be the
  Cartesian product of compact intervals
  $\covspace_1, \ldots, \covspace_\covdim$, and assume
  $\pmone \in \holderball{\holdersmooth}{\holderradius} \eqdef \funcspace\opt$ for
  some $\holderradius > 0$. 
\end{assumption}
\noindent The assumption
$\pmone \in \holderball{\holdersmooth}{\holderradius}$ allows general
functions $\pmone$ while ensuring it is well-approximated by finite
dimensional linear sieves. In particular, sieve spaces in
Examples~\ref{example:polynomials} and~\ref{example:splines} achieve
approximation error
$\inf_{\mu_1 \in \funcspace_n} \linf{\mu_1 - \popfunc} =
O(\comp_n^{-\holdersmooth})$ (see, e.g.,~\cite[Sec.~5.3.1]{Timan63} or
\cite[Thm.~12.8]{Schumaker07}).  Similar guarantees also hold for wavelet
bases (as well as others)~\citep{Daubechies92, Chen07}, though we omit it for
brevity.

By choosing $J_n$ to optimally trade off statistical estimation error and
approximation error, we can achieve optimal nonparametric convergence rates.
The following theorem is a straightforward consequence of~\citet{ChenSh98}.
\begin{lemma}[{\citet{ChenSh98}}]
  \label{lemma:sieve}
  For $\covspace = [0, 1]^d$, consider $\funcspace_n$ in
  Example~\ref{example:polynomials} or~\ref{example:splines} with
  $\comp_n \asymp n^{\frac{1}{2 \holdersmooth + \covdim}}$. Let
  Assumptions~\ref{assumption:residuals},~\ref{assumption:smooth},~\ref{assumption:density}
  hold, and let $\hmone$ be a
  $O_p\left(\left(\frac{\log n}{n}\right)^\frac{2\holdersmooth}{2\holdersmooth
      + \covdim}\right)$-approximate optimizer of the empirical
  problem~\eqref{eqn:opt-emp}. Then, we have
  $\norms{\empfunc - \popfunc}_{2, P} = O_p\left(\left(\frac{\log
        n}{n}\right)^\frac{\holdersmooth}{2\holdersmooth + \covdim}\right)$.
\end{lemma}
\noindent The above result states that when $p > d$, that is when $\pmone$ is
sufficiently smooth relative to the dimension, relevant requirements in
Assumption~\ref{assumption:neyman} are satisfied.

For neural networks with one hidden layer, we have the following result which
builds out of universal approximation guarantees shown by~\citet{Barron93} and
\citet{Makovoz96}.
\begin{lemma}[{\citet{ChenSh98}}]
  \label{lemma:sieve-nn}
  Let Assumptions~\ref{assumption:residuals},~\ref{assumption:density} hold,
  and let $\pmone$ be given by $\pmone(x) = \int \exp(i a^\top x) d\nu(a)$ for
  some complex valued measure $\nu$ on $\R^d$ such that for some $C>0$
  \begin{align*}
    \int \max(\lone{a}, 1) d|\nu|_{\rm tv}(a) \le C,
  \end{align*}
  where $|\nu|_{\rm tv}$ is the total variation of $\nu$. If $\hmone$ is the
  solution to the empirical problem~\eqref{eqn:opt-emp} with the sieve space
  considered in Example~\ref{example:nn}, then
  $\norms{\empfunc - \popfunc}_{2, P} = O_p\left( \left(\frac{\log n}{n}
    \right)^{\frac{1+d^{-1}}{4(1+(2d)^{-1})}} \right)$.
\end{lemma}

%%% local Variables:
%%% mode: latex
%%% TeX-master: "main"
%%% End:

\section{Proof of Theorem~\ref{theorem:clt}: Part I}
\label{section:proof-clt-classical}

In this section, we prove that for each $k \in [K]$,
\makesmall{
\begin{align}
  \label{eqn:conv-classical}
  \sqrt{|I_k|} \left( T\left(\what{P}_k; \hall\right) - T\left( P; \hall\right)
  \right) = \frac{1}{\sqrt{|I_k|}} \sum_{i \in \fold} \psi(D_i) + o_p(1).
\end{align}
}
In the rest of the proof, let $\cfoldinf$ be the set of indices \emph{not} in
$I_k$, as $n \to \infty$.

We begin by showing that the feasibility region in the dual formulation of the
$\mbox{WTE}_{\alpha}$ can be restricted to a compact set.  Let $S_{\alpha}$ be
an interval around $\aq{\pmu}$
\begin{equation*}
  S_{\alpha} \defeq [\aq{\pmu} \pm 1].
\end{equation*}
\begin{proposition}
  \label{proposition:cvar-domain}
  Under the conditions of Theorem~\ref{theorem:clt},
  following occurs eventually with probability $1$
  \makesmall{
  \begin{align*}
    \argmin_{\eta} \left\{ \frac{1}{\alpha}
    \E_{X \sim \what{P}_k} \hinge{\hmu(X) - \eta} + \eta \right\}
    \subseteq S_{\alpha}
    ~~\mbox{and}~~\argmin_{\eta} \left\{ \frac{1}{\alpha}
    \E_{X \sim P} \hinge{\hmu(X) - \eta} + \eta \right\}
    \subseteq S_{\alpha}.
  \end{align*}
  }
\end{proposition}
\noindent See Appendix~\ref{section:proof-cvar-domain} for a proof of
Proposition~\ref{proposition:cvar-domain}.

Define a modified version of the functional~\eqref{eqn:functional} where
inf is taken over $S_{\alpha}$ instead of $\R$
\makesmall{
\begin{align}
  \label{eqn:cpt-functional}
  T_{S_{\alpha}} (Q; \mu_0, \mu_1, e, h)
  \defeq \inf_{\eta \in S_{\alpha}} \left\{ \frac{1}{\alpha}
  \E_{D \sim Q}\hinge{\mu(X) - \eta} + \eta
  \right\} + \E_{D \sim Q}[\kappa(D; \mu_0, \mu_1, e, h)].
\end{align}
}
From Proposition~\ref{proposition:cvar-domain}, the following event happens
almost surely
\makesmall{
\begin{subequations}
  \label{eqn:cpt-domain-ok}
\begin{align}
  \inf_{\eta \in S_{\alpha}} \left\{ \frac{1}{\alpha}
  \E_{X \sim \what{P}_k}\hinge{\hmu(X) - \eta} + \eta \right\}
  & = \inf_{\eta} \left\{ \frac{1}{\alpha}
  \E_{X \sim \what{P}_k} \hinge{\hmu(X) - \eta} + \eta \right\}
  ~~\mbox{eventually} \label{eqn:cpt-domain-emp} \\
  \inf_{\eta \in S_{\alpha}} \left\{ \frac{1}{\alpha}
  \E_{X \sim P}\hinge{\hmu(X) - \eta} + \eta \right\}
  & = \inf_{\eta} \left\{ \frac{1}{\alpha}
  \E_{X \sim P} \hinge{\hmu(X) - \eta} + \eta \right\}
    ~~\mbox{eventually}
    \label{eqn:cpt-domain-pop}
\end{align}
\end{subequations}
}
For the claim~\eqref{eqn:conv-classical}, it hence suffices to show
\ifdefined\useectastyle
{\small
\begin{align}
  \label{eqn:cpt-conv-classical}
  \hspace{-20pt}\sqrt{|I_k|} \left( T_{S_{\alpha}}\left(\what{P}_k; \hall\right) - T_{S_{\alpha}}\left( P; \hall\right)
  \right) = \frac{1}{\sqrt{|I_k|}} \sum_{i \in \fold} \psi(D_i) + o_p(1).
\end{align}
}
\else
\begin{align}
  \label{eqn:cpt-conv-classical}
  \sqrt{|I_k|} \left( T_{S_{\alpha}}\left(\what{P}_k; \hall\right) - T_{S_{\alpha}}\left( P; \hall\right)
  \right) = \frac{1}{\sqrt{|I_k|}} \sum_{i \in \fold} \psi(D_i) + o_p(1).
\end{align}
\fi

Recalling the definition of $\mc{U}$ in the paragraph preceding
Assumption~\ref{assumption:regularity}, define the event
\begin{align}
  \label{eqn:good-event}
  \event_{n, k} \defeq
  \left\{ \hmu \in \mc{U}, \mbox{  and conditions of Assumption~\ref{assumption:neyman} holds for $k$}
  \right\}.
\end{align}
In what follows, we show convergence~\eqref{eqn:cpt-conv-classical}
conditional on $\event_{n, k}$. This conditional convergence implies the
unconditional result~\eqref{eqn:cpt-conv-classical}: for any sequence of
random variables $W_n$ satisfying $\P(W_n > \delta \mid \event_{n, k}) \to 0$,
we have $\P(W_n > \delta) \to 0$ since $\P(\event_{n, k}) \to 1$ by
Assumptions~\ref{assumption:neyman},~\ref{assumption:regularity}.  

We begin by showing that the empirical measure $\what{P}_k$ weakly converges
uniformly (at the $\sqrt{n}$-rate) over the following set of functions
\begin{subequations}
  \label{eqn:functions}
  \begin{align*}
    & f_{n, \eta}(D)
     \defeq 
       \frac{1}{\alpha} \hinge{\hmu(X) - \eta} + \eta 
      ~\mbox{for}~ \eta \in S_{\alpha},\\
    & f_{n, \aq{\pmu} + 2}(D)
      \defeq \kappa(D; \hall),
  \end{align*}
\end{subequations}
recalling that $S_{\alpha} \defeq [\aq{\pmu} \pm 1]$.  The above functions are
identified by elements of the index set
$\Lambda \defeq S_{\alpha} \cup \{\aq{\pmu} + 2\}$.  We denote by
$\ell^{\infty}(\Lambda)$ the space of uniformly bounded functions on $\Lambda$
endowed with the supremum norm, and view measures as bounded functionals on
$\Lambda$ so that $\what{P}_k: \eta \mapsto \E_{D \sim \what{P}_k} f_{n, \eta}(D)$ and
$P_n: \eta \mapsto \E_{D \sim P} f_{n, \eta}(D)$.  We have the following key result, which we
prove in Appendix~\ref{section:proof-donsker}. 
\begin{proposition}
  \label{prop:donsker}
  Under the conditions of Theorem~\ref{theorem:clt}, conditional on
  $\event_{n,k}$,
  \begin{align*}
    \sqrt{n} \left(\E_{D \sim \what{P}_k} f_{n, \eta}(D)
    - \E_{D \sim P} f_{n, \eta}(D)\right)
    \cd \mathbb{G} ~~\mbox{in}~~\ell^{\infty}\left(\Lambda\right),
  \end{align*}
  where $\mathbb{G}$ is a Gaussian process on
  $\Lambda = S_{\alpha} \cup \{\aq{\pmu} + 2\}$ with covariance
  $\Sigma(\eta, \eta')$
  \makesmall{
  \begin{align*}
    & \frac{1}{\alpha^2} \E\left[ \hinge{\pmu(X) - \eta}
      \hinge{\pmu(X) - \eta'} \right]
      + \frac{\eta'}{\alpha} \E\left[ \hinge{\pmu(X) - \eta}  \right]
      + \frac{\eta}{\alpha} \E\left[ \hinge{\pmu(X) - \eta'}  \right]
      ~ \mbox{if}~\eta, \eta' \in S_{\alpha} \\
    & \frac{1}{\alpha}
      \E\left[ \hinge{\pmu(X) - \eta} \kappa(D; \pall) \right]
      ~\mbox{if}~\eta \in S_{\alpha}, \eta' = \aq{\pmu} + 2 \\
    & \E\left[ \kappa(D; \pall)^2 \right] ~\mbox{if}~\eta = \eta' = \aq{\pmu} + 2.
  \end{align*}
  }
\end{proposition}

Next, we apply the functional delta method to the map
$Q \mapsto T_{S_{\alpha}}(Q; \hall)$ by establishing uniform Hadamard
differentiability of the functional.
% Let $(\mc{D}, \mathcal{A})$ be a measurable space, and $\hclass$ be a
% collection of functions $h : \mc{D} \to \R$, where we assume that
% $\hclass$ is $P_0$-Donsker with envelope $M_2 \in L^2(P_0)$
% (Definition~\ref{definition:donsker}).
We begin by formally recalling the functional delta method. Let
$\lambda: \mathbb{D}_{\lambda} \subset \mathbb{D} \to \R$ be a functional on a
metrizable topological vector space $\mathbb{D}$ and its (arbitrary) subset
$\mathbb{D}_{\lambda}$. Let $r_n$ be a sequence of constants such that
$r_n \to \infty$, and let $P_n, P$ be elements of
$\mathbb{D}_{\lambda} \subset \mathbb{D}$ such that $P_n \to P$.  In the
result below, the sets $\Omega_n$ are sample spaces defined for each $n$.
\begin{lemma}[{\citep[Delta method, Theorem 3.9.5]{VanDerVaartWe96}}]
  \label{lemma:delta}
  Let $\mathbb{D}_0 \subseteq \mathbb{D}$. For every converging sequence
  $H_n \in \mathbb{D}$ such that $P_n + r_n^{-1} H_n \in \mathbb{D}_{\lambda}$
  ~for all $n$, and $H_n \to H \in \mathbb{D}_0 \subset \mathbb{D}$, let there
  be a map $d\lambda_P(\cdot)$ on $\mathbb{D}_0$ such that
  \begin{align*}
    r_n(\lambda(P_n + r_n^{-1} H_n) - \lambda(P_n)) \to d\lambda_P(H).
  \end{align*}
  Let $\xi_n: \Omega_n \to \mathbb{D}_{\lambda}$ be maps with
  $\sqrt{n} (\xi_n - P_n) \cd \xi$ in $\mathbb{D}$, where $\xi$ is separable
  and takes values in $\mathbb{D}_0$. If $d\lambda_P(\cdot)$ can be extended
  to the whole of $\mathbb{D}$ as a linear, continuous map, then
  \begin{align*}
    r_n(\lambda(\xi_n) - \lambda(P_n)) - d\lambda_P(r_n(\xi_n - P)) \cp 0.
  \end{align*}
\end{lemma}

Our goal is to apply Lemma~\ref{lemma:delta} to the functinoal
$\lambda = T_{S_{\alpha}}$, where $\mathbb{D} = \ell^{\infty}(\Lambda)$,
$r_n = \sqrt{|\fold|}$,
$\xi_n = \what{P}_k: \eta \mapsto \E_{\what{P}_k} f_{n, \eta}$, and
$P_n: \eta \mapsto \E_P f_{n, \eta}$.  We first show that the infimization
functional
\begin{equation*}
  \lambda_{\rm opt}(H) \defeq \inf_{\eta \in S_{\alpha}} H(\eta)
\end{equation*}
satisfies uniform Hadamard differentiability required in
Lemma~\ref{lemma:delta}.  Let $\mathbb{D}_{\lambda_{\rm opt}}$ be the set of
functions
\begin{align*}
  \eta \mapsto
  \begin{cases}
    \frac{1}{\alpha} \E_{Q}\hinge{\mu(X) - \eta} + \eta
    & ~~\mbox{if}~\eta \in S_{\alpha} \\
    \E_{Q}[\kappa(D; \mu_0, \mu_1, e, h)] 
    & ~~\mbox{if}~\eta = \aq{\pmu} + 2
  \end{cases}
\end{align*}
such that $Q$ is a probability on $\mc{D}$, $\E[\mu^2(X)] < \infty$,
$e(\cdot) \in [c, 1-c]$, and $|h| \le M_h$. We interpret $P$ as a element of
this set with $\mu = \pmu$. In the below lemma, define the set
\begin{equation*}
  \mathbb{D}_0 \defeq \left\{ H \in \ell^{\infty}(\Lambda):
    \eta \mapsto H(\eta)~\mbox{is continuous} \right\}.
\end{equation*}
\begin{lemma}
  \label{lemma:danskin}
  Assume that the hypothesis of Theorem~\ref{theorem:clt} holds.
  On the event $\event_{n, k}$,
  $\lambda_{\rm opt}: \mathbb{D}_{\lambda_{\rm opt}} \subset
  \ell^{\infty}(\Lambda) \to \R$ satisfies the following: for every converging
  sequence $H_n \in \ell^{\infty}(\Lambda)$ s.t.
  $P_n + |I_k|^{-1/2} H_n \in \mathbb{D}_{\lambda_{\rm opt}}$ for all $n$, and
  $H_n \to H \in \mathbb{D}_0$,
  \begin{align*}
    \sqrt{|I_k|} (\lambda_{\rm opt}(P_n + |I_k|^{-1/2} H_n) - \lambda_{\rm opt}(P_n))
    \to H(\aq{\pmu}) \eqdef d\lambda_{{\rm opt},P}(H).
  \end{align*}
\end{lemma}
\noindent We prove the lemma in Appendix~\ref{section:proof-danskin}.

Since there is an almost surely equivalent version of the Gaussian process
$\mathbb{G}$ (given in Proposition~\ref{prop:donsker}) that have continuous
sample paths, we can assume $\mathbb{G}$ takes values in $\mathbb{D}_0$
without loss of generality. Recalling the definition
\begin{align*}
  T_{S_{\alpha}}(\what{P}_k; \hall) & = \lambda_{\rm opt}(\what{P}_k) + \E_{D \sim
                                      \what{P_k}}[\kappa(D; \hall)], \\
  T_{S_{\alpha}}(P_n; \hall) & = \lambda_{\rm opt}(P_n)
                               + \E_{D \sim P}[\kappa(D; \hall)],
\end{align*}
Lemma~\ref{lemma:danskin} confirms the hypothesis of Lemma~\ref{lemma:delta}.
Thus, we have shown that conditional on $\event_{n, k}$, the
convergence~\eqref{eqn:cpt-conv-classical} holds. As argued above, this shows
our final claim~\eqref{eqn:conv-classical}.

\subsection{Proof of Proposition~\ref{proposition:cvar-domain}}
\label{section:proof-cvar-domain}

We use the following elementary result, which is essentially known
(e.g.,~\citet{RockafellarUr00}).
% though the proof becomes more involved, the lemma follows only from assuming F_{\xi}(\aq{\xi}) = 1-\alpha
\begin{lemma}
  \label{lemma:cvar-soln}
  If a random variable $\xi$ has a positive density at the
  $(1-\alpha)$-quantile $\aq{\xi} \defeq \inf\{t: F_{\xi}(t) \ge 1-\alpha\}$,
  then
  \begin{equation*}
    \argmin_{\eta} \left\{ \frac{1}{\alpha} \E\hinge{\xi - \eta} + \eta
      \right\} = \left\{ P^{-1}_{1-\alpha}(\xi)\right\}.
  \end{equation*}
\end{lemma}
\begin{proof-of-lemma}
  Let $F_{\xi}$ be the cumulative distribution of $\xi$.
  From first order optimality conditions, $\eta\opt$ is an optimum if
  $F_{\xi}(\eta\opt) - (1-\alpha) \in [0, \P(\xi = \eta\opt)]$. Since
  $F_{\xi}(P^{-1}_{1-\alpha}(\xi)) = 1-\alpha$ by hypothesis,
  $P^{-1}_{1-\alpha}(\xi)$ is an optimal solution. To see that this solution
  is unique, any optimal solution $\eta\opt$ cannot be smaller than
  $P^{-1}_{1-\alpha}(\xi)$ since it violates the first order optimality
  condition (recall the definition of the quantile $P^{-1}_{1-\alpha}(\xi)$).

  Now, assume that an optimal solution $\eta\opt$ satisfies
  $\eta\opt > P^{-1}_{1-\alpha}(\xi)$. Since $\xi$ has a positive density at
  $\aq{\xi}$, we have $F_{\xi}(\eta\opt) > 1-\alpha$, and
  $\P(\xi = \eta\opt) > 0$ by first order optimality conditions.  By
  convexity, for all $a \in [0, 1]$,
  $a \eta\opt + (1-a) P^{-1}_{1-\alpha}(\xi)$ is an optimal solution; the same
  argument again gives
  $\P(\xi = a \eta\opt + (1-a) P^{-1}_{1-\alpha}(\xi)) > 0$ for all
  $a \in [0, 1]$. This implies that $F_{\xi}$ has an uncountable number of
  jumps, which gives a contradiction since $F_{\xi}$ is a cumulative
  distribution functionx (and hence has at most countable jumps).
\end{proof-of-lemma}
Applying Lemma~\ref{lemma:cvar-soln} to $\xi = \pmu(X)$, we have 
\begin{align*}
 \{ \aq{\pmu} \} = \inf_{\eta} \left\{ \frac{1}{\alpha}
  \E \hinge{\pmu(X) - \eta} + \eta \right\}.
\end{align*}
We proceed by arguing that the solutions to the optimization problems
\begin{align*}
  \argmin_{\eta} \left\{ \frac{1}{\alpha}
    \E_{X \sim \what{P}_k}\hinge{\hmu(X) - \eta} + \eta \right\},
   ~~~~\argmin_{\eta} \left\{ \frac{1}{\alpha}
  \E_{X \sim P}\hinge{\hmu(X) - \eta} + \eta \right\}
\end{align*}
converge to its population counterpart $\aq{\pmu}$. The following result is a direct consequence of the powerful
theory of epi-convergence~\citep{RockafellarWe98}.
\begin{lemma}[{\citet[Theorems 7.17, 7.31]{RockafellarWe98}}]
  \label{lemma:epi}
  Let $g_n, g: \R \to \R$ be proper, closed, convex, and coercive functions,
  and let $\argmin_{\eta} g(\eta) = \{\eta\opt\}$ be unique. If $g_n \to g$
  pointwise, then
  $\sup_{\eta \in \argmin_{\eta'} g_n(\eta')} |\eta - \eta\opt| \to 0$.
\end{lemma}
\noindent To prove our two claims, we apply Lemma~\ref{lemma:epi} with
$g(\eta) \defeq \frac{1}{\alpha} \E\hinge{\pmu(X) - \eta} + \eta$ and
\begin{align*}
  \what{g}_{1, n, k}(\eta) \defeq
  \frac{1}{\alpha}
  \E_{X \sim \what{P}_k}\hinge{\hmu(X) - \eta} + \eta,~~~
    \what{g}_{2, n, k}(\eta) \defeq
  \frac{1}{\alpha}
  \E_{X \sim P}\hinge{\hmu(X) - \eta} + \eta.
\end{align*}
To verify the hypothesis of Lemma~\ref{lemma:epi}, first note that
$\what{g}_{1, n, k}, \what{g}_{2, n, k}, g$ are all proper, continuous,
convex, and coercive, and $g$ has a unique optimum from
Lemma~\ref{lemma:cvar-soln}. Assumption~\ref{assumption:neyman}
implies that $\what{g}_{2, n, k} \to g$ pointwise, which gives our second result.

We now show $\what{g}_{1, n, k} \cas g$ pointwise, where we begin with the
bound
\begin{align}
  |\what{g}_{1, n, k}(\eta) - g(\eta)|
  \le & \frac{1}{\alpha} \left| \E_{X \sim \what{P}_k} \hinge{\hmu(X) - \eta}
  - \E_{X \sim P} \hinge{\hmu(X) - \eta} \right| \nonumber \\
  & + \frac{1}{\alpha} \left| \E_{X \sim P} \hinge{\hmu(X) - \eta}
  - \E_{X \sim P} \hinge{\pmu(X) - \eta} \right|.   \label{eqn:g-conv}
\end{align}
The second term in the right hand side goes to zero pointwise from
Assumption~\ref{assumption:neyman}.  To show that the first term converges to
zero, we use the following strong law of large numbers for triangular arrays.
\begin{lemma}[{\citet[Theorem 2]{HuMoTa89}}]
  \label{lemma:triangular-slln}
  Let $\{\xi_{ni}\}_{i=1}^n$ be a triangular array where
  $X_{n1}, X_{n2}, \ldots$ are independent random variables for any fixed
  $n$. If there exists a real-valued random variable $\xi$ such that
  $|\xi_{ni}|\le\xi$ and $\E[\xi^2] <\infty$, then
  $\frac{1}{n} \sum_{i=1}^n (\xi_{ni} - \E[\xi_{ni}]) \cas 0$.
\end{lemma}
\noindent Conditional on $\{D_i\}_{i \in \cfoldinf}$, we can apply
Lemma~\ref{lemma:triangular-slln} to
$\{\hinge{\hmu(X_i) - \eta}\}_{i \in \fold}$ since each element is mutually
independent conditional on $\{D_i\}_{i \in \cfoldinf}$. For any $\eta \in \R$, we
conclude the first term in the bound~\eqref{eqn:g-conv} converges to zero
almost surely conditional on $\{D_i\}_{i \in \cfoldinf}$.
By dominated convergence, it follows that the first term in the
bound~\eqref{eqn:g-conv} goes to zero almost surely.

% In what follows, we show
% convergence~\eqref{eqn:conv-classical} conditional on
% $\sigma(\{D_i\}_{i \in \cfoldinf})$ and $\event_{n, k}$. Since
% $\P(\event_{n, k}) \to 1$ by
% Assumptions~\ref{assumption:neyman},~\ref{assumption:regularity}, this
% conditional convergence implies the desired result~\eqref{eqn:conv-classical}
% : by dominated convergence, $\P(W_n \le t \mid \mc{F}) \to \P(N(0, 1) \le t)$
% for all $t \in \R$ implies $\P(W_n \le t) \to \P(N(0, 1) \le t)$ for all
% $t \in \R$.

\subsection{Proof of Proposition~\ref{prop:donsker}}
\label{section:proof-donsker}

We first recall the (standard) definition of bracketing numbers, which measure
the size of a set of functions $\mc{F} \subseteq \{\mc{D} \to \R\}$ by the
number of brackets that cover it. (Recall that
$\mc{D} = \mc{X} \times \mc{Y} \times \{0, 1\}$ is the space that observations
$D_i$ take value in.)
\begin{definition}
  \label{def:bracketing}
  Let $\norm{\cdot}$ be a (semi)norm on $\hclass$.  For functions
  $l, u : \mc{D} \to \R$ with $l \le u$, the \emph{bracket} $[l, u]$ is
  the set of functions $f : \mc{D} \to \R$ such that $l \le h \le u$, and
  $[l, u]$ is an \emph{$\epsilon$-bracket} if $\norm{l - u} \le \epsilon$.
  Brackets $\{[l_i, u_i]\}_{i = 1}^m$ \emph{cover} $\hclass$ if for all
  $f \in \hclass$, there exists $i$ such that $f \in [l_i, u_i]$. The
  \emph{bracketing number}
  $N_{[\hspace{1pt}]}(\epsilon, \hclass, \norm{\cdot})$ is the minimum number
  of $\epsilon$-brackets needed to cover $\hclass$.
\end{definition}
We use a variant of an uniform version of the Lindeberg-Feller central limit
theorem, which relies on the definition of bracketing numbers; we refer the
reader to~\citet[Chapter 2.11]{VanDerVaartWe96} for an extensive treatment of
related results.  For any set $\mc{V}$, we let
$\ell^{\infty}(\mc{V})$ the space of uniformly bounded functions on $\mc{V}$;
we will identify probability measures as elements of
$\ell^\infty(\mc{V})$. Recall that a sequence of stochastic processes $\mathbb{G}_n$
taking values in $\ell^\infty(\mc{V})$ is said to be asymptotically tight if
for every $\epsilon_1 > 0$ there exists a compact
$K \subseteq \ell^\infty(\mc{V})$ such that
\begin{equation*}
  \liminf_{n \to \infty} \P(\mathbb{G}_n \in K^{\epsilon_2}) \ge 1-\epsilon_1
  ~~\mbox{for all}~\epsilon_2 > 0,
\end{equation*}
where $K^\epsilon \defeq \{f \in \ell^\infty(\mc{V}): d(f, K) \le \epsilon\}$
is the $\epsilon$-enlargement of $K$ (in the uniform norm).

In the below lemma, we denote by $\emp$ the empirical measure
on $n$ i.i.d. observations; abusing notation, we take samples over $I_k$ in
our subsequent application. 
\begin{lemma}[{\citet[Theorem 2.11.23]{VanDerVaartWe96}}]
  \label{lemma:triangular-donsker}
  For each $n$, let $\hclass_n \defeq \{f_{n, v}: v \in \mc{V}\}$ be a class
  of measurable functions $D \mapsto \R$ indexed by a totally bounded
  semi-metric space $(\mc{V}, d)$. Let there exist an envelope functions $B_n$
  such that $|f_{n, v}| \le B_n$ for all $v \in \mc{V}$, and
  \begin{subequations}
    \label{eqn:nice-envelope}
    \begin{align}
      & \E[ B_n^2] = O(1), \label{eqn:nice-envelope-var} \\
      & \E[B_n^2 \indic{B_n > M\sqrt{n}}] \to 0~~\mbox{for all}~~M > 0, \label{eqn:nice-envelope-dct} \\
      & \sup_{d(v, v') < \delta_n} \E[(f_{n, v} - f_{n, v'})^2] \to 0 ~~\mbox{for any}~~
      \delta_n \downarrow 0. \label{eqn:nice-envelope-conti}
    \end{align}
  \end{subequations}
  If the following bracketing integral goes to zero for any
  $\delta_n \downarrow 0$
  \begin{equation}
    \label{eqn:bracketing-integral}
    \int_0^{\delta_n} \sqrt{\log N_{[\hspace{1pt}]}
      \left(\epsilon\norm{B_n}_{L^2(\mc{D})}, \mc{F}_n, L^2(\mc{D})\right)} d\epsilon
    \to 0,
  \end{equation}
  then $\sqrt{n} \left(\E_{\emp} [f_{n, v}] - \E[f_{n, v}]\right)$ is
  asymptotically tight in $\ell^{\infty}(\mc{V})$.
  % and $\forall v, v' \in \mc{V}$,
  % $\E[f_{n, v} f_{n, v'}] - \E[f_{n, v}] \E[f_{n, v'}]$ converges pointwise to
  % a covariance $\Sigma(v, v')$, then
  % \begin{align*}
  %   \sqrt{n} \left(\E_{\emp} [f_{n, v}] - \E[f_{n, v}]\right) \cd \mathbb{G}~~\mbox{in}~~\ell^{\infty}(\mc{V}),
  % \end{align*}
  % for some centered Gaussian process $\mathbb{G}$ with covariance given by
  % $\Sigma(\cdot, \cdot)$.
\end{lemma}

Recalling the definition~\eqref{eqn:functions}, let $\mc{F}_n$ be the space of
functions
\begin{align*}
  \mc{F}_n \defeq \left\{f_{n, \eta} \indic{\event_{n, k}} \mid \eta \in \Lambda
  \defeq S_{\alpha} \cup \{\aq{\pmu} + 2\}   \right\}.
\end{align*}
We will apply Lemma~\ref{lemma:triangular-donsker} conditional on
$\{D_i\}_{i \in \cfoldinf}$, with
$\mc{V} = \Lambda = S_{\alpha} \cup \{\aq{\pmu} + 2\}$. Conditional on
$\{D_i\}_{i \in \cfoldinf}$, $\{\hmu(X_i)\}_{i \in \fold}$ and
$\{\kappa(D_i; \hall)\}_{i \in \fold}$ are respectively i.i.d., since
$\event_{n, k}$ is $\sigma(\{D_i\}_{i \in \cfoldinf})$-measurable.
% \indic{\event_{n, k}}  $S_{\alpha} \defeq [\aq{\pmu} \pm 1]$, define
% \begin{align*}
%   f_{n, \eta}(D) \defeq \frac{1}{\alpha} \hinge{\hmu(X) - \eta} + \eta
%   ~\mbox{for}~ \eta \in S_{\alpha},~~\mbox{and}~~~
%   f_{n, \aq{\pmu} + 2}(D) \defeq \kappa(D; \hall).
% \end{align*}
To verify the hypothesis of Lemma~\ref{lemma:triangular-donsker}, we first
check that there is an envelope function satisfying the
conditions~\eqref{eqn:nice-envelope}, where we interpret the expectations as
conditional expectations given $\{D_i\}_{i \in \cfoldinf}$.  Noting that on
the event $\event_{n, k}$,
$|\kappa(D; \hall)| \le \frac{M_h}{c} (2|Y| + \bar{\mu}(X))$, define the
envelope function
\begin{equation*}
  B \defeq B_n \defeq \alpha^{-1} \bar{\mu}(X) + (1+ \alpha^{-1}) (|\aq{\pmu}| + 2)
  + \frac{M_h}{c} (2|Y| + \bar{\mu}(X)).
\end{equation*}
By inspection,
conditions~\eqref{eqn:nice-envelope-var},~\eqref{eqn:nice-envelope-dct} hold,
and since $\E[(f_{n, \eta} - f_{n, \eta'})^2] \le (\alpha^{-1} + 1) |\eta - \eta'|^2$ whenever
$|\eta - \eta'| < 1$, condition~\eqref{eqn:nice-envelope-conti} holds.

To verify the bracketing integral condition~\eqref{eqn:bracketing-integral},
we use the following basic result. Let $N(\epsilon, \mc{V}, \norm{\cdot})$ be
the $\epsilon$-covering number of $\mc{V}$, in the metric $\norm{\cdot}$.
\begin{lemma}[{\citet[Chapter 2.7.4]{VanDerVaartWe96}}]
  Let $\hclass = \{f_v: v\in \mc{V}\}$, and let $L(D) > 0$ such that
  $\E[L(D)^2] < \infty$ and $|f_v(D) - f_{v'}(D)| \le L(D)\norm{v - v'}$ for
  all $v, v' \in \mc{V}$. Then,
  $N_{[\hspace{1pt}]} \left(2\epsilon \mbox{diam}(\mc{V}) \norm{L(D)}_{L^2(\mc{D})},
    \mc{F}, L^2(\mc{D})\right) \le N(\epsilon, \mc{V}, \norm{\cdot})$, where
  $\mbox{diam}(\mc{V}) = \sup_{v, v' \in \mc{V}} \norm{v-v'}$.
\end{lemma}
Since $\eta \mapsto \frac{1}{\alpha} \hinge{\hmu(X) - \eta} + \eta$ is
$(\alpha^{-1} + 1)$-Lipschitz, we conclude from the lemma 
\begin{align*}
  \log N_{[\hspace{1pt}]}
  \left(\epsilon, \mc{F}_n, L^2(\mc{D})\right)
  \lesssim \log \frac{1}{\epsilon}
\end{align*}
where $\lesssim$ denotes $\le$ up to a problem dependent constant. Thus,
condition~\eqref{eqn:bracketing-integral} holds, and we have shown that the
stochastic process
\begin{align*}
    \mathbb{G}_n(\eta) \defeq \sqrt{n} \left(\E_{D \sim \what{P}_k} f_{n, \eta}(D)
    - \E_{D \sim P} f_{n, \eta}(D)\right)
\end{align*}
is asymptotically tight as an element of $\ell^{\infty}\left(\Lambda\right)$,
conditional on $\{D_i\}_{i \in \cfoldinf}$ and $\event_{n, k}$. By Fatou's
lemma, this implies that $\mathbb{G}_n$ is asymptotically tight conditional on
the events $\event_{n, k}$.

% define unconditional f_{n, \eta}'s
% since $\event_{n, k} \cp 1$, the unconditional version (define \bar{f} or sth like that) is equivalent to $o_p(1)$
% cite asymptotic tight + marginal then joint lemma
% then say we show marginal conv of unconditional 
% from Lindeberg Feller CLT and Cramer-Wold, suffices to show that covariance converges

% In the definition~\eqref{eqn:functions}, recall that the functions
% $f_{n, \eta}$ were conditioned on the event $\event_{n,k}$. We define its
% unconditional counterpart, and the associated empirical process
% \begin{align*}
%   \bar{f}_{n, \eta}(D)
%   \defeq 
%   \frac{1}{\alpha} \hinge{\hmu(X) - \eta} & + \eta
%   ~\mbox{for}~ \eta \in S_{\alpha},~\mbox{and}~\bar{f}_{n, \aq{\pmu} + 2}(D)
%   \defeq \kappa(D; \hall) \\
%   & \bar{\mathbb{G}}_n(\eta) \defeq \sqrt{n} \left(\E_{D \sim \what{P}_k}
%   \bar{f}_{n, \eta}(D)
%   - \E_{D \sim P} \bar{f}_{n, \eta}(D)\right).
% \end{align*}
We will use the following lemma to prove $\mathbb{G}_n \cd \mathbb{G}$.
\begin{lemma}[Van der Vaart and Wellner~\citep{VanDerVaartWe96}, Theorem 1.5.4]
  \label{lemma:asymptotic-tight-marginal}
  Let $\mathbb{G}_n$ be a sequence of stochastic processes taking values in
  $\ell^\infty(\mc{V})$. Then $\mathbb{G}_n$ converges weakly to a tight limit
  if and only if $\mathbb{G}_n$ is asymptotically tight and the marginals
  $(\mathbb{G}_n(v_1), \ldots, \mathbb{G}_n(v_m))$ converge weakly to a limit
  for every finite subset $\left\{ v_1,\ldots, v_m\right\}$ of $\mc{V}$. If
  $\mathbb{G}_n$ is asymptotically tight and its marginals converge weakly to
  the marginals of $(\mathbb{G}(v_1), \ldots, X(v_k))$ of $\mathbb{G}$, then
  there is a version of $\mathbb{G}$ with uniformly bounded sample paths and
  $\mathbb{G}_n \cd \mathbb{G}$.
\end{lemma}
% \noindent Applying Lemma~\ref{lemma:asymptotic-tight-marginal}, we now show
% that the marginals of $\bar{\mathbb{G}}_n$ converges to that of $\mathbb{G}$.
% To see this, note that
% \begin{align*}
%   \norm{\bar{\mathbb{G}}_n(\eta) - \frac{1}{\sqrt{n}} \sum_{i \in \fold}
%     \left(\bar{f}_{\eta}(D_i)
%     - \E_{D \sim P} [\bar{f}_{\eta}(D)]\right)}_{L^2(\mc{D})}
%     \le \norm{\bar{f}_{n, \eta} - \bar{f}_{\eta}}_{L^2(\mc{D})}
%     + \left| \E_{D \sim P}[\bar{f}_{n, \eta}(D) - \bar{f}_{\eta}(D)]
%     \right|.
% \end{align*}
\noindent To show that the marginals of $\mathbb{G}_n$ converge to that of
$\mathbb{G}$, we use the below result.
% We now verify that the following converges pointwise
% \begin{equation*}
%   \E\left[f_{n, \eta} f_{n, \eta'} \mid \{D_i\}_{i \in \cfoldinf}, \event_{n, k} \right]
%   - \E\left[f_{n, \eta} \mid \{D_i\}_{i \in \cfoldinf}, \event_{n, k}\right]
%   \E\left[f_{n, \eta'} \mid \{D_i\}_{i \in \cfoldinf}, \event_{n, k} \right]
% \end{equation*}
% to the covariance function $\Sigma(\eta, \eta')$ given by
% \begin{align*}
%     & \frac{1}{\alpha^2} \E\left[ \hinge{\pmu(X) - \eta} \hinge{\pmu(X) - \eta'} \right]
%     + \frac{\eta'}{\alpha} \E\left[ \hinge{\pmu(X) - \eta}  \right]
%     + \frac{\eta}{\alpha} \E\left[ \hinge{\pmu(X) - \eta'}  \right]
%     ~ \mbox{if}~\eta, \eta' \in S_{\alpha} \\
%     & \frac{1}{\alpha}
%       \E\left[ \hinge{\pmu(X) - \eta} \kappa(D; \pall) \right] ~\mbox{if}~\eta \in S_{\alpha}, \eta' = \aq{\pmu} + 2 \\
%     & \E\left[ \kappa(D; \pall)^2 \right] ~\mbox{if}~\eta = \eta' = \aq{\pmu} + 2 
% \end{align*}
\begin{lemma}
  \label{lemma:covar-conv}
  Under Assumptions~\ref{assumption:ig}-\ref{assumption:neyman},
  $\forall \eta, \eta' \in S_{\alpha}$, conditional on
  $\{D_i\}_{i \in \cfoldinf}$ and $\event_{n, k}$,
  \begin{align*}
    \E_{X \sim P}\left[ \hinge{\hmu(X) - \eta} \hinge{\hmu(X) - \eta'} \right]
    & \to \E\left[ \hinge{\pmu(X) - \eta} \hinge{\pmu(X) - \eta'}\right], \\
    \E_{D \sim P} \left[\kappa(D; \hall)^2  \right]
    & \to \E\left[\kappa(D; \pall)^2  \right], \\
    \E_{D \sim P} \left[\kappa(D; \hall) \hinge{\hmu(X) - \eta}  \right]
    & \to \E\left[\kappa(D; \pall) \hinge{\pmu(X) - \eta} \right].
  \end{align*}
\end{lemma}
\noindent See Section~\ref{section:proof-covar-conv} for the proof.
From Lemma~\ref{lemma:covar-conv}, Lindeberg-Feller CLT yields
\begin{equation*}
  \sqrt{n} (\E_{D \sim \what{P}_k} f_{n, \eta}(D) - \E_{D \sim P} f_{n, \eta}(D))
  \cd \mathbb{G}(\eta),
\end{equation*}
conditional on $\{D_i\}_{i \in \cfoldinf}$ and $\event_{n, k}$. Since
$\P(\event_{n, k}) \to 1$ by
Assumptions~\ref{assumption:neyman},~\ref{assumption:regularity}, we have that
this convergence holds unconditionally. Using the Cramer-Wold device to show
that all finite dimensional marginals converge, we conclude
$\mathbb{G}_n \cd \mathbb{G}$ from Lemma~\ref{lemma:asymptotic-tight-marginal}.

% Combining the above facts, Lemma~\ref{lemma:triangular-donsker} implies that
% \begin{align}
%   \label{eqn:unif-clt}
%   \sqrt{n} (\E_{D \sim \what{P}_k} f_{n, \eta} - \E_{D \sim P} f_{n, \eta})
%   \cd \mathbb{G} ~~\mbox{in}~~\ell^{\infty}\left(S_{\alpha} \cup \{\aq{\pmu} + 2\}\right),
% \end{align}
% conditional on $\{D_i\}_{i \in \cfoldinf}$ and $\event_{n, k}$. Viewing these
% measures as bounded functionals on $\Lambda$ so that
% $\what{P}_k: \eta \mapsto \what{P}_k f_{n, \eta}$ and
% $P_n: \eta \mapsto P f_{n, \eta}$, the convergence~\eqref{eqn:unif-clt} can be
% written $\sqrt{n}(\what{P}_k - P_n) \cd \mathbb{G}$.

\subsubsection{Proof of Lemma~\ref{lemma:covar-conv}}
\label{section:proof-covar-conv}

By applying Cauchy-Schwarz on the inequality
\begin{align*}
  & |\hinge{\hmu(X) - \eta} \hinge{\hmu(X) - \eta'} - \hinge{\pmu(X) - \eta} \hinge{\pmu(X) - \eta'}| \\
  & \le |\hinge{\hmu(X) - \eta} \hinge{\hmu(X) - \eta'} - \hinge{\hmu(X) - \eta} \hinge{\pmu(X) - \eta'}| \\
  & \qquad + |\hinge{\hmu(X) - \eta} \hinge{\pmu(X) - \eta'} - \hinge{\pmu(X) - \eta} \hinge{\pmu(X) - \eta'}| \\
  & \le \left(\bar{\mu}(X) + |\pmu(X)| + 2|\aq{\pmu}| + 2\right) |\hmu(X) - \pmu(X)|,
\end{align*}
the first limit follows from Assumption~\ref{assumption:neyman} and dominated
convergence.  Under
Assumption~\ref{assumption:ig},~\ref{assumption:overlap},~\ref{assumption:residuals},~\ref{assumption:neyman},
elementary calculations give
\begin{align*}
  & \E_{D \sim P}[|\kappa(D; \pall) - \kappa(D; \hall)|^2] \\
  & \lesssim \Ltwo{\pmzero - \hmzero} + \Ltwo{\pmone - \hmone}
  + \Ltwo{\pprop - \hprop} + \Ltwo{\pthr - \hthr}.
\end{align*}
We show in Section~\ref{section:proof-clt-neyman}, Lemma~\ref{lemma:thr-conv}
that $\Ltwo{\pthr - \hthr} \to 0$ on the event $\event_{n, k}$.  From
Assumption~\ref{assumption:neyman}, it follows that
$\E_{D \sim P}[|\kappa(D; \pall) - \kappa(D; \hall)|^2] \to 0$ conditional on
$\{D_i\}_{i \in \cfoldinf}$ and $\event_{n, k}$, and we have shown the second
limit. For the final result, note that
\begin{align*}
  & \kappa(D; \hall) \hinge{\hmu(X) - \eta} - \kappa(D; \pall) \hinge{\pmu(X) - \eta} \\
  & = \hinge{\hmu(X) - \eta} (\kappa(D; \hall) - \kappa(D; \pall))  \\
  & \qquad  + \kappa(D; \pall)(\hinge{\hmu(X) - \eta} - \hinge{\pmu(X) - \eta}).
\end{align*}
Applying Cauchy-Schwarz and the aforementioned results, we obtain the final
claim.

\subsection{Proof of Lemma~\ref{lemma:danskin}}
\label{section:proof-danskin}
We adapt the approach of~\citet{Romisch05} to show \emph{uniform}
differentiability.  Abusing notation slightly, we use $\eta \mapsto Q(\eta)$
to refer to elements of $\mathbb{D}_{\lambda}$. Denote by $S(F, \epsilon)$ the
set of $\epsilon$-approximate optimizers of $F \in \mathbb{D}_{\lambda_{\rm opt}}$
\begin{align*}
  S(F, \epsilon)
  =\left\{ \eta: F(\eta) \le \inf_{\eta \in S_{\alpha}}  F(\eta) + \epsilon \right\}.
\end{align*}
From Lemma~\ref{lemma:cvar-soln}, we have $S(P, 0) = \{\aq{\pmu}\}$.

First, we show
$\limsup_{n \to \infty} \sqrt{|I_k|} \left(\lambda_{\rm opt}(P_n +
  |I_k|^{-1/2} H_n) - \lambda_{\rm opt}(P_n)\right) \le d\lambda_{{\rm opt},
  P}(H)$. Letting $\eta_n \in S(P_n, |I_k|^{-1})$, note that
\makesmall{
\begin{align*}
  \sqrt{|I_k|} \left(
  \lambda_{\rm opt}(P_n + |I_k|^{-1/2} H_n) - \lambda_{\rm opt}(P_n) \right)
  & \le |I_k|^{1/2} \left( (P_n + |I_k|^{-1/2} H_n)(\eta_n) - P_n(\eta_n)
    + |I_k|^{-1}\right) \\
  & \le H(\eta_n) + \norm{H_n - H} + |I_k|^{-1/2}.
\end{align*}
}
Noting that $\Linf{\hmu - \pmu} \le \delta_n$ on the event $\event_{n, k}$, we
have $\eta_n \in S(P, |I_k|^{-1} + \alpha^{-1} \delta_n)$. We now show
$\lim \eta_n = \aq{\pmu}$ by arguing $\limsup \eta_n = \liminf \eta_n$. For
any convergent subsequence $\eta_{n_m}$, its limit $\eta\opt$ must be
contained in the singleton $S(P, 0)$: since $\eta \mapsto P(\eta)$ is
$(\alpha^{-1} + 1)$-Lipschitz, we have
$\eta\opt \in S(P, (\alpha^{-1}+1) |\eta_{n_m} - \eta\opt| + |I_k|^{-1} +
\alpha^{-1} \delta_{n_m})$. We conclude that
$\lim_{n \to \infty} H(\eta_n) = H(\aq{\pmu}) =  d\lambda_{{\rm opt},
  P}(H)$ by continuity of $H \in \mathbb{D}_0$.

Second, we show
$\liminf_{n \to \infty} \sqrt{|I_k|} \left(\lambda_{\rm opt}(P_n +
  |I_k|^{-1/2} H_n) - \lambda_{\rm opt}(P_n)\right) \ge d\lambda_{{\rm opt},
  P}(H)$. Letting $\eta_n \in S(P_n + |I_k|^{-1/2} H_n, |I_k|^{-1})$, we
have
\begin{align*}
  & \lambda_{\rm opt}(P_n + |I_k|^{-1/2} H_n) - \lambda_{\rm opt}(P_n) \\
  & \ge (P_n + |I_k|^{-1/2} H_n)(\eta_n) - |I_k|^{-1}
    - P_n(\eta_n) \\
  & \ge |I_k|^{-1/2} H(\eta_n) + |I_k|^{-1/2}\norm{H_n - H} - |I_k|^{-1}.
  % & \ge |I_k|^{-1/2} \inf_{\eta \in S(P_n + |I_k|^{-1/2} H_n, |I_k|^{-1})}  H(\eta_n) + |I_k|^{-1/2}\norm{H_n - H} - |I_k|^{-1} \\
  % & \ge  |I_k|^{-1/2} \inf_{\eta \in S(P_n, |I_k|^{-1/2}
  %   \norm{H_n} + |I_k|^{-1})}  H(\eta_n) + |I_k|^{-1/2}\norm{H_n - H} - |I_k|^{-1}
\end{align*}
Since we have the inclusion
\begin{align*}
  S(P_n + |I_k|^{-1/2} H_n, |I_k|^{-1})
  & \subseteq S(P_n, |I_k|^{-1/2}  \norm{H_n} + |I_k|^{-1}) \\
  & \subseteq S(P, |I_k|^{-1/2} \norm{H_n} + |I_k|^{-1} +
  \alpha^{-1}\delta_n)
\end{align*}
on the event $\event_{n, k}$, we again conclude that
$\lim \eta_n = \aq{\pmu}$.  Continuity of $H$ again gives the desired
inequality.

%%% Local Variables:
%%% mode: latex
%%% TeX-master: "main"
%%% End:

%%% Local Variables:
%%% mode: latex
%%% TeX-master: "main"
%%% End:

\section{Proof of Theorem~\ref{theorem:clt}: Part II}
\label{section:proof-clt-neyman}

Our goal in this section is to show that the term~\eqref{eqn:part-neyman}
converges to zero in probability. Throughout the section, we assume that
Assumptions~\ref{assumption:ig},~\ref{assumption:overlap},~\ref{assumption:residuals}
hold. Recalling the definition~\eqref{eqn:remainder} of the remainder function
$\rem(r)$, we begin by showing it is differentiable on $(0, 1)$. In the below
lemma, we use
\begin{align*}
  (\rall) \defeq (\pall) + r((\hall) - (\pall))
\end{align*}
to ease notation; see Appendix~\ref{section:proof-gateaux-deriv} for its proof.
\begin{proposition}
  \label{proposition:gateaux-deriv}
  On the event $\event_{n, k}$ defined in Eq.~\eqref{eqn:good-event},
  $\rem(r)$ is differentiable on $(0, 1)$, and
  \makesmall{
  \begin{align}
   & \rem'(r) =  \E_{X \sim P}\left[ (\hmu - \pmu)(X) \left(\frac{1}{\alpha} \indic{\rmu(X) \ge \aq{\rmu}}
      - \pthr(X) \right) \right] \nonumber \\
    & + r\E_{X \sim P}\left[(\hmone - \pmone)(X) \left\{ \frac{\pprop}{\rprop^2} \rthr(\hprop - \pprop)
      - \frac{2 \pprop}{\rprop} (\hthr - \pthr) 
      + \frac{\pthr}{\rprop} (\hprop - \pprop)
      \right\}(X) \right]  \label{eqn:deriv} \\
    & + r \E_{X \sim P}\left[(\hmzero - \pmzero)(X) \left\{ \frac{ (1-\pprop)}{(1-\rprop)^2} \rthr (\hprop - \pprop)
      + \frac{2 (1-\pprop)}{1-\rprop} (\hthr - \pthr) 
      - \frac{\pthr}{1-\rprop} (\hprop - \pprop)
      \right\}(X) \right]. \nonumber
  \end{align}
  }
\end{proposition}
\noindent Since $\hprop, \pprop \in [c, 1-c]$, and $\hthr \in [-M_h, M_h]$ on
$\event_{n, k}$, elementary calculations and repeated applications of Holder's
inequality yield
\begin{align}
  \sup_{r \in [0, 1]} |\rem'(r)| & \le \Linf{\hmu - \pmu}   \sup_{r \in [0, 1]} \Lone{\frac{1}{\alpha} \indic{\rmu(X) \ge \aq{\rmu}} - \pthr(X)} \nonumber \\
             & \qquad + C \Lone{\hthr - \pthr} \Linf{\hmu - \pmu} \nonumber \\
             & \qquad + C \Ltwo{\hprop - \pprop} \left(\Ltwo{\hmzero - \pmzero} + \Ltwo{\hmone - \pmone}\right)
               \label{eqn:deriv-bound}
\end{align}
where $C$ is a positive constant that only depends on $c$, and $M_h$. The last
term in the bound is bounded by $\delta_n n^{-1/2}$ by the definition of
$\event_{n, k}$.

We proceed by showing that the first two terms in the preceding bound is
$o_p(n^{-1/2})$. First, we show that $\Linf{\hmu - \pmu}$ controls how
quantiles of $\rmu$ converge; see Appendix~\ref{section:proof-quantile-conv}
for the proof.
\begin{lemma}
  \label{lemma:quantile-conv}
  On the event $\event_{n, k}$, we have
  \begin{equation*}
    \sup_{r \in [0, 1]}
    \left| \aq{\rmu} - \aq{\pmu} \right|
    = O\left(\Linf{\hmu - \pmu}\right) =  O(\delta_n n^{-1/3}).
  \end{equation*}
\end{lemma}
\noindent Using Lemma~\ref{lemma:quantile-conv}, we are able to control the
rate of convergence of $\hthr$ to $\pthr$; see
Appendix~\ref{section:proof-thr-conv} for the proof of the below lemma.
\begin{lemma}
  \label{lemma:thr-conv}
  On the event $\event_{n, k}$, we have
  \begin{align*}
    \Lone{\hthr - \pthr} 
    & \lesssim n^{1/6} \left( \Lone{\hmu - \pmu}  + | \what{q}_k - \aq{\hmu}| 
      + | \aq{\hmu} - \aq{\pmu}| \right) \\
    & \qquad + n^{-1/6} \delta_n.
  \end{align*}
\end{lemma}
\noindent We conclude that the second term in the bound~\eqref{eqn:deriv-bound} is
$O(\delta_n n^{-1/2})$ on the event $\event_{n, k}$.

To bound the first term in the bound~\eqref{eqn:deriv-bound}, we use an
argument parallel to the proof of Lemma~\ref{lemma:thr-conv}---where we use
uniform differentiability~\eqref{eqn:unif-diff} of $F_{\rmu}$---to obtain
\begin{align*}
  & \sup_{r \in [0, 1]} \Lone{\frac{1}{\alpha} \indic{\rmu(X) \ge \aq{\rmu}} - \pthr(X)} \\
  & \lesssim n^{1/6} \left( \Lone{\hmu - \pmu}  + | \what{q}_k - \aq{\hmu}| 
    +\sup_{r \in [0, 1]} \left| \aq{\rmu} - \aq{\pmu} \right| \right) \\
  & \qquad + n^{-1/6} \delta_n + \Linf{\hmu - \pmu}.
\end{align*}
By Lemma~\ref{lemma:quantile-conv}, we once again conclude the first term in
the bound~\eqref{eqn:deriv-bound} is $O(\delta_n n^{-1/2})$.

Combining these results, we have shown
$\sup_{r \in [0, 1]} |\rem'(r)| = o(n^{-1/2})$ on the event $\event_{n, k}$.
Recalling $\rem(1) = \rem(0) + \rem'(r) \cdot 1 = \rem'(r)$, we conclude
$|\rem(1)| = o_p(n^{-1/2})$ since $\P(\event_{n, k}) \to 1$ from
Assumptions~\ref{assumption:neyman},~\ref{assumption:regularity}.

\subsection{Proof of Proposition~\ref{proposition:gateaux-deriv}}
\label{section:proof-gateaux-deriv}

We first show that the dual formulation is Gateaux differentiable. See
Appendix~\ref{section:proof-cvar-diff} for the proof.
\begin{lemma}
  \label{lemma:cvar-diff}
  On the event $\event_{n, k}$ defined in Eq.~\eqref{eqn:good-event},
  for all $r \in [0, 1]$, we have
  \begin{align*}
    \frac{d}{dr} \inf_{\eta}\left\{
    \frac{1}{\alpha} \E_{X \sim P} \hinge{\rmu(X) - \eta}
    + \eta
    \right\} 
    = \frac{1}{\alpha} \E\left[(\hmu - \pmu) \indic{\rmu(X) \ge \aq{\rmu}}\right].
  \end{align*}
\end{lemma}
\noindent To compute the Gateaux derivatives of the augmentation term,
$\frac{d}{dr} \E[\kappa(D; \rall)]$, we first verify interchange of
derivatives and expectations hold on the event $\event_{n, k}$.
\begin{lemma}
  \label{lemma:leibniz}
  On the event $\event_{n, k}$, we have
  \begin{align*}
    \frac{d}{dr} \E_{D \sim P}\left[\kappa(D; \rall)\right]
    = \E_{D \sim P}\left[ \frac{d}{dr} \kappa(D; \rall) \right]
  \end{align*}
\end{lemma}
\begin{proof-of-lemma}
  Under Assumptions~\ref{assumption:ig},~\ref{assumption:overlap},~\ref{assumption:residuals},~\ref{assumption:neyman},
  elementary calculations give
\begin{align*}
  & \E_{D \sim P}|\kappa(D; \rrall) - \kappa(D; \rtall)|^2 \\
  & \lesssim t^2 \left( \Ltwo{\pmzero - \hmzero} + \Ltwo{\pmone - \hmone}
    + \Ltwo{\pprop - \hprop} + \Ltwo{\pthr - \hthr}\right)
    \lesssim t^2
\end{align*}
on the event $\event_{n, k}$.  The result follows from dominated convergence.
\end{proof-of-lemma}

We now directly calculate $\frac{d}{dr} \kappa(D; \rall)$. On the event
$\event_{n, k}$, we have
\makesmall{
\begin{align*}
  & \frac{d}{dr}  \E_{D \sim P}\left[ \frac{Z}{\rprop(X)} (Y - \rmone(X))\right] \\
  & =  - \E\left[
  \frac{Z(\hprop - \pprop)(X)}{\rprop^2(X)} \rthr(X) (Y - \rmone(X)) \right]
    + \E\left[ \frac{Z}{\rprop(X)} (\hthr - \pthr)(X) (Y - \rmone(X)) \right] \\
  & \qquad \qquad
    - \E\left[ \frac{Z}{\rprop(X)} \rthr(X) (\hmone - \pmone)(X) \right] \\
  & = \E\left[
  \frac{r\pprop}{\rprop^2} \rthr (\hmone - \pmone) (\hprop - \pprop)(X)
    \right]
    - \E\left[ \frac{r\pprop}{\rprop} (\hthr - \pthr)(\hmone - \pmone)(X) \right]
    - \E\left[ \frac{\pprop}{\rprop} \rthr (\hmone - \pmone)(X) \right] \\
  & = r \E\left[
    \frac{\pprop}{\rprop^2} \rthr (\hmone - \pmone) (\hprop - \pprop)(X) \right]
    - r\E\left[\frac{2\pprop}{\rprop} (\hthr - \pthr)(\hmone - \pmone)(X)\right] \\
  & \qquad \qquad
    + r \E\left[\frac{\pthr(\hprop - \pprop)}{\rprop} (\hmone - \pmone)(X)\right]
    - \E\left[\pthr (\hmone - \pmone)(X) \right]
\end{align*}
}
where we used the tower law
$\E_{D \sim P}[\cdot] = \E_{D \sim P}[ \E_{D \sim P}[\cdot \mid X]]$ and
ignorability (Assumption~\ref{assumption:ig}) in the second
equality. Similarly, we have
\makesmall{
\begin{align*}
  & \frac{d}{dr}  \E_{D \sim P}\left[ \frac{1-Z}{1-\rprop(X)}
    (Y - \rmzero(X))\right] \\
  & =  - \E\left[
  \frac{(1-Z)(\hprop - \pprop)(X)}{(1-\rprop)^2(X)} \rthr(X) (Y - \rmzero(X)) \right] \\
    & \qquad + \E\left[ \frac{1-Z}{1-\rprop(X)} (\hthr - \pthr)(X) (Y - \rmzero(X)) \right] 
   - \E\left[ \frac{1-Z}{1-\rprop(X)} \rthr(X) (\hmzero - \pmzero)(X) \right] \\
  & = \E\left[
  - r\frac{1-\pprop}{(1-\rprop)^2} \rthr (\hmzero - \pmzero) (\hprop - \pprop)(X)
    \right]
    - r \E\left[ \frac{1-\pprop}{1-\rprop} (\hthr - \pthr)(\hmzero - \pmzero)(X) \right] \\
  & \qquad \qquad
    - \E\left[ \frac{1-\pprop}{1-\rprop} \rthr (\hmzero - \pmzero)(X) \right] \\
  & = - r \E\left[
    \frac{1-\pprop}{(1-\rprop)^2} \rthr (\hmzero - \pmzero) (\hprop - \pprop)(X) \right]
    - 2r\E\left[\frac{1-\pprop}{1-\rprop}
    (\hthr - \pthr)(\hmzero - \pmzero)(X)\right] \\
  & \qquad \qquad
    + r \E\left[\frac{\pthr}{1-\rprop} (\hprop - \pprop) (\hmzero - \pmzero)(X)\right]
    - \E\left[\pthr (\hmzero - \pmzero)(X) \right]
\end{align*}
}
where we used the tower law
$\E_{D \sim P}[\cdot] = \E_{D \sim P}[ \E_{D \sim P}[\cdot \mid X]]$ and
ignorability (Assumption~\ref{assumption:ig}) again in the second
equality. Collecting these derivations alongside Lemma~\ref{lemma:cvar-diff},
we obtain expression~\eqref{eqn:deriv}.

\subsubsection{Proof of Lemma~\ref{lemma:cvar-diff}}
\label{section:proof-cvar-diff}

Recalling Lemma~\ref{lemma:dual}, for any $r \in [0, 1]$,
Assumption~\ref{assumption:regularity} implies
\begin{align*}
  \inf_{\eta} \left\{ \frac{1}{\alpha} \E_{X \sim P} \hinge{\rmu(X) - \eta}
  + \eta \right\}  = \frac{1}{\alpha} \E_{X \sim P} \hinge{\rmu(X) - \aq{\rmu}}
  + \aq{\rmu}.
\end{align*}
We will compute the derivative of the right hand side. Instead of applying
Danskin's theorem, we show differentiability of the above quantity directly as
it requires less stringent assumptions. (Danskin's theorem requires $\rmu$ to
have a positive density \emph{on a neighborhood} of $\aq{\rmu}$, while our
following approach use Assumption~\ref{assumption:regularity}, which requires
positive density only at $\aq{\rmu}$.)

We begin by showing $|\aq{\rrmu} - \aq{\rmu}| = O(t)$ as $t \to 0$.  By
definition of quantiles, for any $\epsilon_t > 0$
\begin{align}
  \label{eqn:quantile-rrmu}
  F_{\rrmu}(\aq{\rrmu} - \epsilon_t)
  \le 1-\alpha \le F_{\rrmu}(\aq{\rrmu}).
\end{align}
Choose $\epsilon_t = o(t)$, and write
$F_{\rrmu}(\aq{\rrmu}) = F_{\rmu}(\aq{\rrmu}) + (F_{\rrmu} -
F_{\rmu})(\aq{\rrmu})$. Noting that $\Linf{\hmu - \pmu} \le \delta_n$ on
$\event_{n, k}$, we have
\begin{align*}
   & |(F_{\rrmu} - F_{\rmu})(\aq{\rrmu})| \\
   & \le F_{\rrmu}(\aq{\rrmu} + t\delta_n) -
     F_{\rrmu}(\aq{\rrmu} - t\delta_n) = O(t)
\end{align*}
since $F_{\rrmu}$ has a density at $\aq{\rrmu}$ by the definition of the set
$\mc{U}$, given in the paragraph before
Assumption~\ref{assumption:regularity}.  Hence, we have
$F_{\rrmu}(\aq{\rrmu}) \le F_{\rmu}(\aq{\rrmu}) + O(t)$, and using an
identical reasoning,
$F_{\rrmu}(\aq{\rrmu} - \epsilon_t) \ge F_{\rmu}(\aq{\rrmu} - \epsilon_t) +
O(t)$. Plugging these back into the definition~\eqref{eqn:quantile-rrmu} of
quantiles, we arrive at
\begin{align}
  & F_{\rmu}(\aq{\rrmu} - \epsilon_t) + O(t) \nonumber \\
  & \le 1-\alpha \le F_{\rmu}(\aq{\rrmu}) + O(t).   \label{eqn:quantile-t}
\end{align}
Since $F_{\rmu}$ is monotone and bounded away from $1-\alpha$ on any small
neighborhood of $\aq{\rmu}$, we conclude $\aq{\rrmu} \to \aq{\rmu}$. To see
that the convergence occurs at the rate $O(t)$, use Taylor's theorem to get
\makesmall{
\begin{align*}
  F_{\rmu}(\aq{\rrmu} - \epsilon_t)
  & = 1-\alpha + (f_{\rmu}(\aq{\rmu}) + o(1))
    \left(\aq{\rrmu} - \aq{\rmu}\right) + o(t), \\
  F_{\rmu}(\aq{\rrmu})
  & = 1-\alpha + (f_{\rmu}(\aq{\rmu}) + o(1))
    \left(\aq{\rrmu} - \aq{\rmu}\right) + o(t).
\end{align*}
}
Plugging the above display into the inequality~\eqref{eqn:quantile-t}, we
conclude $|\aq{\rrmu} - \aq{\rmu}| = O(t)$.

Building on this convergence result, we now prove the following
converges to the claimed derivative as $t\to 0$
\begin{align}
  \label{eqn:pre-derivative}
  \frac{1}{t} \Bigg\{
  & \frac{1}{\alpha} \E_{X \sim P} \hinge{\rrmu(X) - \aq{\rrmu}}
  + \aq{\rrmu} \\
  & - \frac{1}{\alpha} \E_{X \sim P} \hinge{\rmu(X) - \aq{\rmu}}
  - \aq{\rmu}
  \Bigg\}. \nonumber
\end{align}
First, note that if
$\rmu(X) - \aq{\rmu} > t \Linf{\hmu - \pmu} + |\aq{\rrmu} - \aq{\rmu}|$, 
\begin{align*}
  & \hinge{\rrmu(X) - \aq{\rrmu}}
    - \hinge{\rmu(X) - \aq{\rmu}} \\
   &  = t (\hmu - \pmu) - (\aq{\rrmu} - \aq{\rmu}).
\end{align*}
Similarly, if
$\rmu(X) - \aq{\rmu} < -t \Linf{\hmu - \pmu} - |\aq{\rrmu} - \aq{\rmu}|$,
\begin{align*}
  \hinge{\rrmu(X) - \aq{\rrmu}}
  - \hinge{\rmu(X) - \aq{\rmu}} = 0.
\end{align*}
Using this fact, we arrive at the bound
\makesmall{
\begin{align*}
  & \Big| \E_{X \sim P}\left[ \hinge{\rrmu(X) - \aq{\rrmu}}
  - \hinge{\rmu(X) - \aq{\rmu}} \right] \\
  & - \E_{X \sim P}\left[
  \left( t (\hmu - \pmu) - (\aq{\rrmu} - \aq{\rmu}) \right)
  \indic{\rmu(X) \ge \aq{\rmu}}
  \right]
    \Big| \\
  & \le 2 (t \Linf{\hmu - \pmu} + |\aq{\rrmu} - \aq{\rmu}|) \\
  & \qquad \times
    \P_{X \sim P}\left( |\rmu - \aq{\rmu}| \le t \Linf{\hmu - \pmu} + |\aq{\rmu} -
    \aq{\rrmu}| \right) = O(t^2),
\end{align*}
}
where we used that $F_{\rmu}$ has a density at $\aq{\rmu}$.  Since
$\E[\indic{\rmu(X) \ge \aq{\rmu}}] = \alpha$, we conclude that
expression~\eqref{eqn:pre-derivative} is equal to
\begin{align*}
  \frac{1}{\alpha} \E[ (\hmu - \pmu) \indic{\rmu(X) \ge \aq{\rmu}}] + O(t).
\end{align*}

\subsection{Proof of Lemma~\ref{lemma:quantile-conv}}
\label{section:proof-quantile-conv}

The proof mirrors that of Lemma~\ref{lemma:cvar-diff}.  First, we show that
$\sup_{r \in [0, 1]} |\aq{\rmu} - \aq{\pmu}| \to 0$ as $n \to \infty$,
conditional on the event $\event_{n, k}$. By definition of quantiles, for any
$\epsilon_n > 0$
\begin{align}
  \label{eqn:quantile-rmu}
  F_{\rmu}(\aq{\rmu} - \epsilon_n)
  \le 1-\alpha \le F_{\rmu}(\aq{\rmu}).
\end{align}
Choose $\epsilon_n = o(\Linf{\hmu - \pmu}) = o(\delta_n n^{-1/3})$, and write
\begin{equation*}
  F_{\rmu}(\aq{\rmu}) = F_{\pmu}(\aq{\rmu}) + (F_{\rmu} -
  F_{\pmu})(\aq{\rmu}).
\end{equation*}
Noting that $\Linf{\hmu - \pmu} \le \delta_n n^{-1/3}$
on $\event_{n, k}$, we have
\begin{align*}
 & \sup_{r \in [0, 1]} |(F_{\rmu} - F_{\pmu})(\aq{\rmu})|\\
&  \le \sup_{r \in [0, 1]} F_{\rmu}\left(\aq{\rmu} + \Linf{\hmu - \pmu}\right) -
                                                              F_{\rmu}\left(\aq{\rmu} - \Linf{\hmu - \pmu}\right) \\
  & = O\left(\Linf{\hmu - \pmu}\right)
\end{align*}
since $F_{\rmu}$ has a density at $\aq{\rmu}$ uniformly in $r \in [0, 1]$, by
the definition of the set $\mc{U}$ as given in the paragraph before
Assumption~\ref{assumption:regularity}.  Hence, we have
$F_{\rmu}(\aq{\rmu}) \le F_{\pmu}(\aq{\rmu}) + O\left( \Linf{\hmu - \pmu}
\right)$ uniformly in $r \in [0,1]$, and using an identical reasoning,
$F_{\rmu}(\aq{\rmu} - \epsilon_t) \ge F_{\pmu}(\aq{\rmu} - \epsilon_t) +
O\left(\Linf{\hmu - \pmu}\right)$ uniformly in $r \in [0,1]$. Plugging these
back into the definition~\eqref{eqn:quantile-rmu} of quantiles, we arrive at
\begin{align*}
  & F_{\pmu}(\aq{\rmu} - \epsilon_t) + O\left(\Linf{\hmu - \pmu} \right) \\
  & \le 1-\alpha \le F_{\pmu}(\aq{\rmu}) + O\left(\Linf{\hmu - \pmu} \right),
\end{align*}
uniformly in $r \in [0, 1]$. Since $F_{\pmu}$ is monotone and bounded away
from $1-\alpha$ on any small neighborhood of $\aq{\pmu}$, we conclude that
$\aq{\rmu} \to \aq{\pmu}$ uniformly in $r \in [0,1]$.

From Taylor theorem, uniformly over $r \in [0, 1]$
\begin{align*}
  F_{\pmu}(\aq{\rmu} - \epsilon_t)
  & = 1-\alpha + (f_{\pmu}(\aq{\pmu}) + o(1))
    \left(\aq{\rmu} - \aq{\pmu}\right) \\
  & \qquad + o\left(\Linf{\hmu - \pmu} \right), \\
  F_{\pmu}(\aq{\rmu})
  & = 1-\alpha + (f_{\pmu}(\aq{\pmu}) + o(1))
    \left(\aq{\rmu} - \aq{\pmu}\right) \\
  & \qquad + o\left(\Linf{\hmu - \pmu} \right).
\end{align*}
We hence conclude
$\sup_{r \in [0,1]} |\aq{\rmu} - \aq{\pmu}| = O\left(\Linf{\hmu - \pmu}
\right)$, since $F_{\pmu}$ has a strictly positive density
$f_{\pmu}(\aq{\pmu})$ at $\aq{\pmu}$.

\subsection{Proof of Lemma~\ref{lemma:thr-conv}}
\label{section:proof-thr-conv}

We define the following Lipschitz version of the indicator function:
\begin{equation*}
  \indica{w \ge 0}
  \defeq \begin{cases}
    1 ~~ & \mbox{if}~w \ge 0 \\
    ax + 1 ~~ & \mbox{if}~ -1/a \le  w  < 0 \\
    0 ~~ & \mbox{if}~w < -1/a
  \end{cases}.
\end{equation*}
By construction, $w \mapsto \indica{w \ge 0}$ is $a$-Lipschitz.  For a
sequence of constants $a_n \uparrow \infty$, which we choose later, we will
approximate each threshold function $\hthr(x)$ with $\indican{\cdot \ge 0}$.
By definition,
\makesmall{
\begin{align*}
  \alpha |\hthr - \pthr|(x)
  & \le \left|\indic{\hmu(x) - \what{q}_k \ge 0}
    - \indican{\hmu(x) - \what{q}_k \ge 0} \right| \\
  & \qquad +  \left|\indican{\hmu(x) - \what{q}_k \ge 0}
    - \indican{\pmu(x) - \aq{\pmu} \ge 0} \right| \\
  & \qquad +  \left|\indican{\pmu(x) - \aq{\pmu} \ge 0} - \indic{\pmu(x) - \aq{\pmu} \ge 0} \right| \\
  & \le \indic{\hmu(x) - \what{q}_k \in [-a_n^{-1}, 0]} \\
  & \qquad + a_n \left| \hmu(x) - \pmu(x) \right|
    + a_n \left| \what{q}_k - \aq{\hmu} \right|
    +  a_n \left| \aq{\hmu} - \aq{\pmu} \right| \\
  & \qquad + \indic{\pmu(x) - \aq{\pmu} \in [-a_n^{-1}, 0]}.
\end{align*}
}
Taking expectations,  we get
\makesmall{
\begin{align*}
  \alpha \Lone{\hthr - \pthr}
  & \le a_n \Lone{\hmu - \pmu} + a_n \left| \what{q}_k - \aq{\hmu} \right|
    +  a_n \left| \aq{\hmu} - \aq{\pmu} \right|  \\
  &  + \P_{X \sim P}\left(\hmu(X) - \what{q}_k \in [-a_n^{-1}, 0]\right)
    + \P_{X \sim P}\left(\pmu(X) - \aq{\pmu} \in [-a_n^{-1}, 0]\right)
\end{align*}
}
Since $|\what{q}_k - \aq{\hmu}| \le \delta_n n^{-1/3}$ on $\event_{n, k}$, and
both $F_{\hmu}$ and $F_{\pmu}$ have densities at $\aq{\hmu}$ and $\aq{\pmu}$,
we obtain the result by letting $a_n = n^{1/6}$.

%%% Local Variables:
%%% mode: latex
%%% TeX-master: "main"
%%% End:

%%% Local Variables:
%%% mode: latex
%%% TeX-master: "main"
%%% End:

\section{Proof of Theorem~\ref{theorem:clt}: Part III}
\label{section:proof-clt-var}

Since $\E[\hinge{\pmu(X) - \aq{\pmu}} \kappa(D; \pall)] = 0$ by ignorability
(Assumption~\ref{assumption:ig}), we have $\var~\psi(D) = \sigma^2_\alpha$,
where $\sigma^2_{\alpha}$ was defined in expression~\eqref{eqn:var}. Below, we
show that the two terms in the below empirical estimate
\begin{align*}
  \what{\sigma}^2_{\alpha, k}
      = \frac{1}{\alpha^2} \var_{X \sim \empfold} \hinge{\hmu(X) - \hq}
  + \var_{D \sim \empfold} \left( \augment\left(D; \hall\right)\right)
\end{align*}
respectively converge to their population counterparts defined with the true
nuisance parameter $(\pall)$. From Slutsky's lemma, this will give our final
result
$\frac{\sqrt{n}}{\what{\sigma}_{\alpha}} (\what{\omega}_{\alpha} -
\mbox{WTE}_{\alpha}) \cd N(0, 1)$.

We first show the convergence
\begin{align}
  \label{eqn:var-consistency}
  \var_{D \sim \what{P}_k} \hinge{\hmu(X) - \hq} \cp \var_{D \sim P}\hinge{\pmu(X) - \aq{\pmu}}.
\end{align}
We use the following weak law of large numbers.
\begin{lemma}[{\citet[Corollary 2.1.14]{Dembo16}}]
  \label{lemma:wlln-tri}
  Suppose that for each $m$, the random variables $\xi_{m,i},\,i=1,\dots,m$
  are pairwise independent, identically distributed for each $m$, and
  $\E[|\xi_{m,1}|] < \infty$.  Setting $S_m = \sum_{i = 1}^m \xi_{m,i}$
  and $a_m = \sum_{i=1}^m \E\overline{\xi}_{m,i}$,
  \begin{equation*}
    m^{-1}(S_m - a_m) \cp 0~\mbox{as}~m \to \infty.
  \end{equation*} 
\end{lemma}
\noindent Recalling the existence of an envelope function $\bar{\mu}(X)$
in Assumption~\ref{assumption:neyman}, Lemma~\ref{lemma:wlln-tri} implies
\begin{equation*}
  \var_{D \sim \what{P}_k} \hinge{\hmu(X) - \hq}
  - \var_{D \sim P} \hinge{\hmu(X) - \hq} \cp 0
\end{equation*}
conditional on $\{D_i\}_{i \in \cfoldinf}$. From dominated convergence, we
also have the unconditional convergence. From the above display, it
suffices for the claim~\eqref{eqn:var-consistency} to show
$\var_{D \sim P} \hinge{\hmu(X) - \hq} \cp \var_{D \sim P} \hinge{\pmu(X) -
  \aq{\pmu}}$.  Since $|\hq - \aq{\hmu}|\le \delta_n n^{-1/3}$ on the event
$\event_{n, k}$ defined in expression~\eqref{eqn:good-event}, we have
\begin{align*}
  \var_{D \sim P} \hinge{\hmu(X) - \hq}
  - \var_{D \sim P} \hinge{\hmu(X) - \aq{\hmu}} \cp 0.
\end{align*}
Now, since  $\Ltwo{\hmu - \pmu} \cp 0$ and $\aq{\hmu} \cp \aq{\pmu}$
from Lemma~\ref{lemma:quantile-conv}, we conclude
\begin{equation*}
  \var_{D \sim P} \hinge{\hmu(X) - \hq} \cp \var_{D \sim P} \hinge{\pmu(X) -
    \aq{\pmu}}.
\end{equation*}

To show $\var_{D \sim \what{P}_k} \kappa(D; \hall) \cp \var \kappa(D; \pall)$,
we begin by noting that on the event $\event_{n, k}$,
$|\kappa(D; \hall)| \le \frac{M_h}{c} (2|Y| +
\bar{\mu}(X))$. Lemma~\ref{lemma:wlln-tri} again yields
\begin{align*}
  \var_{D \sim \what{P}_k} \kappa(D; \hall)
  - \var_P \kappa(D; \hall) \cp 0,
\end{align*}
conditional on $\event_{n, k}$ and $\{D_i\}_{i \in \cfoldinf}$, and hence
unconditionally as well.  Under
Assumption~\ref{assumption:ig},~\ref{assumption:overlap},~\ref{assumption:residuals},~\ref{assumption:neyman},
elementary calculations show
\begin{align*}
  & \E_{D \sim P}[|\kappa(D; \pall) - \kappa(D; \hall)|^2] \\
  & \lesssim \Ltwo{\pmzero - \hmzero} + \Ltwo{\pmone - \hmone}
  + \Ltwo{\pprop - \hprop} + \Ltwo{\pthr - \hthr}
\end{align*}
on the event $\event_{n, k}$. From Lemma~\ref{lemma:thr-conv}, we conclude
\begin{equation*}
  \var~\kappa(D; \hall) \cp \var~\kappa(D; \pall),
\end{equation*}
which gives the desired result.

%%% Local Variables:
%%% mode: latex
%%% TeX-master: "main"
%%% End:

\section{Proof of Theorem~\ref{theorem:efficiency}}
\label{section:proof-efficiency}

\subsection{Formal background}

To formally define the semiparametric efficiency bound, we begin by reviewing
requisite definitions, which are are standard in the semi-parametric
statistics literature. See~\citet{Tsiatis07} and~\citet{VanDerVaart98} for an
overview of standard results. We use the following classical definition
(e.g. see~\citet[Appendix A]{Newey90}) which guarantees the Hajek-Le Cam bound
is a well-defined asymptotic efficiency bound. A set of parameterized
densities of a random vector $\xi$ is smooth in the quadratically mean
differentiable (QMD) sense if it satisfies the following properties.
\begin{definition}
  \label{def:qmd}
  Let $f(\xi; \theta)$ be a density over $\xi$, parameterized by
  $\theta \in \Theta \subseteq \R^r$, defined with respect to some
  $\sigma$-finite measure $\nu$. $\{f(\cdot; \theta)\}_{\theta \in \Theta}$ is
  \emph{smooth in the quadratically mean differentiable} sense if $\Theta$ is
  open, $\theta \mapsto f(\xi; \theta)$ is continuous $\nu$-a.e., and
  $\sqrt{f(\xi; \theta)}$ is differentiable in quadratic mean: there exists a
  score function $S(\xi; \theta) \in \R^r$ such that
  $\int \ltwo{S(\xi; \theta)}^2 f(\xi; \theta) d \nu < \infty$, 
  \begin{equation*}
    \int (\sqrt{f(\xi; \theta')} - \sqrt{f(\xi; \theta)}
    - \sqrt{f(\xi; \theta)} S(\xi; \theta)^\top (\theta' - \theta))^2 d\nu = o(\ltwo{\theta - \theta'}^2),
  \end{equation*}
  and 
  $\int \ltwo{S(\xi; \theta) \sqrt{f(\xi; \theta)} - S(\xi; \theta')
    \sqrt{f(\xi; \theta')}}^2 d\nu \to 0$ as $\ltwo{\theta - \theta'} \to 0$.
\end{definition}
\noindent For a class of parameterized densities satisfying above, the Fisher
information matrix is
\begin{equation*}
  I(\theta) \defeq 4 \int S(\xi; \theta) S(\xi; \theta)^\top f(\xi; \theta) d\nu.
\end{equation*}
A class of regular conditional probability kernels
$f(\xi_1 \mid \xi_2; \theta)$ is smooth in the QMD sense if
Definition~\ref{def:qmd} holds for $f(\cdot | \xi_2; \theta)$ for a.s. all
$\xi_2$, and
{\small $\norm{\E[\ltwo{S(\xi_1 | \xi_2; \theta)}^2 \mid \xi_2]}_{L^{\infty}(\Xi_2)} <
\infty$}.

The density of $D = (X, Y, Z)$ separates by ignorability
(Assumption~\ref{assumption:ig})
\begin{equation*}
  f(d) = (f_1(y \mid x) e\opt(x))^z  (f_{0}(y \mid x) (1-e\opt(x)))^{1-z} f_X(x),
\end{equation*}
where $f_z(\cdot | \cdot)$ is the conditional density of $Y \mid X, Z = z$ for
$z \in \{0, 1\}$, and $f_X(\cdot)$ is the density of $X$.  Consider
parameteric submodels of $f(d)$, given by $\theta \in \Theta$
\begin{equation}
  \label{eqn:submodel}
  f(d; \theta) \defeq (f_1(y \mid x; \theta) e(x; \theta))^z (f_{0}(y \mid x;
  \theta) (1-e(x; \theta)))^{1-z} f_X(x; \theta),
\end{equation}
where $f_0(y | x; \theta), f_1(y | x; \theta), f_X(x; \theta), e(x; \theta)$
are submodels of their respective true versions. We let $\theta\opt$ be the
true parameter so that $f(d; \theta\opt) = f(d)$ under
Assumption~\ref{assumption:ig}. We denote the expectation under $f(d; \theta)$
by $\E_{\theta}[\cdot]$, and let
$\mu_z(X; \theta) \defeq \E_{\theta}[Y(z) \mid X]$ for $z \in \{0, 1\}$, and
$\mu(X; \theta) \defeq \mu_1(X; \theta) - \mu_0(X; \theta)$.

We consider \emph{smooth} parametric submodels over which the Hajek-Le Cam
bound holds.
\begin{definition}
  \label{def:smooth}
  A parametric submodel $\{f(\cdot; \theta)\}_{\theta \in \Theta}$ is
  \emph{smooth} if
\begin{enumerate}
\item All subcomponents are smooth (Definition~\ref{def:qmd}),
  where $f_X(x; \theta)$ is defined w.r.t. the Lebesgue measure. We let
  $S_z(y | x; \theta), S_X(x; \theta)$ be the score functions of
  $f_z(y | x; \theta)$ and $f_X(x; \theta)$ resp.. \label{condition:qmd}
\item \label{condition:lipschitz} For $z \in \{0,1\}$, there is a neighborhood
  $N(\theta\opt)$ such that
  \begin{equation*}
    \sup_{\theta \in N(\theta\opt)}
    \Linf{ \E_{\theta}[Y(1) S_1(Y(1)| X; \theta) \mid X]
      - \E_{\theta}[Y(0) S_0(Y(0)| X; \theta) \mid X]}
    < \infty
  \end{equation*}
\item Overlap holds: $e(x; \theta) \in [c/2, 1-c/2]$, where $c$ is given in
  Assumption~\ref{assumption:overlap} \label{condition:overlap}
\item $\E_{\theta}[Y(z)^2] < \infty$, and $\Linf{\E_{\theta}[(Y(z) - \mu_z(X; \theta))^2 \mid X]} < \infty$ \label{condition:residuals}
\end{enumerate}
If $\{f(\cdot; \theta)\}_{\theta \in \Theta}$ is \emph{smooth}, and its Fisher
information $I(\theta)$ is nonsingular for every $\theta \in \Theta$, we say
that the submodel is \emph{regular}.
\end{definition}
\noindent Condition~\ref{condition:qmd} is a standard condition required for
the Hajek-Le Cam bound to hold. Condition~\ref{condition:lipschitz} guarantees
that if we denote by $P_{\theta}$ the distribution with density
$f(D; \theta)$, then $\theta \mapsto \mbox{WTE}_{\alpha}(P_{\theta})$ is
differentiable at $\theta\opt$. Conditions~\ref{condition:overlap}
and~\ref{condition:residuals} do not change the efficiency bound, but are
standard conditions required for estimators of $\mbox{WTE}_{\alpha}$ to be
asymptotically linear.

Since the Hajek-Le Cam bound holds for smoothly parameterized statistical
functionals, we restrict attention to \emph{differentiable} functionals,
following the by now standard approach of~\citet{KoshevnikLe77}
and~\citet{PfanzaglWe85}.
\begin{definition}
  \label{def:diff}
  We say that a statistical functional $Q \mapsto \beta(Q) \in \R$ is
  \emph{differentiable} if for all \emph{smooth} parametric submodels of the
  true data-generating distribution $P$, $\theta \mapsto \beta(P_{\theta})$ is
  differentiable at the true model $\theta\opt$ ($P_{\theta\opt} = P$), and
  there exists a random variable $\phi$ such that
  $\nabla_{\theta} \beta(P_{\theta\opt}) = \E[\phi S(\xi; \theta\opt)]$.
\end{definition}

In order to exclude superefficient estimators that exhibit dependence on the
local data generating process, we consider estimators that conform to a notion
of uniformity. For a parametric submodel, consider a local data generating
process (LDGP) where the data is distributed from a model $\theta_n$ for each
sample size $n$, and $\sqrt{n}(\theta_n - \theta\opt)$ is bounded. An
estimator $\what{\beta}$ of $\beta(P)$ is regular for this parametric submodel
if $\sqrt{n}(\what{\beta} - \beta(P))$ has a limiting distribution that
doesn't depend on the LDGP. An estimator is \emph{regular} if it is regular in
every parametric submodel, and the limiting distribution of
$\sqrt{n} (\what{\beta} - \beta(P))$ does not depend on the submodel.  The
class of regular statistical estimators precludes nonsmooth superefficient
behavior~\citep{BickelKlRiWe93, Newey90} which are unstable due to high
sensitivity to undetectable changes in the data-generating process; an
estimator's distribution should remain similar when the parameter changes only
by a small amount.~\citet{Newey90} states that ``in finite samples
superefficient estimators do worse than MLEs in neighborhoods of the point of
superefficiency''. See~\citet[Section 3.1]{Tsiatis07} for a discussion of
pitfalls of superefficiency.

The classical Hajek-Le Cam convolution theorem~\cite[Theorem
8.8]{VanDerVaart98} shows that any regular estimator has asymptotic variance
at least
$\nabla_{\theta} \beta(P_{\theta\opt})^\top I(\theta\opt)^{-1} \nabla_{\theta}
\beta(P_{\theta\opt})$; going outside the regular class of estimators do not
help much as improvements beyond this bound can only be made on a measure zero
set of parameters~\cite[Theorem 8.9]{VanDerVaart98}.  Moreover, the Hajek-Le
Cam local asymptotic minimax theorem~\cite[Theorem 8.11]{VanDerVaart98} states
that for any regular parametric (sub)model, any \emph{arbitrary} estimator
$\what{\beta}_n$ suffers the following mean squared error in the local minimax
sense
\begin{align*}
  \sup_{A \subset \R^r: \mbox{\scriptsize finite}}
  \liminf_{n \to \infty} \sup_{h \in A}
  ~n \cdot \E_{\theta\opt + v / \sqrt{n}}  \cdot \left(
  \what{\beta}_n - \beta(P_{\theta\opt + v/\sqrt{n}})\right)^2
  \ge \nabla_{\theta} \beta(P_{\theta\opt})^\top
    I(\theta\opt)^{-1} \nabla_{\theta} \beta(P_{\theta\opt}).
\end{align*}
The Hajek-Le Cam bound
$\nabla_{\theta} \beta(P_{\theta\opt})^\top I(\theta\opt)^{-1} \nabla_{\theta}
\beta(P_{\theta\opt})$ can also be thought of as an asymptotic Cramer-Rao
bound, as it coincides with the finite-sample variance bound for unbiased
estimators.

Since the error of any semiparametric estimator is bounded below by the
Hajek-Le Cam bound of any regular parametric submodel, the
\emph{semiparametric efficiency bound} is defined as the supremum of the
Hajek-Le Cam bound over all regular parametric submodels
\begin{equation*}
  \mathbb{V}(\beta; P) \defeq \sup\left\{
    \nabla_{\theta} \beta(P_{\theta\opt})^\top
    I(\theta\opt)^{-1} \nabla_{\theta} \beta(P_{\theta\opt}):
    ~\mbox{all regular parametric submodels}
  \right\}.
\end{equation*}
By construction, any regular semiparametric estimator of $\beta(P)$ has
asymptotic variance necessarily larger than $V(\beta; P)$.

\begin{remark}
  Using classical arguments~\citep[Theorem 2.2]{Newey90}, we can formally show
  that our augmented estimator $\what{\omega}_{\alpha}$ is \emph{regular}
  under sufficient smoothness of
  $\pmzero(\cdot), \pmone(\cdot), \pprop(\cdot)$ that guarantee standard
  nonparametric estimators (e.g. sieves~\citep{Chen07}) converge at
  the rates required by Assumption~\ref{assumption:neyman}.  These additional
  conditions can be added to our definition of smoothness above; carefully
  following the proof of Theorem~\ref{theorem:efficiency}, we can see that
  these additional smoothness requirements do not affect the semiparametric
  efficiency bound. We omit the details since this is a purely theoretical
  concern, and is of little practical consequence; the true practical power of
  Algorithm~\ref{alg:cross-fitting} comes from its ability to fit nuisance
  parameters using black-box ML models.
\end{remark}

\subsection{Proof of main result}

We proceed in the style of classical arguments outlined in~\citep{Newey90,
  BickelKlRiWe93}, affording special care to establish differentiablity of the
functional $\mbox{WTE}_{\alpha}$.

We first characterize the tangent space $\mc{T}$, the $L^2(P)$-closure of the
space spanned by elements of the score function $S(D; \theta\opt)$, for all
\emph{smooth} parametric submodels
\makesmall{
\begin{align*}
  \mc{T} \defeq \Big\{
  s& (D): ~ \E[s(D)^2] < \infty,~\mbox{and}~\exists a(m) \in \R^{r_m},
           ~\mbox{and sequence of \emph{smooth} parametric submodels} \\
         & \mbox{with score}~ S(D; \theta\opt(m)) ~\mbox{at}~\theta\opt =
           \theta\opt(m)
           ~~\mbox{s.t.}~
           ~\E[(s(D) - a(m)^\top S(D; \theta\opt(m)))^2] \to 0~\mbox{as}~m \to \infty
           \Big\}.
\end{align*}
}
From the decomposition~\eqref{eqn:submodel} (which holds under
ignorability~\ref{assumption:ig}), the score of a \emph{smooth} parametric
submodel of $D$ is given by
\makesmall{
\begin{align}
  \label{eqn:score}
  S(D; \theta)
  = Z S_1(Y | X; \theta) + (1-Z) S_0(Y | X; \theta) + \frac{Z - e(X; \theta)}{e(X; \theta) (1-e(X;\theta))} \nabla_{\theta} e(x; \theta)
  + S_X(X; \theta).
\end{align}
}
Using this formula, we derive an explicit representation for the tangent
space.
\begin{lemma}
  \label{lemma:tangent}
  Under the conditions of Theorem~\ref{theorem:efficiency}, the tangent space
  $\mc{T}$ is the set of measurable functions
  $d \mapsto z S_1(y \mid x) + (1-z) S_0(y \mid x) + A(x)(z - \pprop(x)) +
  S_X(x)$ s.t. $\E[S_X(x)] = 0$,
  $\E[S_z(Y(z) \mid X) \mid X] = 0$ a.s., and
  $\E[A^2(X)] < \infty$.
\end{lemma}
\noindent See Appendix~\ref{section:proof-tangent} for the proof.  With this
characterization of the tangent space, our efficiency bound follows from the
following classical result; see~\citet[Theorem 1, Section 3.3]{BickelKlRiWe93}
and~\citet[Theorem 3.1]{Newey90}.
\begin{lemma}
  \label{lemma:geometric}
  Let $\beta$ be a differentiable parameter, and let the tangent space
  $\mc{T}_{\beta}$ be linear. For $\phi$ defined in Definition~\ref{def:diff},
  let us denote by $\Pi(\phi \mid \mc{T})$ the $L^2$-projection of $\phi$ to
  $\mc{T}$, which is well-defined since $\mc{T}$ is a linear Hilbert space.
  If $\E[\Pi(\phi \mid \mc{T})^2] > 0$, then
  $\mathbb{V}(\beta; P) = \E[ \Pi(\phi \mid \mc{T})^2]$.
\end{lemma}
\noindent The tangent space characterized in Lemma~\ref{lemma:tangent} is
clearly linear. The below lemma---whose proof we give in
Appendix~\ref{section:proof-diff}---shows differentiability of
$\theta \mapsto \mbox{WTE}_{\alpha}(P_{\theta})$ at $\theta\opt$.
\begin{lemma}
  \label{lemma:diff}
  Let the conditions of Theorem~\ref{theorem:efficiency} hold. Then, for any
  \emph{smooth} parameteric submodel,
  $\theta \mapsto \mbox{WTE}_{\alpha}(P_{\theta})$ is differentiable at
  $\theta\opt$ with
  \begin{align*}
    \nabla_{\theta} \mbox{WTE}_{\alpha}(P_{\theta\opt})
    =  & \frac{1}{\alpha} \int  \hinge{\pmu(x) - \aq{\pmu}} 
    S_X(x; \theta\opt) f_X(x) dx \\
    & + \int \nabla_{\theta} \mu(x; \theta\opt) h\opt(x)  f_X(x) dx,
  \end{align*}
  where integrals over vectors is taken elementwise, and we use
  \begin{equation*}
  \nabla_{\theta} \mu(x; \theta\opt) = \int y S_1(y | x; \theta\opt)
  f_1(y | x) d\nu(y) - \int y S_0(y | x; \theta\opt) f_0(y | x) d\nu(y).
  \end{equation*}
\end{lemma}
\noindent Recalling the score function $S(D; \theta\opt)$ given in
expression~\eqref{eqn:score}, and the influence
function~\eqref{eqn:influence}, we have
$\nabla_{\theta} \mbox{WTE}_{\alpha}(P_{\theta\opt}) = \E[\psi(D) S(D;
\theta\opt)]$ by inspection. We conclude that $\mbox{WTE}_{\alpha}$ is a
\emph{differentiable} parameter in the sense of Definition~\ref{def:diff}.

The influence function $\psi$ is in the tangent space $\mc{T}$. Define
$A(x) = 0$, and
\begin{align*}
  S_X(x) & \defeq \frac{1}{\alpha} \hinge{\pmu(x) - \aq{\pmu}} + \aq{\pmu} - \mbox{WTE}_{\alpha}(P) \\
  S_1(y | x) & \defeq \frac{\pthr(x)}{\pprop(x)} (y - \pmone(x)), ~~~~
               S_0(y | x) \defeq -\frac{\pthr(x)}{1-\pprop(x)} (y - \pmzero(x)).
\end{align*}
From Lemma~\ref{lemma:tangent} and
Assumptions~\ref{assumption:overlap},~\ref{assumption:residuals}, we have
$\psi \in \mc{T}$. So the first result follows from Lemma~\ref{lemma:geometric}.

When the true propensity score $\pprop(\cdot)$ is known, a similar argument as
Lemma~\ref{lemma:tangent} shows that the tangent space is given by
$d \mapsto z S_1(y \mid x) + (1-z) S_0(y \mid x) + S_X(x)$ s.t.
$\E[S_X(x)] = 0$, and $\E[S_z(Y(z) \mid X) \mid X] = 0$ a.s.. From the tangent
space, a nearly identical argument as above gives the same efficiency bound.

\subsection{Proof of Lemma~\ref{lemma:tangent}}
\label{section:proof-tangent}

Given the form of the score function~\eqref{eqn:score} at $\theta = \theta\opt$,
the definition of \emph{smoothness} (Definition~\ref{def:smooth}) implies that
$\mc{T}$ is included in the set characterized in the statement of the lemma.

To show the other direction, let $S_1(y \mid x), S_0 (y \mid x), A(x), S_X(x)$
be bounded functions that satisfy conditions characterized in the
statement. Let $\theta \in \R$ be small enough, such that
\begin{align*}
  & f_z(y | x; \theta) \defeq f_z(y \mid x) (1+\theta S_z(y\mid x))
    ~~\mbox{for}~~z \in \{0, 1\}, \\
  & f_X(x; \theta) \defeq f_X(x)(1+ \theta S_X(x)),
  ~~~e(x; \theta) \defeq e\opt(x) + \theta A(x) e\opt(x) (1-e\opt(x)),
\end{align*}
are valid kernels/densities/propensity scores.
Standard arguments~\citep[Prop. 2.1.1]{BickelKlRiWe93} show that the above
quantities are smoothness in the QMD sense (Definition~\ref{def:qmd}), with
scores at $\theta\opt$ given by $S_z(y \mid x)$, $S_X(x)$, and
$A(x)(z - e\opt(x))$.
At other values of $\theta$, we have
$S_z(y | x; \theta) = (1+\theta S_z(y |x))^{-1} S_z(y | x)$.  For
$z \in \{0, 1\}$, boundedness of $S_z(y | x)$ and $S_X(x)$, and
Assumption~\ref{assumption:efficiency} implies
that there exists a neighborhood $N(\theta\opt)$ such that
\begin{align*}
  \sup_{\theta' \in N(\theta)} \Linf{\int y S_z(y |x; \theta) f_z(y | x;\theta)
  d\nu} < \infty.
\end{align*}
It is also straightforward to verify conditions~\ref{condition:overlap},
and~\ref{condition:residuals}.  Since bounded functions are dense in the space
of $L^2$ functions, we conclude $f(D; \theta)$ is \emph{smooth} in the sense
of Definition~\ref{def:smooth}.

\subsection{Proof of Lemma~\ref{lemma:diff}}
\label{section:proof-diff}

From condition~\ref{condition:qmd} of Definition~\ref{def:smooth}, we have
\begin{equation*}
  \nabla_{\theta} \mu(x; \theta) = \int y S_1(y | x; \theta)
  f_1(y | x; \theta)) d\nu(y) - \int y S_0(y | x; \theta) f_0(y | x; \theta) d\nu(y).
\end{equation*}
From condition~\ref{condition:lipschitz} of Definition~\ref{def:smooth}, there
exists a neighborhood of $\theta\opt$, and a constant $C > 0$ such that
$\Linf{\ltwo{\nabla_{\theta} \mu(x; \theta)}} \le C$ over $\theta$ in this
neighborhood. This implies that for all $t$ small enough, we have
\begin{equation}
  \label{eqn:sup-bound-mu}
  \Linf{\mu(x; \theta\opt + t \theta) - \mu(x; \theta\opt)}
  \le t C \ltwo{\theta}.
\end{equation}

We begin by showing that as $t \to 0$,
\begin{align}
  & P_{1-\alpha, \theta\opt + t \theta}^{-1}(\mu(\cdot; \theta\opt + t\theta))
    - P_{1-\alpha, \theta\opt}^{-1}(\mu(\cdot; \theta\opt)) \nonumber \\
    & = P_{1-\alpha, \theta\opt + t \theta}^{-1}(\mu(\cdot; \theta\opt + t\theta))
  - \aq{\pmu}
  = O(t)    \label{eqn:quantile-conv-param}
\end{align}
Let us denote by $F_{\theta}$ the cumulative distribution of $\mu(x; \theta)$
under $f_X(\cdot; \theta)$, and let $\epsilon_t$ be a strictly positive
sequence such that $\epsilon_t= o(t)$.  Write
\begin{align*}
  & F_{\theta\opt + t \theta}
  \left(P_{1-\alpha, \theta\opt + t \theta}^{-1}(
    \mu(\cdot; \theta\opt + t\theta))\right) \\
  & = F_{\theta\opt}
  \left(P_{1-\alpha, \theta\opt + t \theta}^{-1}(
    \mu(\cdot; \theta\opt + t\theta))\right) 
  + (F_{\theta\opt + t \theta} - F_{\theta\opt})
  \left(P_{1-\alpha, \theta\opt + t \theta}^{-1}(
  \mu(\cdot; \theta\opt + t\theta))\right),
\end{align*}
and note that since $\pmu(X)$ has a strictly positive density, we have
\makesmall{
\begin{align*}
  & \left| (F_{\theta\opt + t \theta} - F_{\theta\opt})
  \left(P_{1-\alpha, \theta\opt + t \theta}^{-1}(
  \mu(\cdot; \theta\opt + t\theta))\right) \right|  \\
  & \le F_{\theta\opt}\left(P_{1-\alpha, \theta\opt + t \theta}^{-1}(
  \mu(\cdot; \theta\opt + t\theta)) + t C \ltwo{\theta}\right)
  - F_{\theta\opt}\left(P_{1-\alpha, \theta\opt + t \theta}^{-1}(
  \mu(\cdot; \theta\opt + t\theta)) - t C \ltwo{\theta}\right) = O(t).
\end{align*}
}
Arguing similarly for the input
$P_{1-\alpha, \theta\opt + t \theta}^{-1}( \mu(\cdot; \theta\opt + t\theta)) -
\epsilon_t$, we get
\begin{align*}
  & F_{\theta\opt + t \theta} \left(P_{1-\alpha, \theta\opt + t \theta}^{-1}(
  \mu(\cdot; \theta\opt + t\theta))\right) = F_{\theta\opt} \left(P_{1-\alpha,
    \theta\opt + t \theta}^{-1}( \mu(\cdot; \theta\opt + t\theta))\right) +
  O(t) \\
  & F_{\theta\opt + t \theta} \left(P_{1-\alpha, \theta\opt + t \theta}^{-1}(
    \mu(\cdot; \theta\opt + t\theta)) - \epsilon_t\right) = F_{\theta\opt}
    \left(P_{1-\alpha, \theta\opt + t \theta}^{-1}( \mu(\cdot; \theta\opt +
    t\theta)) - \epsilon_t\right) + O(t).
\end{align*}
Recalling the definition of
$P_{1-\alpha, \theta\opt + t \theta}^{-1}(\mu(\cdot; \theta\opt + t\theta))$,
conclude
\makesmall{
\begin{align}
  & F_{\theta\opt + t \theta}
  \left(P_{1-\alpha, \theta\opt + t \theta}^{-1}(\mu(\cdot; \theta\opt + t\theta))
  - \epsilon_t\right)
  = F_{\theta\opt} \left(P_{1-\alpha,
  \theta\opt + t \theta}^{-1}( \mu(\cdot; \theta\opt + t\theta)) - \epsilon_t\right) +
  O(t)  \nonumber \\
  & \le 1-\alpha
  \le  F_{\theta\opt + t \theta}
  \left(P_{1-\alpha, \theta\opt + t \theta}^{-1}(
  \mu(\cdot; \theta\opt + t\theta))\right)
  = F_{\theta\opt}
    \left(P_{1-\alpha, \theta\opt + t \theta}^{-1}( \mu(\cdot; \theta\opt +
    t\theta)) \right) + O(t).
    \label{eqn:quantile-inequality}
\end{align}
}
Since $F_{\theta\opt}$ has a positive density at $\aq{\pmu}$, we have shown
$P_{1-\alpha, \theta\opt + t \theta}^{-1}(\mu(\cdot; \theta\opt + t\theta))
\to P_{1-\alpha, \theta\opt}^{-1}(\mu(\cdot; \theta\opt))$.

To show that the convergence happens at rate $O(t)$,  Taylor theorem yields
\begin{align*}
& F_{\theta\opt} \left(P_{1-\alpha,
  \theta\opt + t \theta}^{-1}( \mu(\cdot; \theta\opt + t\theta))\right) \\
  & = 1-\alpha + f_{\theta\opt}(\aq{\pmu})\left( P_{1-\alpha, \theta\opt + t \theta}^{-1}(\mu(\cdot; \theta\opt + t\theta))
    - P_{1-\alpha, \theta\opt}^{-1}(\mu(\cdot; \theta\opt))
    \right) \\
  & \qquad + o\left( P_{1-\alpha, \theta\opt + t \theta}^{-1}(\mu(\cdot; \theta\opt + t\theta))
    - P_{1-\alpha, \theta\opt}^{-1}(\mu(\cdot; \theta\opt))
    \right) \\
    & F_{\theta\opt} \left(P_{1-\alpha,
  \theta\opt + t \theta}^{-1}( \mu(\cdot; \theta\opt + t\theta)) - \epsilon_t \right) \\
  & = 1-\alpha + f_{\theta\opt}(\aq{\pmu}) \left( P_{1-\alpha, \theta\opt + t \theta}^{-1}(\mu(\cdot; \theta\opt + t\theta))
    - P_{1-\alpha, \theta\opt}^{-1}(\mu(\cdot; \theta\opt))
    \right) \\
  & \qquad + o\left( P_{1-\alpha, \theta\opt + t \theta}^{-1}(\mu(\cdot; \theta\opt + t\theta))
    - P_{1-\alpha, \theta\opt}^{-1}(\mu(\cdot; \theta\opt))
    \right) + o(t),
\end{align*}
where we used $f_{\theta\opt}$ to denote the (positive) density of
$F_{\theta\opt}$.  Plugging these approximations into the
inequality~\eqref{eqn:quantile-inequality}, the desired
convergence~\eqref{eqn:quantile-conv-param} follows.

We are now ready to directly show differentiability of the mapping
$\theta \mapsto \mbox{WTE}_{\alpha}(P_{\theta})$. We draw heavily on the dual
representation of $\mbox{WTE}_{\alpha}$ (see Lemma~\ref{lemma:dual}). Recalling Lemma~\ref{lemma:dual}, for $t \ge 0$
\begin{align*}
  \mbox{WTE}_{\alpha}(P_{\theta\opt + t \theta})
  & = \frac{1}{\alpha} \E_{\theta\opt + t \theta}
  \hinge{\mu(X; \theta\opt + t \theta) - P_{1-\alpha, \theta\opt + t \theta}^{-1}(
  \mu(\cdot; \theta\opt + t\theta))} \\
  & \qquad + P_{1-\alpha, \theta\opt + t \theta}^{-1}(
    \mu(\cdot; \theta\opt + t\theta)).
\end{align*}
Note that if
\begin{equation*}
  \pmu(X) - \aq{\pmu} > t C\ltwo{\theta} + \left|P_{1-\alpha, \theta\opt + t
    \theta}^{-1}(\mu(\cdot; \theta\opt + t\theta)) - P_{1-\alpha,
    \theta\opt}^{-1}(\mu(\cdot; \theta\opt)) \right|,
\end{equation*}
then
\begin{align*}
  & \hinge{\mu(X; \theta\opt + t \theta) - P_{1-\alpha, \theta\opt + t \theta}^{-1}(
    \mu(\cdot; \theta\opt + t\theta))}
  - \hinge{\mu(X; \theta\opt) - P_{1-\alpha, \theta\opt}^{-1}(
    \mu(\cdot; \theta\opt))} \\
  & = \mu(X; \theta\opt + t \theta) - \mu(X; \theta\opt)
    - \left(P_{1-\alpha, \theta\opt + t \theta}^{-1}(
    \mu(\cdot; \theta\opt + t\theta))
     - P_{1-\alpha, \theta\opt}^{-1}(
    \mu(\cdot; \theta\opt))\right)
\end{align*}
from the uniform bound~\eqref{eqn:sup-bound-mu}. Similarly, if
\begin{equation*}
  \pmu(X) - \aq{\pmu} < -t C\ltwo{\theta} - \left|P_{1-\alpha, \theta\opt + t
    \theta}^{-1}(\mu(\cdot; \theta\opt + t\theta)) - P_{1-\alpha,
    \theta\opt}^{-1}(\mu(\cdot; \theta\opt)) \right|,
\end{equation*}
then we have
\begin{align*}
  \hinge{\mu(X; \theta\opt + t \theta) - P_{1-\alpha, \theta\opt + t \theta}^{-1}(
    \mu(\cdot; \theta\opt + t\theta))}
  - \hinge{\mu(X; \theta\opt) - P_{1-\alpha, \theta\opt}^{-1}(
  \mu(\cdot; \theta\opt))}
  = 0.
\end{align*}
Since $\pmu(X)$ has a density at $\aq{\pmu}$, the above implies
\makesmall{
\begin{align}
  & \Bigg| \frac{1}{t} \int
  \left( \hinge{\mu(x; \theta\opt + t \theta)
  - P_{1-\alpha, \theta\opt + t \theta}^{-1}(
  \mu(\cdot; \theta\opt + t\theta))}
  - \hinge{\pmu(x) - \aq{\pmu}}\right) f_X(x) dx \nonumber \\
  & ~-   \alpha \int
  \theta^\top \nabla_{\theta} \mu(x; \theta\opt + t \theta)
    h\opt(x)  f_X(x) dx
    + \frac{\alpha}{t} \left(P_{1-\alpha, \theta\opt + t \theta}^{-1}(
  \mu(\cdot; \theta\opt + t\theta)) - \aq{\pmu}\right)
    \Bigg| \nonumber \\
  & \le 2 \left( C \ltwo{\theta} + \frac{1}{t}
    \left| P_{1-\alpha, \theta\opt + t \theta}^{-1}(
    \mu(\cdot; \theta\opt + t\theta)) - \aq{\pmu} \right|
    \right)  \nonumber \\
  & \qquad  \times P\left( | \pmu(X) - \aq{\pmu} | \le t C \ltwo{\theta}
    + | P_{1-\alpha, \theta\opt + t \theta}^{-1}(
    \mu(\cdot; \theta\opt + t\theta)) - \aq{\pmu} | \right)
    = o(1),
    \label{eqn:second-piece}
\end{align}
}
where we used the convergence~\eqref{eqn:quantile-conv-param} in the last
equality. Now, rewrite
\makesmall{
\begin{align*}
  & \frac{1}{t} \left( \mbox{WTE}_{\alpha}(P_{\theta\opt + t\theta})
    - \mbox{WTE}_{\alpha}(P_{\theta\opt}) \right) \\
  & = \frac{1}{t}
    \Bigg\{ \frac{1}{\alpha} \E_{\theta\opt + t \theta}
    \hinge{\mu(X; \theta\opt + t \theta) - P_{1-\alpha, \theta\opt + t \theta}^{-1}(
    \mu(\cdot; \theta\opt + t\theta))}
    + P_{1-\alpha, \theta\opt + t \theta}^{-1}(
    \mu(\cdot; \theta\opt + t\theta))   \\
  & \qquad \qquad - \frac{1}{\alpha} \E_{\theta\opt}
    \hinge{\mu(X; \theta\opt + t \theta) - P_{1-\alpha, \theta\opt + t \theta}^{-1}(
    \mu(\cdot; \theta\opt + t\theta))}
    - P_{1-\alpha, \theta\opt + t \theta}^{-1}(
    \mu(\cdot; \theta\opt + t\theta)) \Bigg\} \\
  & \qquad + \frac{1}{t} \Bigg\{ \frac{1}{\alpha} \E_{\theta\opt}
    \hinge{\mu(X; \theta\opt + t \theta) - P_{1-\alpha, \theta\opt + t \theta}^{-1}(
    \mu(\cdot; \theta\opt + t\theta))}
    + P_{1-\alpha, \theta\opt + t \theta}^{-1}(
    \mu(\cdot; \theta\opt + t\theta)) \\
  & \qquad \qquad \qquad - \frac{1}{\alpha} \E_{\theta\opt}
    \hinge{\pmu(X) - \aq{\pmu}}
    - \aq{\pmu} \Bigg\}.
\end{align*}
}
Using the bound~\eqref{eqn:second-piece}, the above display is equal to
\begin{align*}
  &  \int \left( \frac{1}{\alpha} \hinge{\mu(X; \theta\opt + t \theta) - P_{1-\alpha, \theta\opt + t \theta}^{-1}(
    \mu(\cdot; \theta\opt + t\theta))} + P_{1-\alpha, \theta\opt + t \theta}^{-1}(
    \mu(\cdot; \theta\opt + t\theta)) \right) \\
  & \qquad \times \frac{1}{t} \left( f_X(x; \theta\opt + t\theta) - f_X(x; \theta\opt)\right) dx
    + \int \theta^\top \nabla_{\theta} \mu(x; \theta\opt) h\opt(x)  f_X(x) dx
    + o(1) \\
  & =  \int \left( \frac{1}{\alpha} \hinge{\pmu(x) - \aq{\pmu}} + \aq{\pmu} \right)
    \theta^\top S_X(x; \theta\opt) f_X(x) dx \\
   & \qquad  + \int \theta^\top \nabla_{\theta} \mu(x; \theta\opt) h\opt(x)  f_X(x) dx
    + o(1),
\end{align*}
where the last inequality follows from dominated convergence and 
\emph{smoothness} of $f(D; \theta)$.

% To see that $\mu(X; \theta)$ has a strictly positive density on $\mc{X}$ under
% the distribution $f_X(\cdot; \theta)$, we use the decomposition
% \begin{equation*}
%   \mu(X; \theta) = \pmu(X) + \mu(X; \theta) - \pmu(X) = \pmu(X) + \theta \cdot r(X).
% \end{equation*}
% We use the following lemma due to~~\citet{ChernozhukovFeLu18}.
% \begin{lemma}[{\citet[Lemma A.1]{ChernozhukovFeLu18}}]
%   Under Assumption~\ref{assumption:efficiency}, $\pmu(X)$ and $r(X)$
%   have densities under $f_X(\cdot; \theta)$, respectively given by
%   \begin{equation*}
%     t \mapsto \int_{x: \pmu(x) = t} \frac{f_X(x; \theta)}{\norm{\nabla \pmu(x)}} ~d \mbox{\rm Vol},
%     ~~\mbox{and}~~
%     t \mapsto \int_{x: r(x) = t} \frac{f_X(x; \theta)}{\norm{\nabla r(x)}} ~d \mbox{\rm Vol},
%   \end{equation*}
%   where both integrals are well defined since
%   $x \mapsto \nabla \pmu(x), \nabla r(x)$ are finite, continuous, and bounded away from zero.
% \end{lemma}
% Since $f_X(x; \theta) > 0$ on $\mc{X}$, both $\pmu(X)$ and $r(X)$ have
% strictly positive densities on $\mc{X}$. By the convolution of probability
% densities, we conclude that $\mu(X; \theta)$ has a strictly positive density
% under $f_X(\cdot; \theta)$.

%%% Local Variables:
%%% mode: latex
%%% TeX-master: "main"
%%% End:

%%% Local Variables:
%%% mode: latex
%%% TeX-master: "main"
%%% End:

\end{document}